\documentclass{article}
\pdfoutput=1

\usepackage{jmlr2e}
\usepackage{lastpage}

\jmlrheading{23}{2022}{1-\pageref{LastPage}}{7/21; Revised
3/22}{4/22}{21-0808}{S\'ebastien Forestier, R\'emy Portelas, Yoan Mollard and Pierre-Yves Oudeyer}

\ShortHeadings{Intrinsically Motivated Goal Exploration Processes}{Forestier, Portelas, Mollard and Oudeyer}
\firstpageno{1}

\usepackage[utf8]{inputenc} 
\usepackage[T1]{fontenc}    
\usepackage[english]{babel}
\usepackage{hyperref}       
\usepackage{url}            
\usepackage{booktabs}       
\usepackage{amsfonts}       
\usepackage{nicefrac}       
\usepackage{microtype}      

\usepackage{tikz}
\usetikzlibrary{shapes,shadows}
\tikzstyle{abstractbox} = [draw=black, fill=white, rectangle, 
inner sep=10pt, drop shadow={fill=black,
opacity=0.5}]
\tikzstyle{abstracttitle} =[fill=white]

\newcommand{\boxabstract}[3][fill=white]{
\begin{center}
  \begin{tikzpicture}
    \node [abstractbox, #1] (box)
    {\begin{minipage}{0.98\linewidth}
        #3
      \end{minipage}};
    \node[abstracttitle, right=10pt] at (box.north west) {\large ~ #2 ~~};
  \end{tikzpicture}
\end{center}
}

\usepackage{amsmath}
\usepackage{amssymb}
\usepackage{mathabx}
\usepackage{bm}
\usepackage{graphicx}
\usepackage[noend]{algpseudocode} 
\usepackage{algorithm}
\usepackage{multirow}
\usepackage{array}
\usepackage{verbatim}
\usepackage{xcolor}

\makeatletter
\newenvironment{architecture}[1][htb]{%
    \renewcommand{\ALG@name}{Architecture}
   \begin{algorithm}[#1]%
  }{\end{algorithm}}
\makeatother

\makeatletter
\newcommand{\algmargin}{\the\ALG@thistlm}
\makeatother
\newlength{\whilewidth}
\algdef{SE}[parWHILE]{parWhile}{EndparWhile}[1]
  {\parbox[t]{\dimexpr\linewidth-\algmargin}{%
     \hangindent\whilewidth\strut\algorithmicwhile\ #1\ \algorithmicdo\strut}}{\algorithmicend\ \algorithmicwhile}%
\algnewcommand{\parState}[1]{\State%
  \parbox[t]{\dimexpr\linewidth-\algmargin}{\strut #1\strut}}
  
\makeatletter
\newlength{\trianglerightwidth}
\algnewcommand{\LineComment}[1]{\Statex \hskip\ALG@thistlm $\triangleright$ #1}
\algnewcommand{\LineCommentCont}[1]{\Statex \hskip\ALG@thistlm%
  \parbox[t]{\dimexpr\linewidth-\ALG@thistlm}{\hangindent=\trianglerightwidth \hangafter=1 \strut$\triangleright$ #1\strut}}
\makeatother

\usepackage[caption=false,font=footnotesize]{subfig}

\begin{document}

\title{Intrinsically Motivated Goal Exploration Processes with Automatic Curriculum Learning}

\author{\name Sébastien Forestier \email seb@forest.bio \\
    \addr Inria Bordeaux Sud-Ouest \\
    200 avenue de la Vieille Tour, 33405 Talence, France
    \AND 
    \name Rémy Portelas \email remy.portelas@inria.fr \\
    \addr Inria Bordeaux Sud-Ouest \\
    200 avenue de la Vieille Tour, 33405 Talence, France
    \AND
    \name Yoan Mollard \email yoan.mollard@inria.fr \\
    \addr Inria Bordeaux Sud-Ouest \\
    200 avenue de la Vieille Tour, 33405 Talence, France
    \AND
    \name Pierre-Yves Oudeyer \email pierre-yves.oudeyer@inria.fr \\
    \addr Inria Bordeaux Sud-Ouest \\
    200 avenue de la Vieille Tour, 33405 Talence, France
}

\editor{George Konidaris}

\maketitle

\begin{abstract}
Intrinsically motivated spontaneous exploration is a key enabler of autonomous developmental learning in human children. It enables the discovery of skill repertoires through autotelic learning, i.e. the self-generation, self-selection, self-ordering and self-experimentation of learning goals. We present an algorithmic approach called Intrinsically Motivated Goal Exploration Processes (IMGEP) to enable similar properties of autonomous learning in machines. The IMGEP architecture relies on several principles: 1) self-generation of goals, generalized as parameterized fitness functions; 2) selection of goals based on intrinsic rewards; 3) exploration with incremental goal-parameterized policy search and exploitation with a batch learning algorithm; 4) systematic reuse of information acquired when targeting a goal for improving towards other goals. We present a particularly efficient form of IMGEP, called AMB, that uses a population-based policy and an object-centered spatio-temporal modularity. We provide several implementations of this architecture and demonstrate their ability to automatically generate a learning curriculum within several experimental setups. One of these experiments includes a real humanoid robot exploring multiple spaces of goals with several hundred continuous dimensions and with distractors. While no particular target goal is provided to these autotelic agents, this curriculum allows the discovery of diverse skills that act as stepping stones for learning more complex skills, e.g. nested tool use.
\\
\end{abstract}


\section{Introduction}

An extraordinary property of natural intelligence in humans is their capacity for lifelong autonomous learning. 
During cognitive development, processes of autonomous learning in infants have several properties that are fundamentally different from many current machine learning systems. 
Among them is the capability to spontaneously explore their environments, driven by an intrinsic motivation to discover and learn new tasks and problems that they imagine and select by themselves \citep{berlyne1966curiosity, gopnik1999scientist}. 
Crucially, there is no engineer externally imposing one target goal that they should explore, hand providing a curriculum for learning, nor providing a ready-to-use database of training examples. 
Rather, children are autotelic learners: they self-select their objectives within a large, potentially open-ended, space of goals they can imagine, and they collect training data by physically practicing these goals. 
In particular, they explore goals in an organized manner, attributing to them values of interestingness that evolve with time, and allowing them to self-define a learning curriculum that is called a developmental trajectory in developmental sciences \citep{thelen1996dynamic}. 
This self-generated learning curriculum prevents infants from spending too much time on goals that are either too easy or too difficult, and allows them to focus on goals of the right level of complexity at the right time. 
Within this developmental learning process, the new learned solutions are often stepping stones for discovering how to solve other goals of increasing complexity. 
Thus, while they are not explicitly guided by a final target goal, these mechanisms allow infants to discover highly complex skills. For instance, biped locomotion or tool use would be extremely difficult to learn by focusing only on these targets from the start as the rewards for those goals are typically rare or deceptive. 

An essential component of such organized spontaneous exploration is the intrinsic motivation system, also called curiosity-driven exploration system \citep{kidd2015psychology,oudeyer2016evolution,gottlieb2018towards}. In the last two decades, a series of computational and robotic models of intrinsically motivated exploration and learning in infants have been developed \citep{oudeyer_what_2007, baldassarre2013intrinsically,BAZHYDAI2020370}, opening new theoretical perspectives in neuroscience and psychology \citep{gottlieb_information-seeking_2013}.
Two key ideas have allowed to simulate and predict important properties of infant spontaneous exploration, ranging from vocal development \citep{moulin-frier_self-organization_2014,forestier2017unified}, to object affordance and tool learning \citep{forestier2016curiosity, forestier2016overlapping}. 
The first key idea is that infants might select experiments that maximize an intrinsic reward based on empirical learning progress \citep{oudeyer_intrinsic_2007}. This mechanism would generate automatically developmental trajectories (e.g. learning curricula) where progressively more complex tasks are practiced, learned and used as stepping stones for more complex skills. 
The second key idea is that beyond selecting actions or states based on the predictive learning progress they provide, a more powerful way to organize intrinsically motivated exploration is to select goals, i.e. self-generated fitness functions, based on a measure of competence progress, i.e. a measure of progress in learning to produce diverse and controlled behavioral features \citep{baranes2010maturationally, baranes_active_2013}. 
Here, the intrinsic reward is the empirical improvement towards solving self-selected goals \citep{oudeyer_what_2007, forestier2016curiosity}, happening through lower-level policy search mechanisms that generate physical actions. 
The efficiency of such goal exploration processes relies on a form of hindsight learning that leverages the fact that the data collected when targeting a goal can be informative to find better solutions to other goals (for example, a learner trying to achieve the goal of pushing an object on the right but actually pushing it on the left fails to progress on this goal, but learns as a side effect how to push it on the left). These general ideas have been instantiated and studied in the context of population-based learning architectures (e.g. \cite{baranes_active_2013,pere2018unsupervised}), as well as more recently in goal-conditioned reinforcement learning architectures (e.g. \cite{colas2018curious,nair2018rig,choi_variational_2021}), used to build \textbf{autotelic agents} \citep{colas2020intrinsically}.

Beyond neuroscience and psychology, we believe these models open new perspectives in artificial intelligence, contributing to the foundations of the new field of \textit{developmental artificial intelligence} \citep{eppe2021intelligent}. 
In particular, algorithmic architectures for intrinsically motivated goal exploration were shown to allow the efficient acquisition of repertoires of high-dimensional motor skills with automated curriculum learning in several robotics experiments \citep{baranes_active_2013, forestier2016curiosity}. 
This includes for example learning omnidirectional locomotion or learning multiple ways to manipulate complex soft objects \citep{rolf,baranes_active_2013}. \\ \\

In this article, we make several contributions:
\begin{itemize}
\item We present a formalization of Intrinsically Motivated Goal Exploration Processes (IMGEP), that is both more compact and more general than these previous models. In particular, it considers a generalized definition of the concept of goals, construed as abstract parameterized fitness functions that can express arbitrary objectives over full behavioural trajectories and include constraints. This enables to express a diversity of goal exploration algorithms in the same framework, including Quality-Diversity algorithms that were not previously formalized as goal exploration algorithms.
\item We present a new population-based IMGEP algorithmic architecture, called AMB, implementing two forms of object-centered modularity and using learning-progress to sample associated modular goal spaces. First, we introduce spatial modularity: each object of the environment is associated to a goal space. Second, we introduce temporal modularity: the temporal structure of objects' movement is leveraged for more  efficient leveraging of discovered stepping-stones in goal exploration, through a stepping-stone preserving mutation operator (SSPMutation). We also present various instantiations.
\item We present a systematic experimental study of this new IMGEP algorithm in diverse environments providing opportunities for discovering complex skills like tool use, as well as including complex distractors: a 2D simulated environment, a Minecraft environment, and a real humanoid robotic setup. 
We compare several variants of IMGEP algorithms, including ablations, in terms of sample efficiency to discover a diversity of behavioral features. We also compare IMGEPs algorithms with algorithms exploring only one target object: we show that letting agents self-organize exploration of diverse objects is vastly more efficient for discovering how to control the target object than channeling the agent to explore only this object. 
We also compare the exploration resulting from the self-organized learning curriculum of intrinsically motivated agents with the exploration following a curriculum designed by hand with expert knowledge of the task, showing a similar exploration efficiency.
\end{itemize}


\begin{figure}
\boxabstract{Generalized Goals and Goal Spaces}{

In the general case, the agent has algorithmic tools to construct any goal as any function $f_g$ (parameterized by vector $g$), taking as input a state-action trajectory $\tau$, and returning the fitness of $\tau$ for achieving the goal. As a shorthand, we call the goal $g$ (vector of parameters), but the full representation of the goal consists in the combination of the parameters $g$ \textit{and} the function or program computing the fitness function\footnote{The term "fitness function" is here a synonym for "goal-achievement reward function" sometimes used in related work.}.

\paragraph{}
Formally, given a behavioural trajectory  $\tau=\{s_{t_0},a_{t_{0}}, \cdots, s_{t_{end}},a_{t_{{end}}}\}$, we assume the agent can compute a set of \textbf{behavioural features} $\varphi_1(\tau), ... , \varphi_n(\tau)$, also called \textbf{outcomes}, over the full trajectory.
Those features are vectors that encode any static or dynamic property of the environment or the agent itself. 
Examples include the average speed of an object, an encoding of an object's trajectory or relation to other objects, the frequency of a movement, or the result of a test checking whether a movement or an object fulfills some properties specified in a sub-part of vector $g$. The features may be given to the agent, or learned, for instance using generative models (see \citet{pere2018unsupervised, nair2018rig, pmlr-v87-laversanne-finot18a,reinke2019intrinsically}).

\paragraph{}
We also assume that computational tools, in the form of mathematical operators\footnote{These operators are pre-defined in existing work and in this paper, but they could in principle be learned.}, are available to the agent for constructing goals, i.e. for constructing fitness functions
$f_g$ using these features. Examples include:
\begin{itemize}
    \item $f_g(\tau) = \varphi_{g}(\tau)$: the goal is to produce a trajectory $\tau$ that maximizes feature $\varphi_{g}(\tau)$, where the goal parameter vector $g$ is a simple one dimensional index of the target behavioural feature. Sampling the space of goals thus amounts to sampling which feature to maximize, e.g. maximize agent's speed or number of objects collected.
    \item $f_g(\tau) = -||\varphi_{\psi_{1}(g)}(\tau) - \psi_{2}(g)||$: the goal $g$ is to produce a trajectory $\tau$ so that its features $\varphi_{\psi_{1}(g)}(\tau)$ are as close as possible to the target vector $\psi_{2}(g)$, using a measure $||.||$, e.g. move the hand or a ball to a particular 3D position \citep{baranes_active_2013,pere2018unsupervised}, or produce a sound with a particular target spectrum \citep{moulin-frier_self-organization_2014}.
    \item $f_{g}(\tau) = \varphi_i(\tau) ~~\texttt{if}~~ \varphi_j(\tau) \in \psi_{3}(g) ~~\texttt{else}~~ 0$: the goal $g$ is to produce a trajectory $\tau$ which maximizes feature $i$ while keeping feature vector $j$ inside a local region specified by $\psi_{3}(g)$, e.g. maximize agent's speed while displaying a certain type of pattern of leg movement. When goal sampling is made in this form of function space, IMGEPs correspond to quality-diversity algorithms \citep{pugh2016quality} such as MAP-Elite \citep{cully2015robots}. 
    \item $f_{g}(\tau) = f_{g_1}(\tau) ~~\texttt{if}~~ f_{g_2}(\tau) < f_{g_3}(\tau) ~~\texttt{else}~~ f_{g_4}(\tau)$: goals can be combined to form more complex constrained optimization problems, e.g. move the ball to follow a target while not getting too close to the walls and holes and minimizing the energy spent.
\end{itemize}

\paragraph{}
A \textbf{goal space} is a set of goals (fitness functions) parameterized by a vector $g$. Diverse forms of structured parameterization can be used, as shown by the examples above, and corresponding to diverse types of goals.
In the experiments of this paper, we define several goal spaces, each with a parameterization representing the target trajectory of positions or sound or light of an object in the environment. 
Each object $k$ and modality $m$ defines a goal space $\mathcal{G}^{k,m}$ containing goals $g$ of the form $f^{k,m}_{g}(\tau) = -||\varphi_{\psi^{k,m}_{1}(g)}(\tau) - \psi^{k,m}_{2}(g)||$ where $\psi^{k,m}_{2}(g)$ denotes the target trajectory of object $k$ in modality $m$ (positions, sound or light), and $\varphi_{\psi^{k,m}_{1}(g)}(\tau)$ denotes the features of trajectory $\tau$ of the same object and modality.
}
\end{figure}

\section{Intrinsically Motivated Goal Exploration Processes}
\label{sec:IM}
    
We define a framework for the intrinsically motivated exploration of multiple goals, where the data collected when exploring a goal give some information to help reach other goals. 
This framework considers that when the agent performed an experiment, it can compute the fitness of that experiment for achieving any goal, not only the one it was trying to reach.
Importantly, it does not assume that all goals are achievable, nor that they are of a particular form, enabling to express complex objectives that do not simply depend on the observation of the end policy state but might depend on several aspects of entire behavioral trajectories (see Box on features, goals and goal spaces).
Also, the agent autonomously builds its goals but does not know initially which goals are achievable or not, which are easy and which are difficult, nor if certain goals need to be explored so that other goals become achievable.

\subsection{Notations and Assumptions}
Let's consider an agent that executes continuous \textbf{actions} $a \in \mathcal{A}$ in continuous \textbf{states} $s \in \mathcal{S}$ of an \textbf{environment} $E$.
We consider policies producing time-bounded rollouts through the dynamics $\delta_{E}(\bm{s_{t+1}}~|~\bm{s_{t_0:t}},\bm{a_{t_0:t}})$ of the environment, and we denote the corresponding behavioral trajectories $\tau = \{s_{t_0},a_{t_{0}}, \cdots, s_{t_{end}},a_{t_{{end}}}\} \in \mathbb{T}$. 


We assume that the agent is able to construct a goal space $\mathcal{G}$ parameterized by $g$, representing fitness functions $f_g$ giving the fitness $f_g(\tau)$ of an experimentation $\tau$ to reach a goal $g$ (see Box on features, goals and goal spaces).
Also, we assume that given a trajectory $\tau$, the agent can compute $f_g(\tau)$ for any $g \in \mathcal{G}$. 

Given $\mathcal{S}$, $\mathcal{A}$ and $\mathcal{G}$, the agent explores the environment by sampling goals in $\mathcal{G}$ and searching for good solutions to those goals, and learns a \textbf{goal-parameterized policy} $\Pi(\bm{a_{t+1}} ~|~\bm{g},\bm{s_{t_0:t+1}}, \bm{a_{t_0:t}})$ to reach any goal from any state.

We can then evaluate the agent's exploration and learning efficiency either by observing its behavior and estimating the diversity of its skills and the reached stepping-stones, or by ``opening'' agent's internal models and policies to analyze their properties.

\subsection{Algorithmic Architecture}

We present Intrinsically Motivated Goal Exploration Processes (IMGEP) as an algorithmic architecture that can be instantiated into many particular algorithms sharing several general principles (see pseudo-code in Architecture \ref{algo:IMGEP}):
\begin{itemize}
	\item The agent autonomously builds and samples goals as fitness functions, possibly using intrinsic rewards, 
	\item Two processes are running in parallel: 1) an exploration loop samples goals and searches for good solutions to those goals with the exploration policy; 2) an exploitation loop uses the data collected during exploration to improve the goal-parameterized policy over the goal space(s), 
	\item The data acquired when exploring solutions for a particular goal is reused to extract potential solutions to other goals. 
\end{itemize}

\begin{architecture}[t]
	\caption{~ Intrinsically Motivated Goal Exploration Process (IMGEP)}
	\label{algo:IMGEP}
	\begin{algorithmic}[1]
	\Require Action space $\mathcal{A}$, State space $\mathcal{S}$
	
	\State Initialize knowledge $\mathcal{E}=\emptyset$
	\State Initialize goal space $\mathcal{G}$ and goal policy $\Gamma$ 
	\State Initialize policies $\Pi$ and $\Pi_\epsilon$ 
	
	\State Launch asynchronously the two following loops:
	
	\Loop \Comment Exploration loop 
	\State Choose goal $g$ in $\mathcal{G}$ with $\Gamma$
	
	\State Execute a roll-out of $\Pi_\epsilon$, observe trajectory $\tau$
	\LineCommentCont{From now on $f_{g'}(\tau)$ can be computed to estimate the fitness of the current experiment $\tau$ for achieving any goal $g' \in \mathcal{G}$}
	\State Compute the fitness $f = f_g(\tau)$ associated to goal $g$
	\State Compute intrinsic reward $r_i = IR(\mathcal{E}, g, f)$ associated to $g$
	\State Update exploration policy $\Pi_\epsilon$ with ($\mathcal{E}, g, \tau, f$) \Comment e.g. fast incremental algo.
	\State Update goal policy $\Gamma$ with ($\mathcal{E}, g, \tau, f, r_i$)
	\State Update knowledge $\mathcal{E}$ with ($g, \tau, f, r_i$)
	\EndLoop

	\Loop \Comment Exploitation loop
	\State Update policy $\Pi$ with $\mathcal{E}$ \Comment e.g. batch training of deep NN, SVMs, GMMs
    \State Update goal space $\mathcal{G}$ with $\mathcal{E}$
	\EndLoop
	\State \Return $\Pi$
	
	\end{algorithmic}
\end{architecture}

\subsection{Goal Exploration}
\label{goal_exploration}

In the exploration loop, the agent samples a goal $g$, executes its exploration policy $\Pi_\epsilon$, and observes the resulting trajectory $\tau$. 
This new experiment $\tau$ can be used to:
\begin{itemize}
    \item compute the fitness associated to goal $g$,
    \item compute an intrinsic reward evaluating the interest of the choice of $g$,
    \item update the goal policy (sampling strategy) using this intrinsic reward,
    \item update the exploration policy $\Pi_\epsilon$ with a fast incremental learning algorithm, 
    \item update the learning database $\mathcal{E}$.
\end{itemize}
Then, asynchronously, this learning database $\mathcal{E}$ can be used to learn a target policy $\Pi$ with a slower or more computationally demanding algorithm, but on the other end resulting in a more accurate policy.
The goal space may also be updated based on this data.

\subsection{Intrinsic Rewards}
\label{intrinsic_rewards}

In goal exploration, a goal $g \in \mathcal{G}$ is chosen at each iteration. $\mathcal{G}$ may be infinite, continuous and of high-dimensionality, making the choice of goal important and non-obvious.
Indeed, even if the fitness function $f_{g'}(\tau)$ may give information about the fitness of a trajectory $\tau$ to achieve many goals $g' \in \mathcal{G}$, the policy leading to $\tau$ has been chosen with the goal $g$ to solve in mind, thus it may not give as much information about other goals than the execution of another policy chosen when targeting other goals.

Intrinsic rewards provide a mean for the agent to self-estimate the expected interest of exploring particular goals for learning how to achieve all goals in $\mathcal{G}$.
An intrinsic reward signal $r_i$ is associated to a chosen goal $g$, and based on a heuristic (denoted $IR$) such as outcome novelty, progress in reducing outcome prediction error, or progress in competence to solve problems \citep{oudeyer_what_2007}. 

\begin{figure}[ht]
\centering
\includegraphics[width=0.47\textwidth]{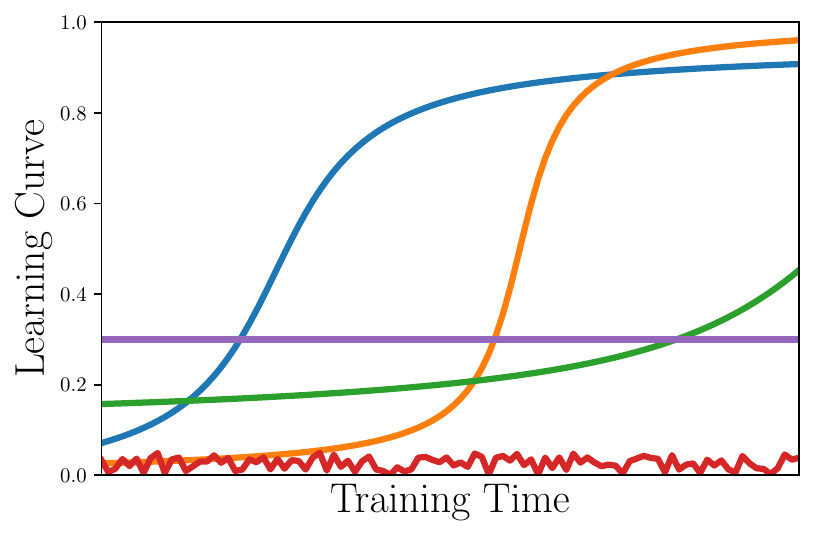}\hspace{0.2cm}
\includegraphics[width=0.47\textwidth]{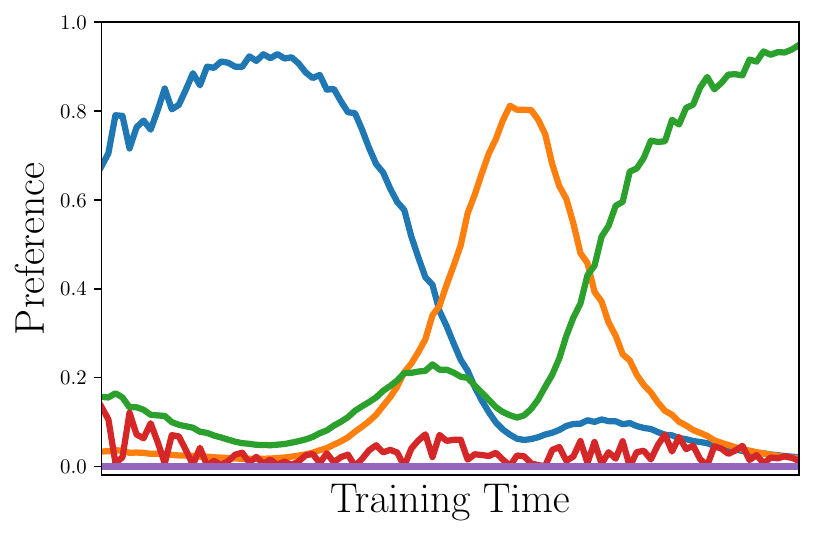}
\caption{Schematic representation of possible learning curves for different goals and the associated exploration preference for an agent with intrinsic rewards based on learning progress.
Left: schematic learning curves associated to $5$ imaginary goals: the y axis represents the competence of the agent to achieve the goal ($1$ is perfect, $0$ is chance level), and the x axis is training time on a goal. The blue, orange and green curves represent achievable goals, for which agent's competence increases with training, at different rates, and saturates after a long training time. The purple curve represents a goal on which the agent always has the same competence, with no progress. The red curve is the learning curve on an unreachable goal, e.g. moving an uncontrollable object. 
Right: exploration preference of an agent with a learning progress heuristic (competence derivative) to explore the $5$ goals defined by the learning curves. 
At the beginning of exploration, the agent makes the most progress on goal blue so it prefers to train on this one, and then its preference will shift towards goals orange and green. The agent is making no progress on goal purple so will not choose to explore it, and goal red has a noisy but low estimated learning progress.
}
\label{curves}
\end{figure}

In the experiments of this paper, we use intrinsic rewards based on measuring the competence progress towards the self-generated goals, which has been shown to be particularly efficient for learning repertoires of high-dimensional robotics skills \citep{baranes_active_2013}. 
Figure \ref{curves} shows a schematic representation of possible learning curves and the exploration preference of an agent with intrinsic rewards based on learning progress.

\section{Modular Population-Based IMGEP Architecture}

\begin{figure*}[ht]
\centering
\includegraphics[width=0.9\textwidth]{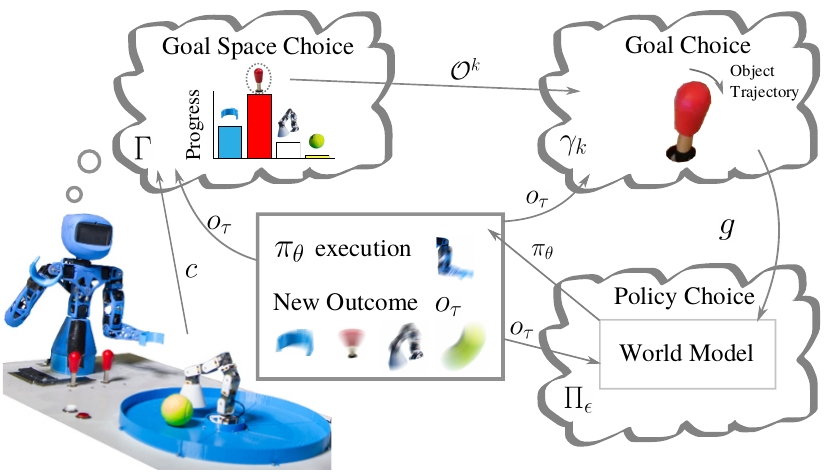}
\caption{Summary of AMB, our modular population-based IMGEP implementation.
At each iteration, the agent observes the current context $c$ and chooses a goal space (an object) to explore based on intrinsic rewards (the learning progress to move each object) with $\Gamma$.
Then a particular goal $g$ for the chosen object is sampled with $\gamma_k$, for instance to push the left joystick to the right. 
The agent chooses the best policy parameters $\theta$ to reach this goal, with the exploration meta-policy $\Pi_\epsilon$, and using an internal model of the world implementing object-centered modularity in goals and mutations.
The agent executes policy $\pi_\theta$, observes the trajectory $\tau$ and compute the outcome $o_\tau$ encoding the movement of each object.
Finally, each component is updated with the result of this experiment.
}
\label{main_drawing}
\end{figure*}

In the previous section, we defined the most general IMGEP architecture without specifying the implementation of its components such as goals and policies.
We define here a particular IMGEP architecture, used in the experiments of this paper, where the goal-parameterized policy $\Pi$ is based on a population of solutions, the goal space $\mathcal{G}$ is constructed in a modular manner from a set of objects and the exploration mutations is temporally modular through taking into account the movement of those objects.
This particular architecture is called Modular Population-Based IMGEP, and we detail its ingredients in the following sections.
Its pseudo-code is provided in Architecture \ref{algo:IMMGEP}. Figure \ref{main_drawing} summarizes the different components of Active Model Babbling (AMB), our implementation of the Modular Population-Based IMGEP architecture, using learning progress for goal sampling and stepping-stone preserving mutations.


\begin{architecture}[t]
	\caption{~ Modular Population-Based IMGEP}
	\label{algo:IMMGEP}
	\begin{algorithmic}[1]
	
	\Require Action space $\mathcal{A}$, State space $\mathcal{S}$, Context space $\mathcal{C}$, Outcome space $\mathcal{O}$
	
	\State Initialize knowledge $\mathcal{E}=\emptyset$
	\State Initialize goal space $\mathcal{G}$, goal policies $\gamma_k$ and goal space policy $\Gamma$
	\State Initialize meta-policies $\Pi$ and $\Pi_\epsilon$
	
	\State Launch asynchronously the two following loops:
	\Loop \Comment Exploration loop 
	\State Observe context $c$
	\State Choose goal space $\mathcal{G}^k$ with $\Gamma$
	\State Choose goal $g$ in $\mathcal{G}^k$ with $\gamma_k$
	\State Choose policy parameters $\theta$ to explore $g$ in context $c$ with $\Pi_\epsilon$
	\State Execute a roll-out of $\pi_\theta$, observe trajectory $\tau$
	\State Compute outcome $o_{\tau}$ from trajectory $\tau$
	\LineCommentCont{From now on, $f_{g'}(\tau)$ can be computed to estimate the fitness of the experiment $\tau$ for achieving any goal $g' \in \mathcal{G}$}
	\State Compute the fitness $f = f_g(\tau)$ associated to goal $g$
	\State Compute intrinsic reward $r_i = IR(\mathcal{E}, c, g, \theta, o_{\tau}, f)$ associated to $g$ in context $c$
	\State Update exploration meta-policy $\Pi_\epsilon$ with ($\mathcal{E}, c, \theta, \tau, o_{\tau}$) \Comment e.g. fast incr. algo.
	\State Update goal policy $\gamma_k$ with ($\mathcal{E}, c, g, o_{\tau}, f, r_i$)
	\State Update goal space policy $\Gamma$ with ($\mathcal{E}, c, k, g, o_{\tau}, f, r_i$)
	\State Update knowledge $\mathcal{E}$ with ($c, g, \theta, \tau, o_{\tau}, f, r_i$)
	\EndLoop
	
	\Loop \Comment Exploitation loop
	\State Update meta-policy $\Pi$ with $\mathcal{E}$ \Comment e.g. batch training of DNN, SVM, GMM
	\EndLoop
	\State \Return $\Pi$
	
	\end{algorithmic}
\end{architecture}

\subsection{Population-Based Meta-Policies \texorpdfstring{$\Pi$}{Pi} and \texorpdfstring{$\Pi_\epsilon$}{Pi\_epsilon}}
\label{policies}

In this version of the IMGEP framework, the goal-parameterized policy $\Pi$ is based on a population of low-level policies.
We consider that the starting state $s_{t_0}$ of a trajectory is characterized by a feature vector $c$ called \textbf{context}.
The policy $\Pi$ is built from a set of low-level policies $\pi_\theta$ parameterized by $\theta \in \Theta$, and from a meta-policy $\Pi(\bm{\theta}~|~\bm{g},\bm{c})$ which, given a goal and context, chooses the best policy $\pi_\theta$ to achieve the goal $g$. 
The policies $\pi_\theta(\bm{a_{t+1}}~|~\bm{s_{t_0:t+1}}, \bm{a_{t_0:t}})$ can be implemented for instance by stochastic black-box generators or small neural networks (see experimental section).

During the goal exploration loop, the main objective consists in collecting data that covers well the space of goals: finding $\theta$ parameters that yield good solutions to as many goals as possible. 
The \textbf{exploration meta-policy} $\Pi_\epsilon(\bm{\theta}~|~\bm{g}, \bm{c})$ is learned and used to output a distribution of policies $\pi_\theta$ that are interesting to execute to gather information for solving in context $c$ the self-generated goal $g$ and goals similar to $g$.
To achieve the objective of collecting interesting data, the exploration meta-policy $\Pi_\epsilon$ must have fast and incremental updates. 
As the aim is to maximize the coverage of the space of goals, being very precise when targeting goals is less crucial than the capacity to update the meta-policy quickly and incrementally. 
In our experiments, the exploration meta-policy $\Pi_\epsilon(\bm{\theta}~|~\textbf{g}, \textbf{c})$ is implemented as a fast memory-based nearest neighbor search with a kd-tree \citep{kdtree}.
\textcolor{black}{More precisely, we record for each policy $\pi_{\theta}$ in our database the associated context $c$ it was used in and the outcome $o_{\tau}$ it produced. Then, given a new context $c^{new}$ and a goal $g$ to attain, which correspond to an (object-specific) outcome $o$ to produce, $\Pi_\epsilon$ selects for interaction the policy $\pi_{\theta'}$ whose associated context and (object-specific) outcome is closest to $c^{new}$ and $g$ (using a nearest neighbor search in the concatenated context-outcome space). Note that $c$ is not used for goal selection, such that a single learning progress-based motivation level can be inferred per goal space.}

On the contrary, the purpose of the \textbf{target meta-policy} $\Pi$ is to be used in exploitation mode: later on, it can be asked to solve as precisely as possible some goals $g$ with maximum fitness. 
As the training of this meta-policy can be done asynchronously from data collected by the goal exploration loop, this allows the use of slower training algorithms, possibly batch, that might generalize better, e.g. using Gaussian mixture models, support vector regression or (deep) neural networks. These differences justify the fact that IMGEP uses in general two different representations and learning algorithms for $\Pi_\epsilon$ and $\Pi$. 
This two-level learning scheme has similarities with the Complementary Learning Systems Theory used to account for the organization of learning in mammalian brains \citep{kumaran2016learning}.
\textcolor{black}{To keep our experiments tractable and focus on studying intrinsically motivated exploration, we consider a simple target meta-policy $\Pi$, corresponding to our exploration meta-policy without mutations mechanisms (section \ref{temporal}).}

\subsection{Object-Centered Modular Goal Construction}
\label{goal_construction}

In the IMGEP architecture, the agent builds and samples goals autonomously. 
Here, we consider the particular case where the agent builds several goal spaces that correspond to moving each object in the environment. 
Each goal space represents features of the movement of an object in the environment, such as its end position in $\tau$ or its full trajectory.

We define the \textbf{outcome} $o_\tau \in \mathcal{O}$ of an experiment $\tau$ as the features of the movement of all objects, so that $\mathcal{O} = \prod\limits_{k} \mathcal{O}^k$ where $o_\tau^k \in \mathcal{O}^k$ are the features of object $k$.
Those features come from a perceptual system that may be given to the agent or learned by the agent.
From feature space $\mathcal{O}^k$, the agent can autonomously generate a corresponding goal space $\mathcal{G}^k$ that contains fitness functions of the form $f_g(\tau) = - ||g - o_\tau^k||_k$.
The norm $||.||_k$ is a distance in the space $O^k$, which can be normalized to be able to compare the fitness of goals across goal spaces. 
The goal space is thus modular, composed of several object-related subspaces: $\mathcal{G}=\bigcup\limits_{k} \mathcal{G}^k$.

With this setting, goal sampling is hierarchical in the sense that the agent first chooses a goal space $\mathcal{G}^k$ to explore with a goal space policy $\Gamma$ and then a particular goal $g \in \mathcal{G}^k$ with the corresponding goal policy $\gamma_k$. Those two levels of choice can make use of self-computed intrinsic rewards $r_i$ (see Sec. \ref{intrinsic_rewards}).

Given an outcome $o_{\tau}$, the fitness $f_g(\tau)$ can thus be computed by the agent for all goals $g \in \mathcal{G}$ and at any time after the experiment $\tau$. 
For instance, if while exploring the goal of moving object A to the left, object B moved to the right, that outcome can be taken into account later when the goal is to move object B.

\subsection{Object-Centered Temporal Modularity: Stepping-Stone Preserving Mutations}
\label{temporal}

When targeting a new goal $g$, the internal model (a memory-based nearest neighbor search in our experiments) infers the best policy $\pi_\theta$ to reach the goal $g$.
The exploration meta-policy $\Pi_\epsilon$ then performs a mutation on $\theta$ in order to explore new policies.
\textcolor{black}{This mutation step can be seen as analogous to existing approaches injecting noise in parametric RL policies to foster exploration \citep{fortunato2018noisy,plappert2018noise}.}
The mutation operator could just add a random noise on the parameters $\theta$, however, those parameters do not all have the same influence on the execution of the policy, in particular with respect to time.
In our implementations, the parameters are sequenced in time, with some parameters influencing more the beginning of the policy roll-out and some more the end of the trajectory.
However, in the context of tool use, the reaching or grasping of a tool is necessary for executing a subsequent action on an object.
A random mutation of policy parameters, irrespective of the moment when the tool is grasped, can result in an action where the agent do not grasp the tool and thus cannot explore the corresponding object.

The Stepping-Stone Preserving Mutation operator (\texttt{SSPMutation}) analyzes the trajectory of the target object while the previous motor policy $\pi_\theta$ was run, to find the moment when the object started to move. 
The operator does not change the variables of $\theta$ concerning the movement before the object moved and mutates the variables of $\theta$ concerning the movement after the object moved (see an example mutation in Fig. \ref{example_traj}). 
When the goal of the agent is to move the tool and it already succeeded to move the tool in the past with policy $\pi_\theta$, then the application of this mutation operator changes the behavior of the agent only when the tool start to move, which makes the agent explore with the tool once grasped and avoid missing the tool.
Similarly, when the goal of the agent is to move a toy controlled by a tool, the mutation changes the behavior only when the toy starts to move, which makes the agent grasp the tool and reach the toy before exploring new actions, so that the agent do not miss the tool nor the toy.
The idea of this stepping-stone preserving operator is similar to the Go-Explore approach \citep{ecoffet2021first}.

\subsection{Active Model Babbling (AMB) and Random Model Babbling (RMB)}

The Modular Population-Based IMGEP architecture gives a high-level description of the learning agent with a population-based policy and an object-centered modularity in goals and mutations. Each component of this architecture may be instantiated in various ways. For instance, in the main loop of Architecture \ref{algo:IMMGEP}, many aspects are not constrained such as how the goal is chosen, how the parameters $\theta$ are computed, how the policies $\pi_\theta$ are implemented, how the intrinsic rewards are defined.

An important contribution of the present work is to design and showcase the effectiveness of one particular instantiation of modular population-based IMGEPs, called Active Model Babbling (AMB, see fig. \ref{main_drawing}), which uses a learning-progress based mechanism for sampling goal spaces (see section \ref{lp-sampling-imple} for examples of implementations) as well as the stepping-stone preserving mutations mechanism described in section \ref{temporal} (see section \ref{sspm-imple} for implementation details). 

In addition to AMB, we also present an alternative variant, called Random Model Babbling (RMB), which also uses the stepping-stone preserving mutations, but implements a simpler goal space policy: instead of using learning progress estimates, goal spaces are selected randomly throughout training. Although apparently simplistic, this approach can perform surprisingly well, especially when all goal spaces are relevant, i.e. when there are no distractors (see figure \ref{explo_distractors}).

In Section \ref{exploalgo}, we detail implementations of these types of Modular Population-Based IMGEP architectures, as well as other baselines.

\section{Experiments}

In this section, we evaluate the Modular Population-Based IMGEP architecture by designing several algorithmic implementations and several environments suitable for curriculum learning, where the exploration of a task brings information to later solve other tasks. 
In particular, we study environments where agents discover objects that can be used as tools to move other objects. 
A good exploration of a tool and of its functioning will yield a better exploration of the objects on which this tool can act.
Those tasks provide the opportunity for an intrinsically motivated agent to build on the skills it has learned to explore and learn new skills on its own. 

Here, we first describe three tool-use learning environments and we detail our implementations of IMGEP and of several control conditions.
Then, we study the behavior of the different agents in the different environments depending on the learning architecture.
We investigate in particular the benefits of a modular representation of the sensory feedback with goals based on objects, and how the exploration mutations can take into account the movement of the goal object.
We further examine how and in which conditions the intrinsic motivation component of the IMGEP architecture improves the learning of skills that can be reused, such as using a tool to move an object.

\subsection{Tool-Use Environments}

We design three tool-use environments. 
The first one is a 2D simulated robotic arm with 3 joints and a gripper that can grab sticks and move toys.
It is a simple environment with no physics and only 2D geometric shapes so very fast to execute. 
The second environment is a Minecraft scene where an agent is able to move, grab and use tools such as a pickaxe to break blocks. 
The third one is a real robotic setup with a Torso robot moving its arm that can reach joysticks controlling a toy robot. 
This setup has complex high-dimensional motor and sensory spaces with noise both in the robot physical arm and in the interaction between objects such as its hand and the joysticks.
It is a high-dimensional and noisy environment with a similar stepping-stone structure as the robotic environments but with a completely different sensorimotor setup.
The code of the different environments and experiments is available on GitHub\footnote{Code of the IMGEP experiments: \href{https://github.com/sebastien-forestier/IMGEP}{https://github.com/sebastien-forestier/IMGEP}}.

\subsubsection{2D Simulated Tool-Use Environment}
    
\begin{figure}[t]
	\centering
	\includegraphics[width=0.9\textwidth]{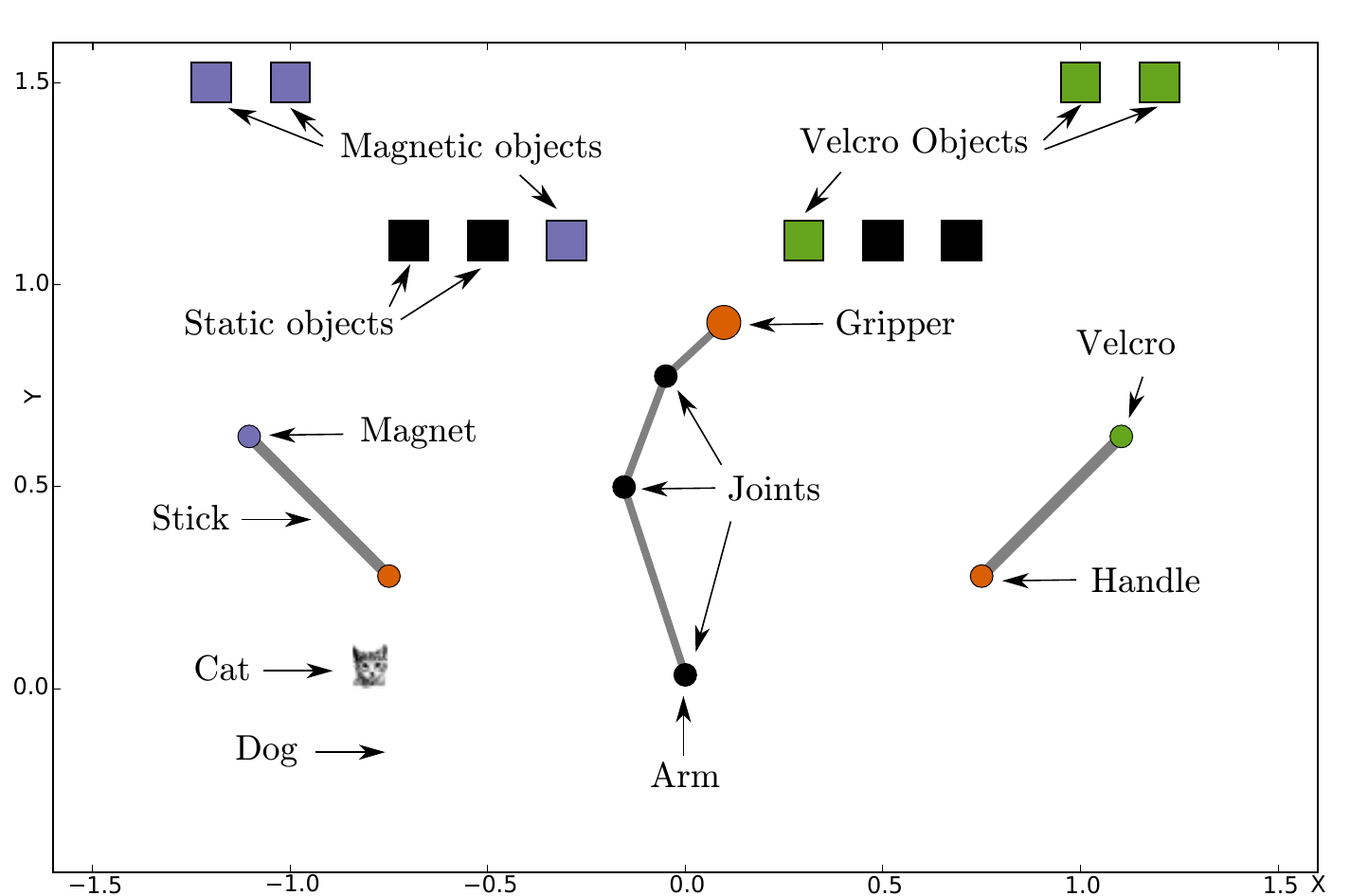}
	\caption{2D Simulated Tool-Use Environment. A simulated robotic arm with a gripper can grab sticks and move toys. The gripper has to close near the handle of a stick to grab it. One magnetic toy and one Velcro toy are reachable with their corresponding stick. Other toys cannot be moved (static or too far away). The cat and the dog are distractors: they move randomly, independently of the arm.}
	\label{2Denv}
\end{figure}

In the 2D Simulated Environment (see Fig. \ref{2Denv}), the learning agent controls a robotic arm with a gripper, that can grab one of two sticks, one with a magnet at the end and one with Velcro, that can themselves be used to move several magnets or Velcro toys. 
Some other objects cannot be moved, they are called static distractors, and finally a simulated cat and dog are randomly moving in the scene, they are random distractors.

The 2D robotic arm has $3$ joints that can rotate from $-\pi~rad$ to $\pi~rad$.
The length of the $3$ segments of the arm are $0.5$, $0.3$ and $0.2$ so the length of the arm is $1$ unit.
The starting position of the arm is vertical with joints at position $0~rad$ and its base is fixed at position $(0, 0)$.
The gripper $gr$ has 2 possible positions: \textit{open} ($gr \geq 0$) and \textit{closed} ($gr < 0$).
The robotic arm has 4 degrees of freedom represented by a vector in $[-1,1]^4$.

Two sticks of length $0.5$ can be grasped by the handle side (orange side) in order to catch an out-of-reach object.
The magnetic stick can catch magnetic objects (in blue), and the other stick has a Velcro tape to catch Velcro objects (in green).
If the gripper closes near the handle of one stick, this stick is considered grasped and follows the gripper's position and the orientation of the arm's last segment until the gripper opens.	
If the other side of a stick reaches a matching object (magnetic or Velcro), the object then follows the stick.
There are three magnetic objects and three Velcro objects, but only one of each type is reachable with its stick.
A simulated cat and dog are following a random walk, they have no interaction with the arm nor with other object.
Finally, four static black squares have also no interaction with other objects.
The arm, tools and other objects are reset to their initial state at the end of each iteration of $50$ steps.

The agent receives a sensory feedback representing the result of its actions.
This feedback (or outcome) is either composed of the position of each object at $5$ time points during the $50$ steps trajectory, or just the end state of each object, depending on the experiments.
First, the hand is represented by its $X$ and $Y$ position and the aperture of the gripper ($1$ or $-1$).
The sticks are represented by the $X$ and $Y$ position of their tip.
Similarly, each other object is represented by their $X$ and $Y$ positions.
Each of the $15$ objects defines a sensory space $S_i$ \textcolor{black}{($10$ of those objects are uncontrollable distractors)}.
The total sensory space $S$ has either $155$ dimensions if trajectories are represented, or $31$ dimensions if only the end state of each object is represented.

\subsubsection{Minecraft Mountain Cart}

\begin{figure}[t]
	\centering
	\includegraphics[width=0.9\textwidth]{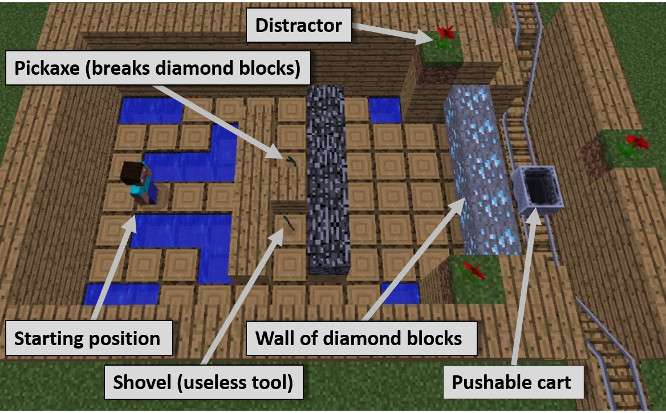}
	\caption{Minecraft Mountain Cart Environment. If the agent manages to avoid falling into water holes it may retrieve and use a pickaxe to break diamond blocks and access the cart. A shovel is also located in the arena and serves as a controllable distractor.}
	\label{MMCenv}
\end{figure}

The Minecraft Mountain Cart (MMC) extends the famous Mountain Car control benchmark in a 3D environment with a multi-goal setting (see Fig. \ref{MMCenv}).

In this episodic task, the agent starts on the left of the rectangular arena and is given ten seconds ($40$ steps) to act on the environment using 2 continuous commands: \textit{move} and \textit{strafe}, both using values in $[-1;1]$. \textit{move(1)} moves the agent forward at full speed,  \textit{move(-0.1)} moves the agent slowly backward, etc. Similarly \textit{strafe(1)} moves the agent left at full speed and \textit{strafe(-0.1)} moves it slowly to the right. Additionally, a third binary action allows the agent to use the currently handled tool.

The first challenge of this environment is to learn how to navigate within the arena's boundaries without falling in water holes (from which the agent cannot get out). Proper navigation might lead the agent to discover one of the two tools of the environment: a shovel and a pickaxe. The former is of no use but the latter enables to break diamond blocks located further ahead in the arena. A last possible interaction is for the agent to get close enough to the cart to move it along its railroad. If given enough speed, the cart is able to climb the left or right slope. The height and width of these slopes were made in such a way that an agent simply hitting the cart at full speed will not provide enough inertia for the cart to climb the slope. The agent must at least partially support the cart along the track to propel it fast enough to fully climb the slope.

The outcome of an episode is a vector composed of the end position of the agent ($2$D), shovel ($2$D), pickaxe ($2$D), cart ($1$D) and $3$ distractors ($2$D each) positions along with a binary vector ($5$D) encoding the $5$ diamond blocks' states \textcolor{black}{($3$ objects out of $8$ are uncontrollable distractors)}.

This environment is interesting to study modular IMGEP approaches since it is composed of a set of linked tasks of increasing complexity. Exploring how to navigate will help to discover the tools and, eventually, will allow to break blocks and move the cart.

\subsubsection{Robotic Tool-Use Environment}

\begin{figure*}[t]
\centering
\includegraphics[width=6.7cm]{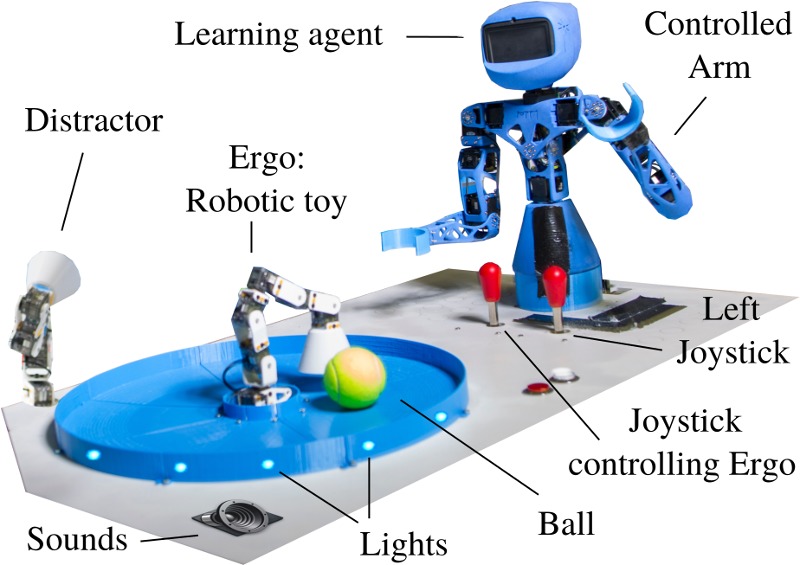}\hspace{0.4cm}
\includegraphics[width=6.7cm]{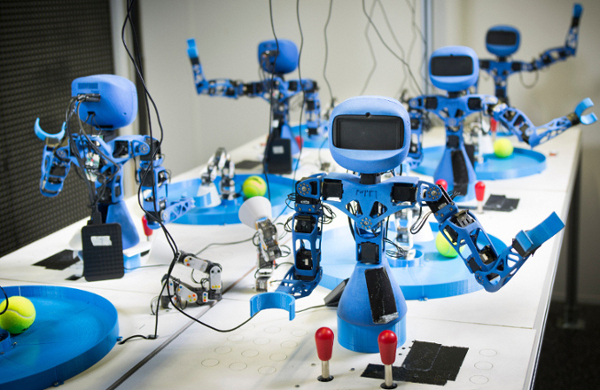}
\vspace{0.2cm}
\caption{Robotic Tool-Use Environment. Left: a Poppy Torso robot (the learning agent) is mounted in front of two joysticks that can be used as tools to act on other objects: a Poppy Ergo robotic toy and a ball that can produce light and sound. Right: 6 copies of this setup are running in parallel to gather more data. Several Ergo robots are placed between robots: they act as distractors that move randomly, independently of the agents.}
\label{TorsoEnv}
\end{figure*}

In order to benchmark different learning algorithms in a realistic robotic environment with high-dimensional action and outcome spaces, we designed a real robotic setup composed of a humanoid arm in front of joysticks that can be used as tools to act on other objects (see Fig. \ref{TorsoEnv}). 
We recorded a video of an early version of the experimental setup\footnote{Early version of the experimental setup: \href{https://youtu.be/NOLAwD4ZTW0}{https://youtu.be/NOLAwD4ZTW0}}.

A Poppy Torso robot (the learning agent) is mounted in front of two joysticks and explores with its left arm. 
A Poppy Ergo robot (seen as a robotic toy) is controlled by the right joystick and can push a ball that controls some lights and sounds.
Poppy is a robust and accessible open-source 3D printed robotic platform \citep{poppy}.

The left arm has $4$ joints, with a hook at the tip of the arm.
A trajectory of the arm is here generated by radial basis functions with $5$ parameters on each of the $4$ degrees of freedom ($20$ parameters in total).

Two analogical joysticks (Ultrastick 360) can be reached by the left arm and moved in any direction.
The right joystick controls the Poppy Ergo robotic toy, and the left joystick do not control any object.
The Poppy Ergo robot has 6 motors, and moves with hardwired synergies that allow control of rotational speed and radial extension.

A tennis ball is freely moving in the blue arena which is slightly sloped so that the ball comes close to the center at the end of a movement.
The speed of the ball controls (above a threshold) the intensity of the light of a LED circle around the arena.
Finally, when the ball touches the border of the arena, a sound is produced and varied in pitch depending on ball position.

Several other objects are included in the environment, with which the agent cannot interact. 
Two Ergo robots ($2$D objects) are moving randomly, independently of the agent.
Six objects are static: the right hand ($3$D) of the robot that is disabled in this experiment, the camera recording the ball trajectory ($3$D), the blue circular arena ($2$D), an out-of-reach yellow toy ($2$D), the red button also out-of-reach ($2$D) and the lamp ($2$D). All distractor objects are reset after each roll-out. 

The context $c$ of this environment represents the current configuration of objects in the scene.
In practice, since only the Ergo and ball are not reset after each roll-out, this amounts to measuring the rotation angle of the Ergo and of the ball around the center of the arena.

The agent is given a perceptual system providing sensory feedback that represents the trajectories of all objects in the scene.
First, the $3$D trajectory of the hand is computed through a forward model of the arm as its $x$, $y$ and $z$ position.
The $2$D states of each joystick and of the Ergo are read by sensors, and the position of the ball retrieved through the camera.
The states of the $1$D intensity of the light and the $1$D pitch of the sound are computed from the ball position and speed.
Each of the $15$ objects defines a sensory space $S_i$ representing its trajectory \textcolor{black}{($8$ of those objects are uncontrollable distractors)}.
The total sensory space $S$ has $310$ dimensions.

\subsection{Implementation of the Modular Population-Based IMGEP Architecture}
\label{exploalgo}

In the following subsections, we detail our implementations of the algorithmic parts of the modular population-based IMGEP architecture (see architecture \ref{algo:IMMGEP}).

\subsubsection{Motor Policy \texorpdfstring{$\pi_{\theta}$}{pi\_theta}}
\label{pitheta}

\begin{figure}[ht]
\centering
\subfloat[Basis Functions shape]{\includegraphics[width=0.475\textwidth]{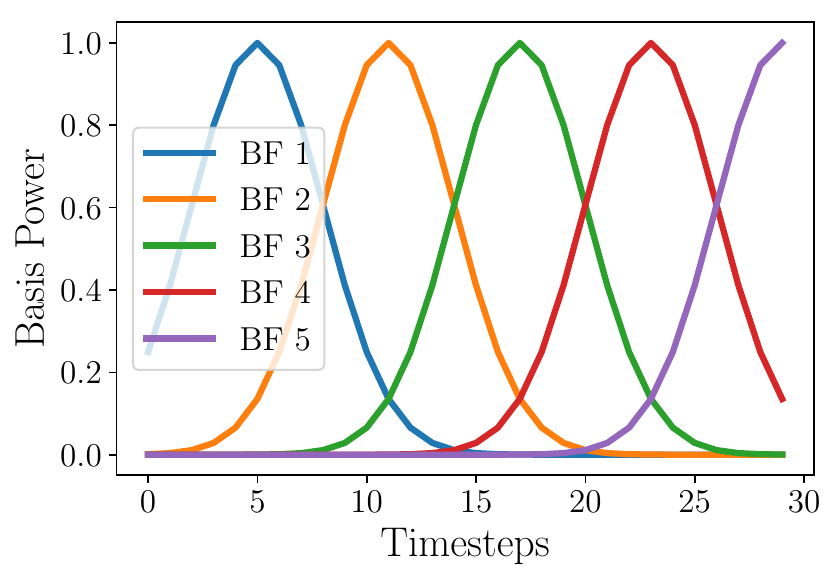}}\hspace{0.2cm}
\subfloat[Example Joints Trajectory]{\includegraphics[width=0.495\textwidth]{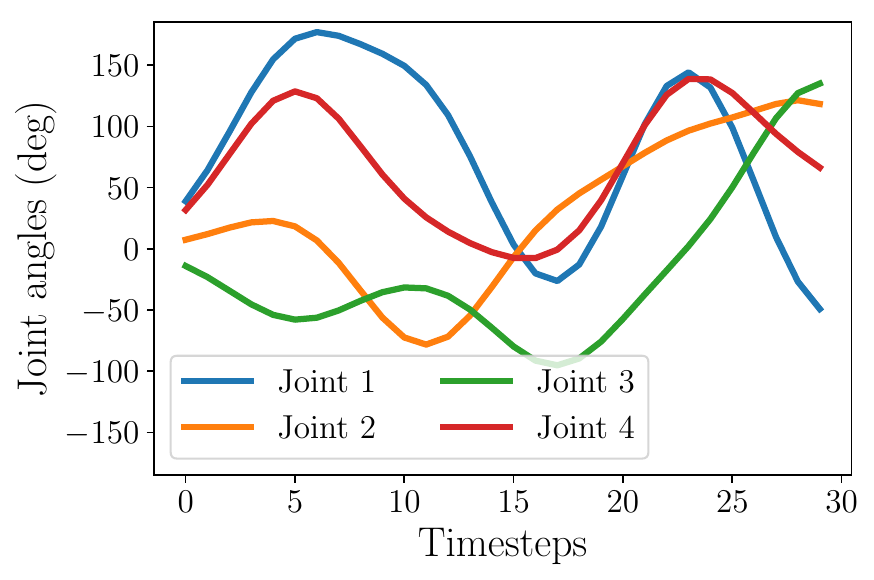}}
\caption{Implementation of motor policies $\pi_{\theta}$ through Radial Basis Functions.
(a) 5 Gaussian bases with different centers but same shape.
(b) the movement of each joint is the result of a weighted sum of the product of $5$ parameters and the $5$ basis.
The total vector $\theta$ has $20$ parameters, in both the 2D Simulated and the Robotic environment. 
}
\label{bf}
\end{figure}

In the 2D Simulated environment and the Robotic environment, we implement the motor policies with Radial Basis Functions (RBF). 
We define $5$ Gaussian basis functions with the same shape ($\sigma=5$ for a $50$ steps trajectory in the 2D environment and $\sigma=3$ for $30$ steps in the Robotic environment) and with equally spaced centers (see
Fig. \ref{bf}).
The movement of each joint is the result of a weighted sum of the product of $5$ parameters and the $5$ basis.
The total vector $\theta$ has $20$ parameters, in both the 2D Simulated and the Robotic environment. In the 2D environment, the fourth joint is a gripper that is considered open if its angle is positive and closed otherwise.

In the Minecraft Mountain Cart environment, trajectories are sampled in a closed-loop fashion using neural networks. 
The observation vector has the same structure as the outcome vector: it provides the current positions of all objects normalized in $[-1;1]$ ($18$D). 
Each neural network is composed of one hidden layer of $64$ Relu units and a $3$D output with tanh activation functions.
The 1411 policy parameters are initialized using the initialization scheme of \citet{he2015}.
\textcolor{black}{Note that in our experiments, neural networks are not trained through backpropagation: as for RBF policies, their parameters are \textit{mutated}. Our mutation-based approach is close to neuroevolution approaches \citep{stanley2019designing}.}

\subsubsection{Stepping-Stone Preserving Mutations}
\label{sspm-imple}
The Stepping-Stone Preserving Mutation operator (\texttt{SSPMutation}) does not change the variables of $\theta$ concerning the movement before the object moved and modifies the variables of $\theta$ concerning the movement after the object moved.
\texttt{SSPMutation} adds a Gaussian noise around those values of $\theta$ in the 2D simulated environment ($\sigma=0.05$) and in Minecraft Mountain Cart ($\sigma=0.3$), or adds the Gaussian noise around the previous motor positions (in the robotic environment with joysticks).
In the experimental section we compare it to the \texttt{FullMutation} operator that adds a Gaussian noise to $\theta$ irrespective of the moment when the target object moved.

\subsubsection{Goal Space Policy \texorpdfstring{$\Gamma$}{Gamma}}
\label{lp-sampling-imple}

The agent estimates its learning progress globally in each goal space (or for each model learned).
At each iteration, the context $c$ is observed, a goal space $k$ is chosen by $\Gamma$ and a random goal $g$ is sampled by $\gamma_k$ in $\mathcal{G}^k$ (corresponding to a fitness function $f_g$).
Then, in $80\%$ of the iterations, the agent uses $\Pi_\epsilon(\bm{\theta}~|~g, c)$ to generate with exploration a policy $\theta$ and does not update its progress estimation.
In the other $20\%$, it uses $\Pi$, without exploration, to generate $\theta$ and updates its learning progress estimation in $\mathcal{G}^k$, with the estimated progress in reaching $g$.			
To estimate the learning progress $r_i$ made to reach the current goal $g$, the agent compares the outcome $o_\tau$ with the outcome $o_\tau'$ obtained for the previous context and goal ($g'$, $c'$) most similar (Euclidean distance) to ($g$, $c$): $~r_i = f_g(\tau) - f_g(\tau')$.
Finally, $\Gamma$ implements a non-stationary bandit algorithm to sample goal spaces. The bandit keeps track of a running average $r_i^k$ of the intrinsic rewards $r_i$ associated to the current goal space $\mathcal{G}^k$. 
With probability $20\%$, it samples a random space $\mathcal{G}^k$, and with probability $80\%$, the probability to sample $\mathcal{G}^k$ is proportional to $r_i^k$ in the 2D Simulated and Minecraft environments, or $exp\big(\frac{r_i^k}{\sum_k r_i^k}\big)$ if $r_i^k > 0$ and $0$ otherwise, in the Robotic environment.

\subsubsection{Control Conditions}

We design several control conditions.
In the Random Model Babbling (RMB) condition, the choice of goal space is random: $\Gamma(\textbf{k}~|~\textbf{c})$, and $\gamma_k(\textbf{g}~|~\textbf{c})$ for each $k$ are always uniform distributions.
Agents in the Single Goal Space (SGS) condition always choose the same goal space, of high interest to the engineer: the magnet toy in the 2D Simulated environment, and the ball in the robotic environment. 
The Fixed Curriculum (FC) condition defines $\Gamma$ as a curriculum sequence engineered by hand: the agents explore objects in a sequence from the easiest to discover to the most complex object while ignoring distractors.
The conditions SGS and FC are thus extrinsically motivated controls. 
We define the Flat Random Goal Babbling (FRGB) condition with a single outcome/goal space containing all the variables of all objects, to compare modular and non-modular representations of the environment. The agents in this condition choose random goals in this space, and use the \texttt{FullMutation} operator.
Finally, agents in the Random condition always choose random motor policies $\theta$.

\subsection{Results}

In this section we show the results of several experiments with the three environments and the different learning conditions. 
We first study in details the Active Model Babbling (AMB) learning algorithm, a modular implementation of the IMGEP architecture.
Then, in order to understand the contribution of the different components of this learning algorithm, we compare it to several controls (or ablations): without a modular representation of goals, without the goal sampling based on learning progress, or without the stepping-stone preserving mutation operator.
In those experiments, goals are sampled in spaces representing the sensory feedback from the environment.
We thus compare several possible encodings of the feedback: with the trajectory of each object or with only the end point of the trajectories.
We included distractors that cannot be controlled by the learning agent in the three tool-use environments. 
We also test the learning conditions with and without distractors to evaluate their robustness to distractors.

\subsubsection{Exploration Measure and Summary Results}

Table \ref{table_explo} shows a summary of the exploration results at the end of the runs, in all conditions in all spaces of all environments, 
We give the $25$, $50$ and $75$ percentiles of the exploration results of all seeds. 
Exploration measures the percentage of reached cells in a discretization of each goal space.
The best condition in each space is highlighted in bold, based on Welch's t-tests (with threshold $p<0.05$): if several conditions are not significantly different, they are all highlighted.
In the $2$D Simulated environment, there are $100$ seeds for each condition, and the exploration measures the number of cells reached in a discretization of the $2$D space of the end position of each object with $100$ bins on each dimension.
In the Minecraft environment, there are $20$ runs for conditions Random, SGS, FRGB, FC and $42$ for AMB and RMB. The exploration metric for the agent, pickaxe and shovel spaces is the number of reached cells in a discretization of the $2$D space in $450$ bins ($15$ on the x axis, $30$ on the y axis). The same measure is used for the block space, which is discrete with $32$ possible combinations. For the cart space we measure exploration as the number of different outcomes reached.
In the Robotic environment, there are $6$ runs with different seeds for condition SGS, $8$ for FRGB, $16$ for RMB, $23$ for AMB, $12$ for FC and 6 for Random, and the exploration also measures the number of cells reached in a discretization of the space of the end position of each object with $1000$ bins in $1$D, $100$ bins on each dimension in $2$D, and $20$ bins in $3$D.

\begin{table}[H]
\setlength\tabcolsep{3.2pt}
\begin{tabular}{|>{\small}c|>{\small}c|>{\small}c|>{\small}c|>{\small}c|>{\small}c|>{\small}c|>{\small}c|}
\hline
\multicolumn{2}{|c|}{Env, Space \textbackslash ~ Condition} & Rdm & SGS & Flat & RMB & AMB & FC \\ \hline
\small
\multirow{2}{*}{\begin{tabular}[c]{@{}c@{}}2D Simulated\\ Environment\end{tabular}} & Magnet Tool & 0,0,0 & 0,0,0 & 8.0,11,13 & 33,36,39 & 57,\textbf{61},65 & 61,\textbf{67},70 \\ \cline{2-8} 
 & Magnet Toy & 0,0,0 & 0,0,0 & 0,0,0 & 0,0,5.0 & 0,\textbf{3.0},16 & 0,\textbf{3.0},19 \\ \hline
\multirow{5}{*}{\begin{tabular}[c]{@{}c@{}}Minecraft\\ Mountain\\ Cart\end{tabular}} & Agent Pos. & 28,29,30 & 29,29,30 & 34,36,40 & 48,50,54 & 55,58,61 & 59,\textbf{63},67 \\ \cline{2-8} 
 & Shovel & 5,5,6 & 5,6,7 & 8,11,13 & 25,27,30 & 32,34,37 & 34,\textbf{37},42  \\ \cline{2-8} 
 & Pickaxe & 6,6,7 & 6,7,8 & 11,15,19 & 33,35,39 & 41,45,48 & 43,\textbf{51},61 \\ \cline{2-8} 
 & Blocks & 3,3,3 & 3,3,3 & 3,11,19 & 69,77,84 & 73,\textbf{84},93 & 100,\textbf{100},100 \\ \cline{2-8} 
 & Cart & 0,0,0 & 0,0,0 & 0,0,1 & 5,162,409 & 56,360,886 & 386,\textbf{787},1207 \\ \hline
\multirow{7}{*}{\begin{tabular}[c]{@{}c@{}}Robotic\\ Environment\end{tabular}} & Hand & 24,24,25 & 18,19,20 & 20,21,22 & 22,24,25 & 22,23,24 & 21,22,23 \\ \cline{2-8} 
 & L. Joystick & 4.2,4.7,5.9 & 1.9,3.3,4.6 & 0.1,0.1,0.3 & 15,18,19 & 20,\textbf{22},26 & 23,\textbf{26},29 \\ \cline{2-8} 
 & R. Joystick & 0.6,0.9,1.0 & 0.3,0.4,0.5 & 0,0,0 & 10,11,13 & 16,\textbf{18},22 & 15,17,18 \\ \cline{2-8} 
 & Ergo & 0.2,0.3,0.4 & 0.1,0.1,0.2 & 0,0,0 & 1.2,\textbf{1.5},1.7 & 1.5,\textbf{1.7},1.8 & 1.7,\textbf{1.7},1.9 \\ \cline{2-8} 
 & Ball & 0,0,0.1 & 0,0,0 & 0,0,0 & 0.8,1.0,1.0 & 0.9,\textbf{1.1},1.2 & 0.9,\textbf{0.9},1.0 \\ \cline{2-8} 
 & Light & 0.1,0.1,0.1 & 0.1,0.2,0.2 & 0.1,0.1,0.1 & 0.8,1.8,3.0 & 2.0,\textbf{3.6},4.9 & 1.8,\textbf{2.2},3.7 \\ \cline{2-8} 
 & Sound & 0.1,0.1,0.1 & 0.1,0.1,0.1 & 0.1,0.1,0.1 & 0.8,1.1,2.6 & 1.7,\textbf{2.8},3.6 & 1.2,1.6,2.3 \\ \hline
\end{tabular}
\vspace{0.3cm}
\caption{Exploration results in all environments and conditions. 
}
\label{table_explo}
\end{table}

\subsubsection{Intrinsically Motivated Goal Exploration}

\begin{figure}[t]
\centering
\subfloat[2D Simulated Environment]{\includegraphics[width=0.32\textwidth]{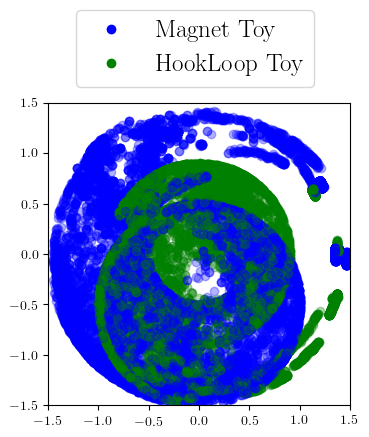}}\hspace{0.2cm}
\subfloat[Minecraft Mountain Cart]{\includegraphics[width=0.3\textwidth]{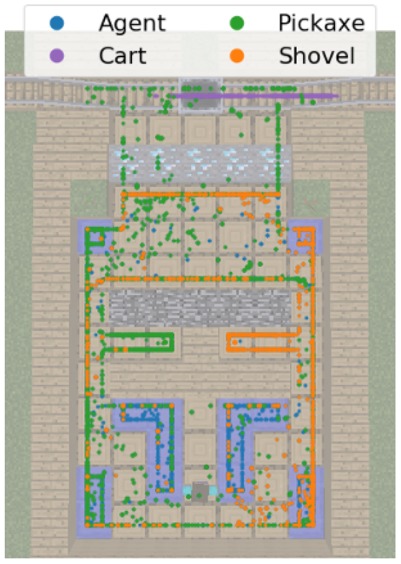}}\hspace{0.2cm}
\subfloat[Robotic Environment]{\includegraphics[width=0.32\textwidth]{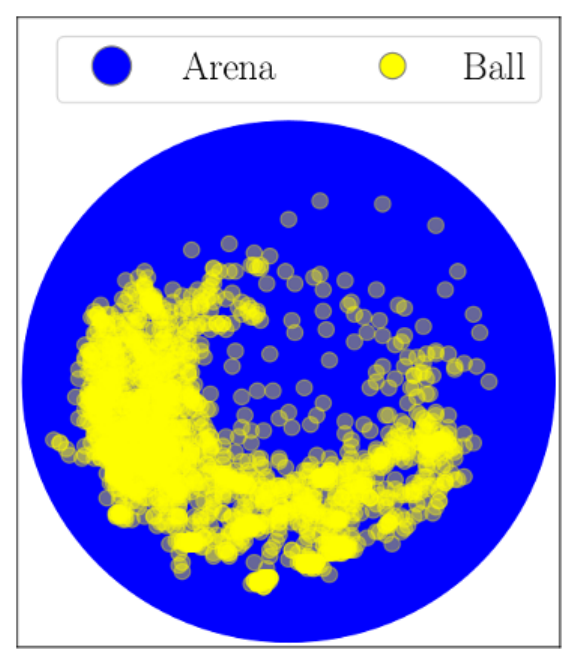}}
\caption{Examples of exploration map of one IMGEP agent in each environment. (a) in the 2D Simulated Environment, we plot the position of the reachable magnet toy at the end of each iteration with a blue point, and the Velcro toy in green.
(b) in Minecraft Mountain Cart we plot the end position of the agent, the agent with pickaxe, the agent with shovel, and the cart.
(c) in the Robotic environment, the position of the ball is plotted when it moved in the arena.
}
\label{explo_example}
\end{figure}

Here we study in detail the Active Model Babbling (AMB) learning algorithm.
AMB agents encode the sensory feedback about objects with a modular representation: each object is associated with one independent learning module. 
At each iteration, they first select an object to explore, then a particular goal to reach for this object.
They execute a motor policy to reach this goal, and observe the outcome.
The selection of the object to explore is based on a self-estimation of the learning progress made to move each object according to chosen goals.
The AMB algorithm is thus a modular implementation of the IMGEP architecture.

\paragraph{}\textit{Exploration Maps ---}~
We first plot examples of exploration results as cumulative exploration maps, one per environment.
Those maps show all the positions where one AMB agent succeeded to move objects. 

Fig. \ref{explo_example}(a) shows the position of the reachable toys of the 2D simulated environment at the end of each iteration in one trial of intrinsically motivated goal exploration. The reachable area for those two toys is the inside the circle of radius $1.5$ and center $0$. We can see that in $100$k iterations, the agent succeeded to transport the toys in many places in this area. The experiments with other seeds are very similar.
Fig. \ref{explo_example}(b) shows an exploration map of a typical run in Minecraft Mountain Cart after 40k iterations. As you can see the agent successfully managed to (1) navigate within the arena boundaries, (2) move the pickaxe and shovel, (3) use the pickaxe to break blocks and (4) move the cart located behind these blocks.  
An example in the robotic environment is shown in Fig. \ref{explo_example}(c) where we plot the position of the ball when it moved in the first $10k$ iterations of the exploration of one agent.

Overall, they show that IMGEP agents discovered how to use the different tools in each environment within the time limit: the sticks to grab the toys in the 2D simulated environment, the pickaxe to mine blocks to reach the cart in Minecraft Mountain Cart, the joysticks to move the toy and push the ball in the robotic experiment.

\begin{figure}[p]
\centering
\subfloat[Magnet Tool]{\includegraphics[width=0.45\textwidth]{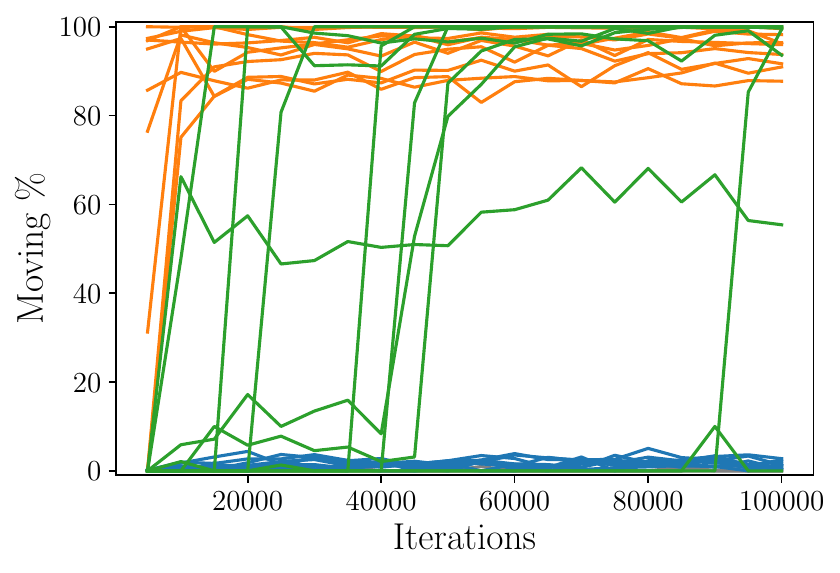}}\hspace{0.2cm}
\subfloat[Magnet Toy]{\includegraphics[width=0.45\textwidth]{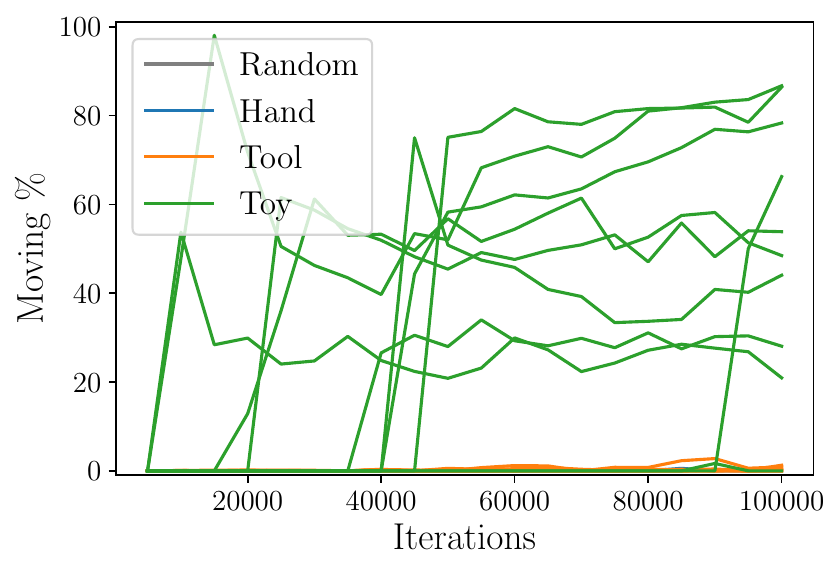}}\\

\subfloat[Pickaxe]{\includegraphics[width=0.31\textwidth]{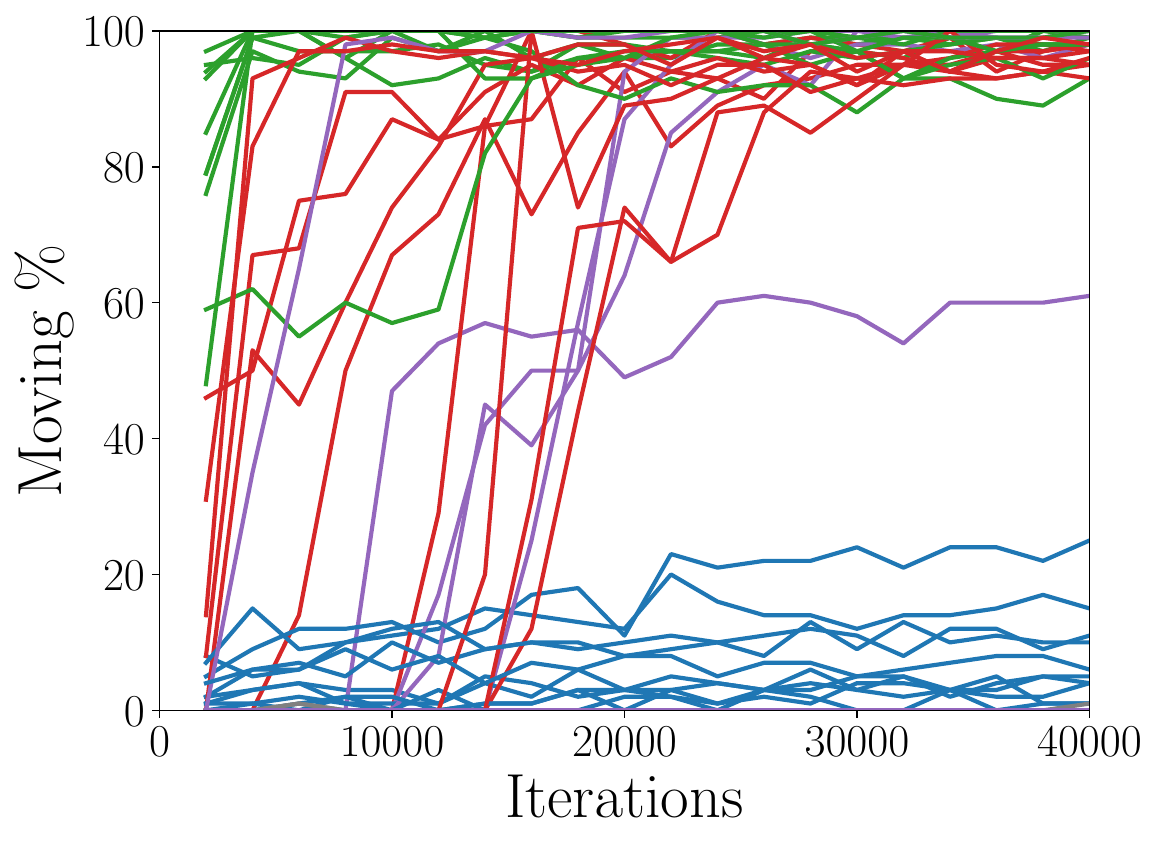}}\hspace{0.2cm}
\subfloat[Blocks]{\includegraphics[width=0.31\textwidth]{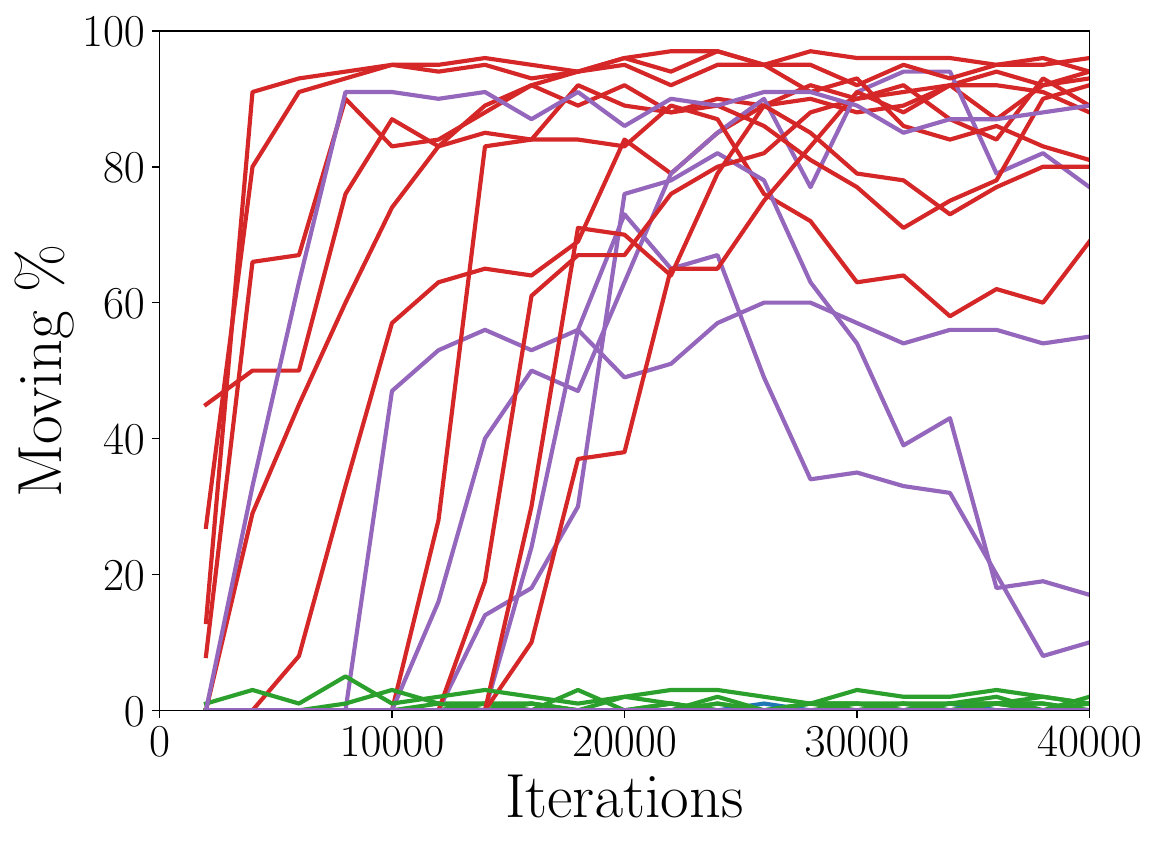}}\hspace{0.2cm}
\subfloat[Cart]{\includegraphics[width=0.31\textwidth]{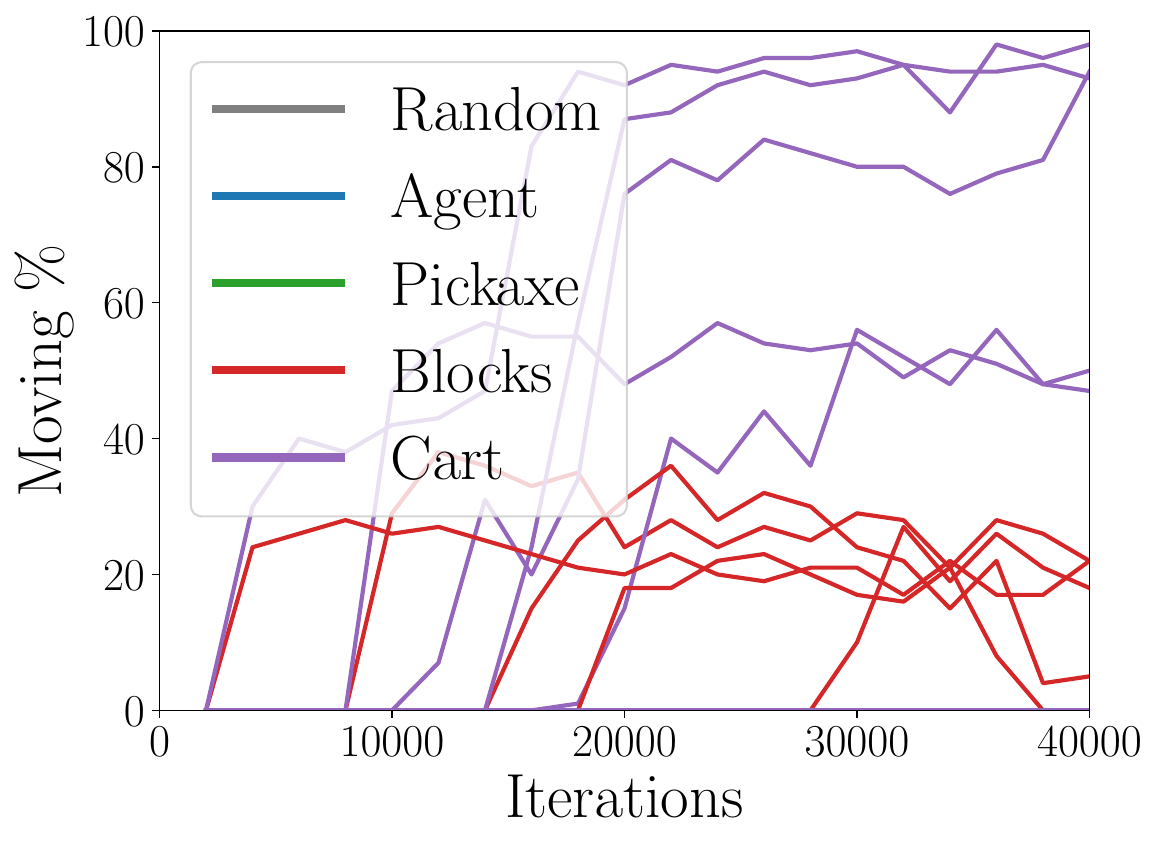}}\\

\subfloat[Left Joystick]{\includegraphics[width=0.31\textwidth]{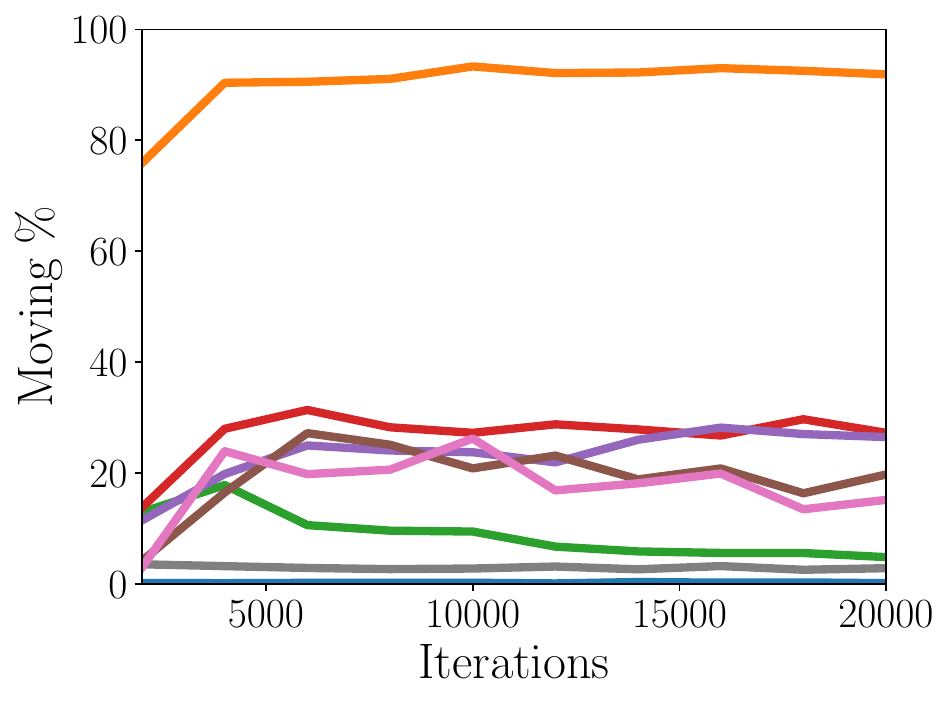}}\hspace{0.2cm}
\subfloat[Right Joystick]{\includegraphics[width=0.31\textwidth]{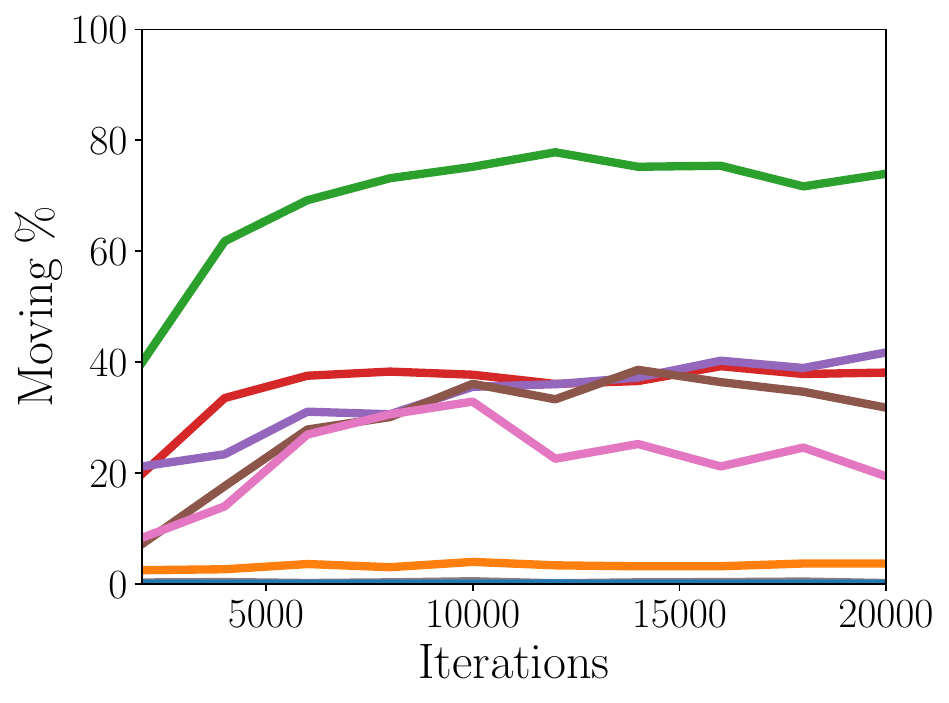}}\hspace{0.2cm}
\subfloat[Ergo]{\includegraphics[width=0.31\textwidth]{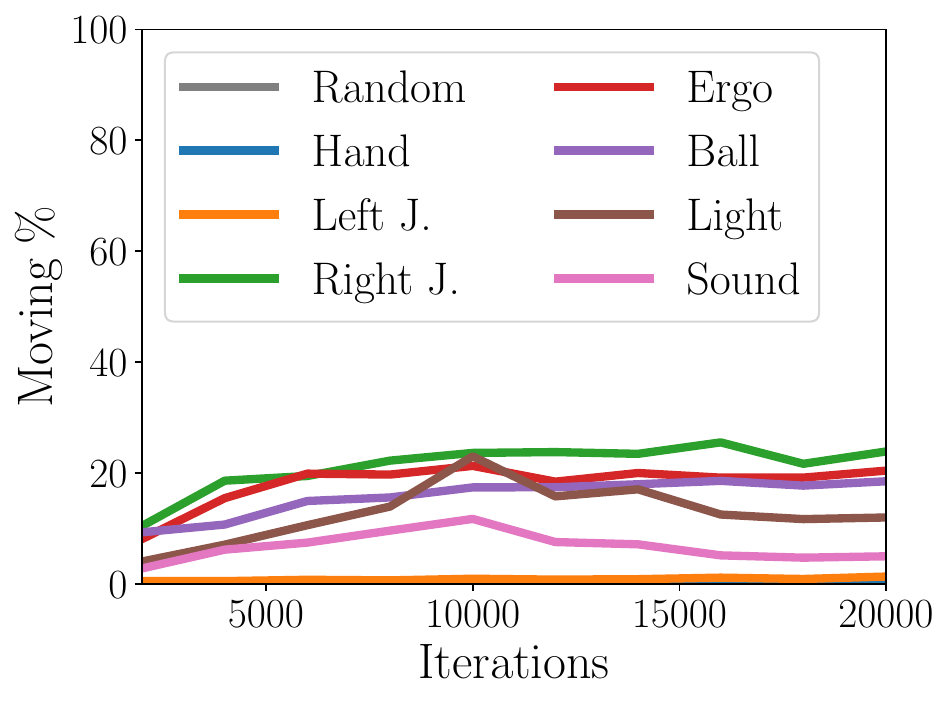}}

\caption{Stepping-stone structure of the three environments. In the 2D Simulated environment, we show the proportion of iterations that allowed to (a) move the magnet tool, (b) move the magnet toy,
depending on the currently explored goal space (or random movements), for $10$ IMGEP agents.
The fastest way to discover the tool is to explore the hand and to discover the toy is to explore the tool.
In the Minecraft Mountain Cart environment, we show the proportion of iterations that allowed to (c) move the pickaxe, (d) mine diamond blocks, and (e) move the cart,
depending on the currently explored goal space (or random movements), for $10$ agents with different seeds.
Exploring the agent space helps discover the pickaxe, exploring the pickaxe helps discover the blocks, and exploring the blocks helps discover the cart.
In the Robotic environment, we show the proportion of iterations that allowed to (f) reach the left joystick, (g) reach the right joystick, and (h) move the Ergo robot, depending on the currently explored goal space (or random movements), averaged for $11$ IMGEP agents with different seeds.
Exploring random movements or the Hand space helps discover the left joystick, exploring the left joystick helps discover the right one, which helps discover the Ergo toy.
}
\label{transfer}
\end{figure}

\paragraph{}\textit{Discoveries ---}~
In order to understand the tool-use structure of the exploration problem in each environment, we can look in more details how agents succeeded to move objects while exploring other objects. 
Indeed, to the agents starting to explore, tools are objects like any other object (e.g. the hand, the stick and the ball have the same status). However, if a tool needs to be used to move another object, then this tool will be discovered before that object, so the exploration of this tool is a stepping-stone giving more chances to discover novelty with that object than the exploration of any other object. To quantify these dependencies between objects in our tool-use environments, we show the proportion of movements where an object of interest has been moved depending on the currently explored object.

Concerning the 2D simulated environment, Fig. \ref{transfer} shows the proportion of the iterations with a goal in a given space that allowed to move (a) the magnet tool, (b) the magnet toy, in 10 runs with different seeds.
First, random movements of the arm have almost zero chances to reach the magnet tool or toy.
Exploring movements of the hand however have about $1.5\%$ chances to move the magnet tool, but still almost zero chances to reach the toy. Exploring the magnet tool makes this tool move in about $93\%$ of the iterations, and makes the toy move in about $0.1\%$ of movements. Finally, exploring the toy makes the tool and the toy move with a high probability as soon as the toy was discovered.
Those results illustrate the stepping-stone structure of this environment, where each object must be well explored in order to discover the next step in complexity (Hand $\rightarrow$ Tool $\rightarrow$ Toy).

In Minecraft Mountain Cart (see Fig. \ref{transfer}(c,d,e)), random exploration with neural networks in this environment is extremely challenging. An agent following random policies has $0.04\%$ chances to discover the pickaxe, $0.00025\%$ chances to break a single block and it never managed to move the cart (over 800k episodes). IMGEP agents reach better performances by leveraging the sequential nature of the environment: when exploring the agent space there is around $10\%$ chances to discover the pickaxe, and exploring the pickaxe space has around $1\%$ chances to break blocks. Finally, exploring the block space has about $8\%$ chances to lead an agent to discover the cart.

In the Robotic environment, a similar stepping-stone exploration structure is displayed (see Fig. \ref{transfer}(f,g,h)): in order to discover the left joystick, the robots needs to do random movements with its arm, which have about $2.9\%$ chances to makes the left joystick move, or explore its hand ($0.2\%$ chance). To discover the right joystick, the agent has to explore the left joystick, which gives a probability of $3.3\%$ to reach the right one.
To discover the Ergo (the white robotic toy in the center of the blue arena), the exploration of the right joystick gives $23\%$ chances to move it, whereas the exploration of the Hand, the left joystick or random movements has a very low probability to make it move.

\subsubsection{Learned Skills}

\begin{figure}[t]
\centering
\subfloat[Agent moving the cart]{\includegraphics[width=0.4\textwidth]{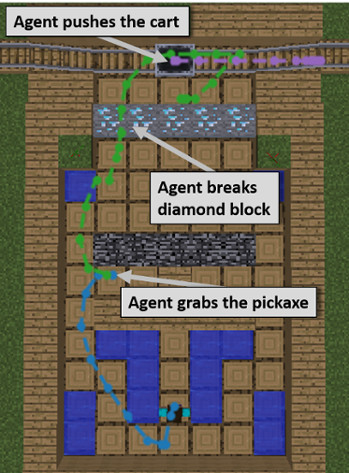}}\hspace{0.2cm}
\subfloat[5 cart goals]{\includegraphics[width=0.5\textwidth]{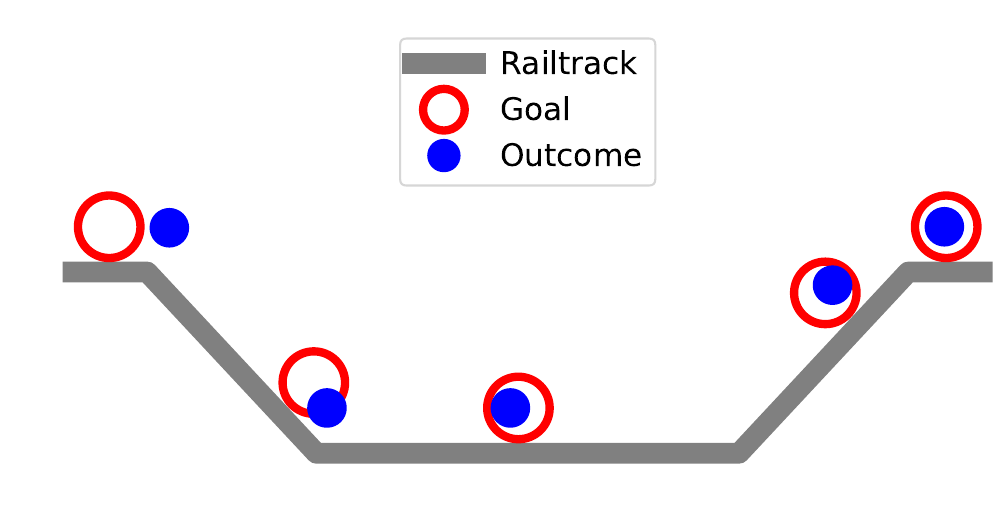}}
\caption{Example of learned skills in the Minecraft Mountain Cart. (a) One AMB agent's trajectory for a single cart goal. (b) Five final cart positions reached by an AMB agent when tasked to reach five different targets.
This agent successfully learned to push the cart along the track.}
\label{learned_skills}
\end{figure}

In Minecraft Mountain Cart we performed post-training tests of competence in addition of exploration measures. Using modular approaches allows to easily test competence on specific objects of the environment. Fig. \ref{learned_skills}(b) shows an example in the cart space for an AMB agent. This agent successfully learned to move the cart close to the 5 queried locations.

For each of the RMB, AMB and FC runs we performed a statistical analysis of competence in the cart and pickaxe spaces using 1000 and 800 uniformly generated goals, respectively. We were also able to test SGS trials for cart competence as this condition has the cart as goal space. A goal is considered reached if the Euclidean distance between the outcome and the goal is lower than $0.05$ in the normalized space (in range $[-1,1]$) for each object. Since the pickaxe goal space is loosely defined as a rectangular area around the environment's arena, many goals are not reachable. Results are shown in Table \ref{table_mmc_comp}. SGS agents never managed to move the cart for any of the given goals. AMB appears to be significantly better than RMB on the pickaxe space ($p<0.01$ on Welch's t-tests). However it is not in the cart space ($p=0.09$), which might be due to the stochasticity of the environment. FC is not significantly better than AMB on the cart and pickaxe spaces.

\begin{table}[t]
\centering
\setlength\tabcolsep{6pt}
\begin{tabular}{|l|c|c|}
  \hline
   & Pickaxe goals & Cart goals \\
  \hline
  FC & 39,49,55 & 12,17,25 \\
  \hline
  AMB & 41,45,49 & 8,11,18 \\
  \hline
  RMB & 37,40,43 & 6,9,15 \\
  \hline
  SGS & N/A &  0,0,0 \\
    \hline
\end{tabular}
\vspace{0.3cm}
\caption{Competence results in Minecraft Mountain Cart. We give the 25, 50 and 75 percentiles of the competence results of all seeds.}
\label{table_mmc_comp}
\end{table}

\paragraph{}\textit{Intrinsic Rewards based on Learning Progress ---}~
The IMGEP agents self-evaluate their learning progress to control each object.
When they choose a goal for an object, they monitor what is the actual movement given to the object and compare it to the goal.
If the distance between the goals and the actual reached movements decrease over time on average, this tells the agents it is making progress to control this object.
This signal is used as an intrinsic reward signal that the agent will seek to maximize by choosing to explore objects that yield a high learning progress.
We can analyze this signal to understand at which point the agent perceived progress to control each object and how its exploration behavior changed over time.

\begin{figure}[p]
\centering
\includegraphics[width=0.432\textwidth]{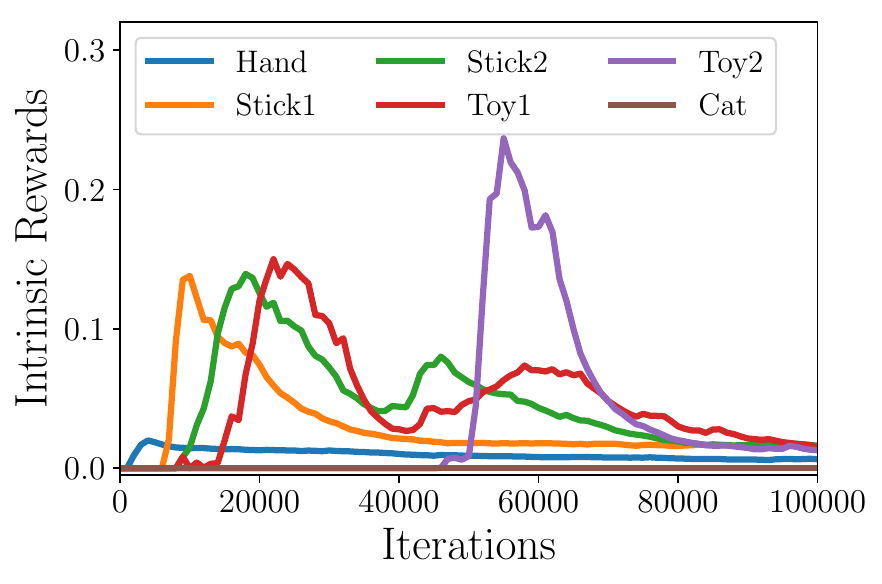}\hspace{0.2cm}
\includegraphics[width=0.44\textwidth]{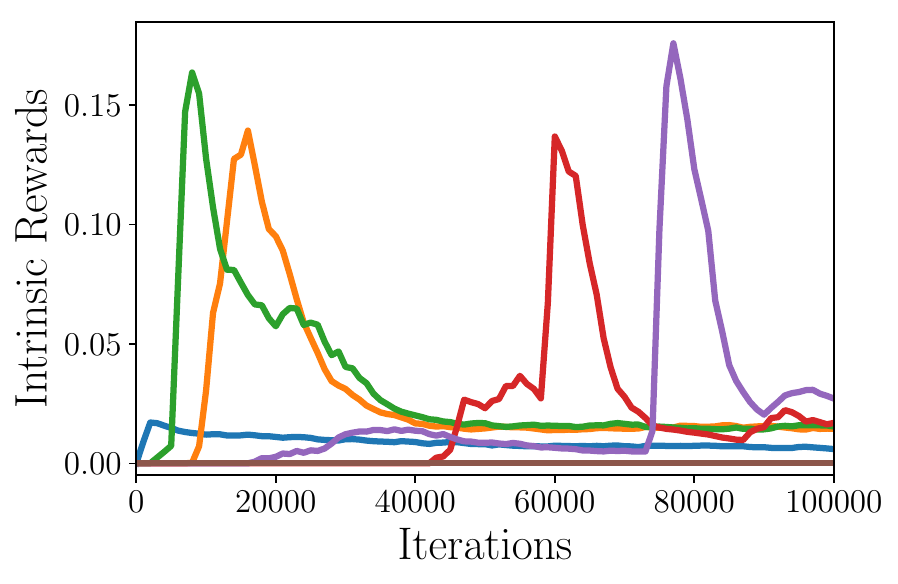}\\

\includegraphics[width=0.44\textwidth]{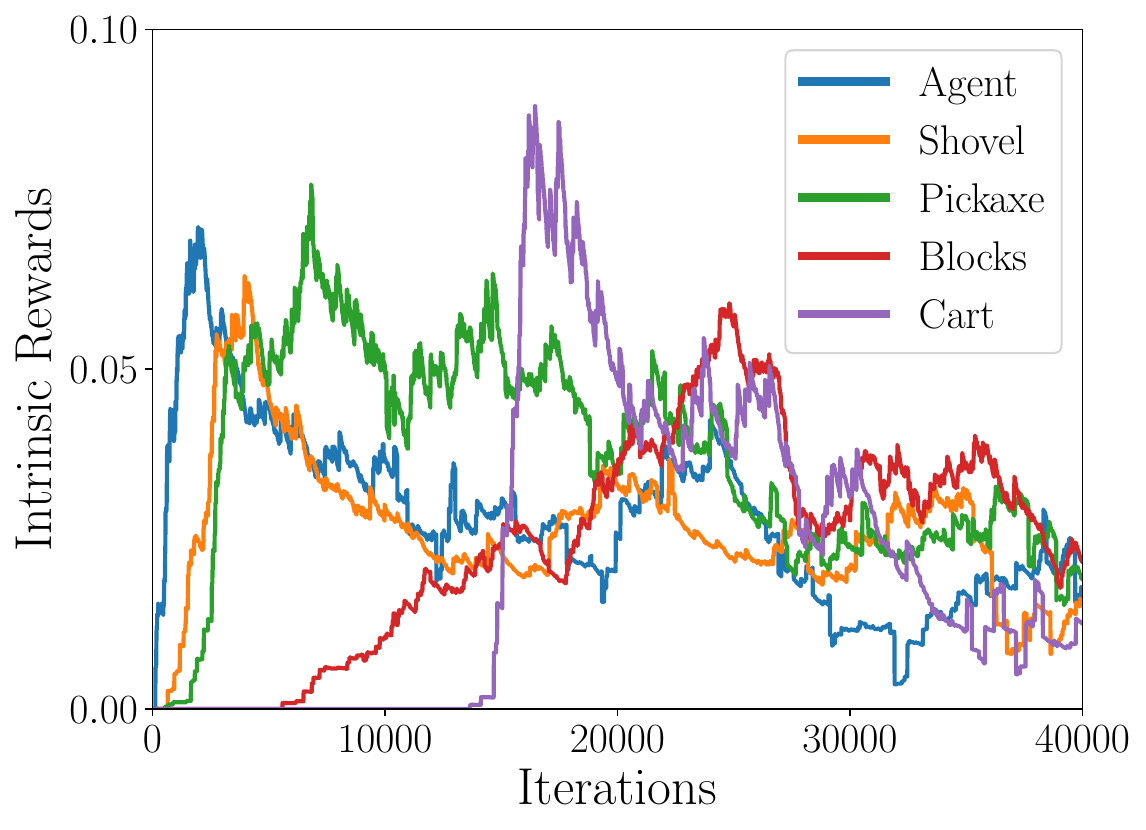}\hspace{0.2cm}
\includegraphics[width=0.44\textwidth]{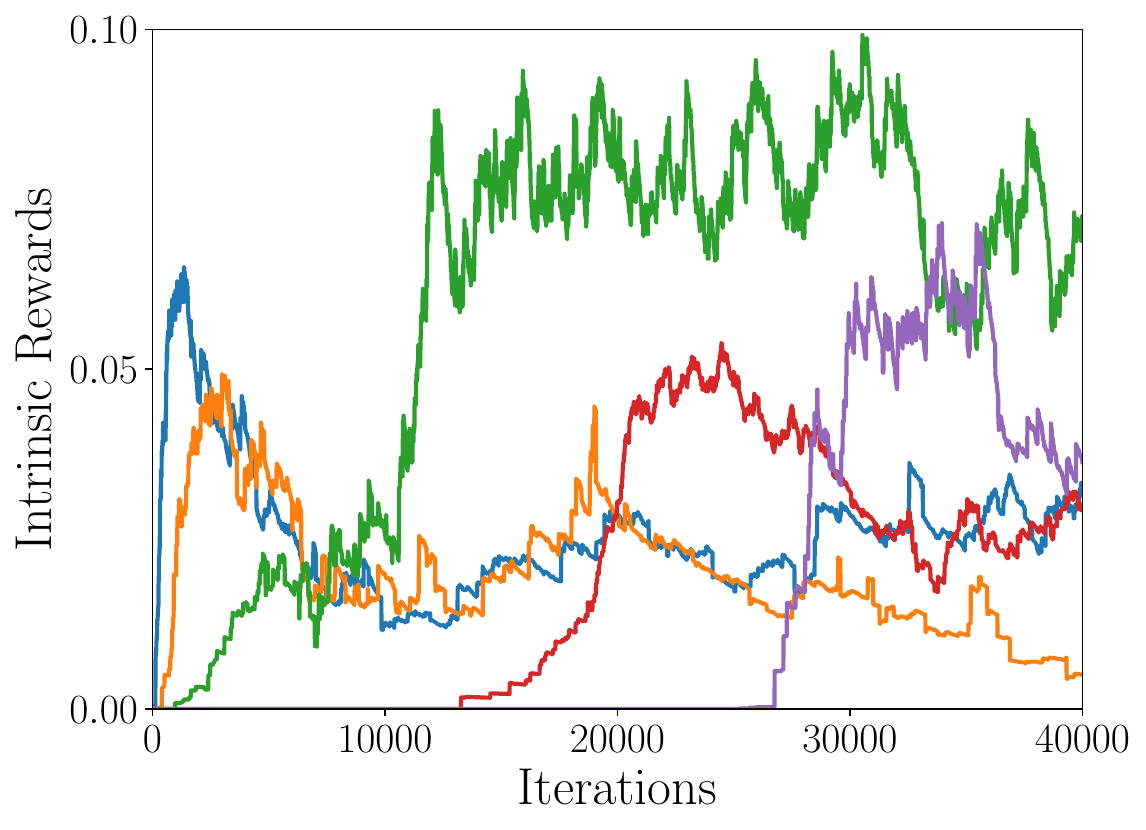}\\

\includegraphics[width=0.44\textwidth]{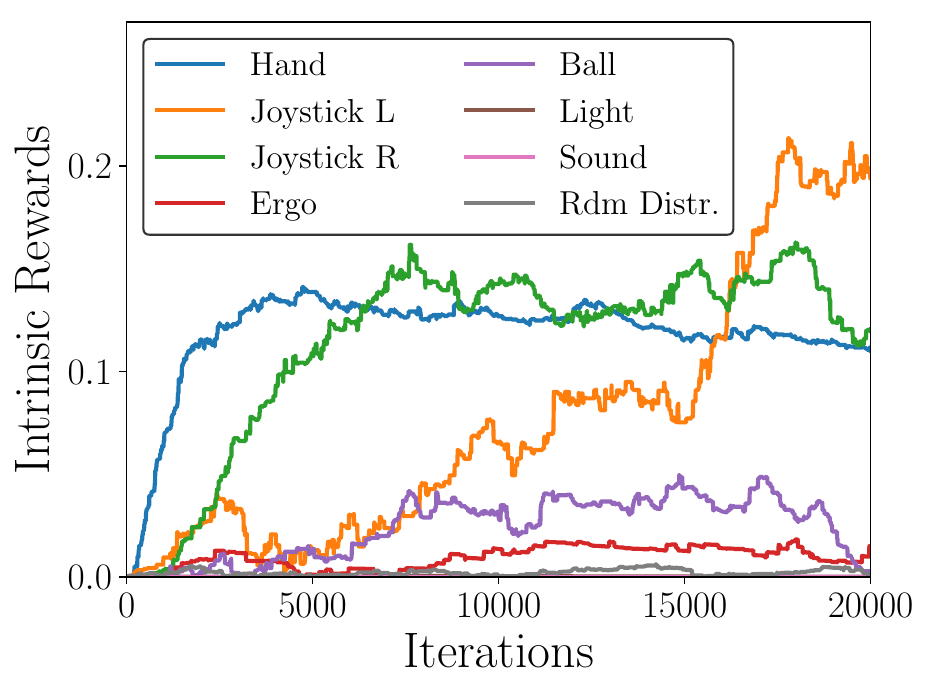}\hspace{0.2cm}
\includegraphics[width=0.44\textwidth]{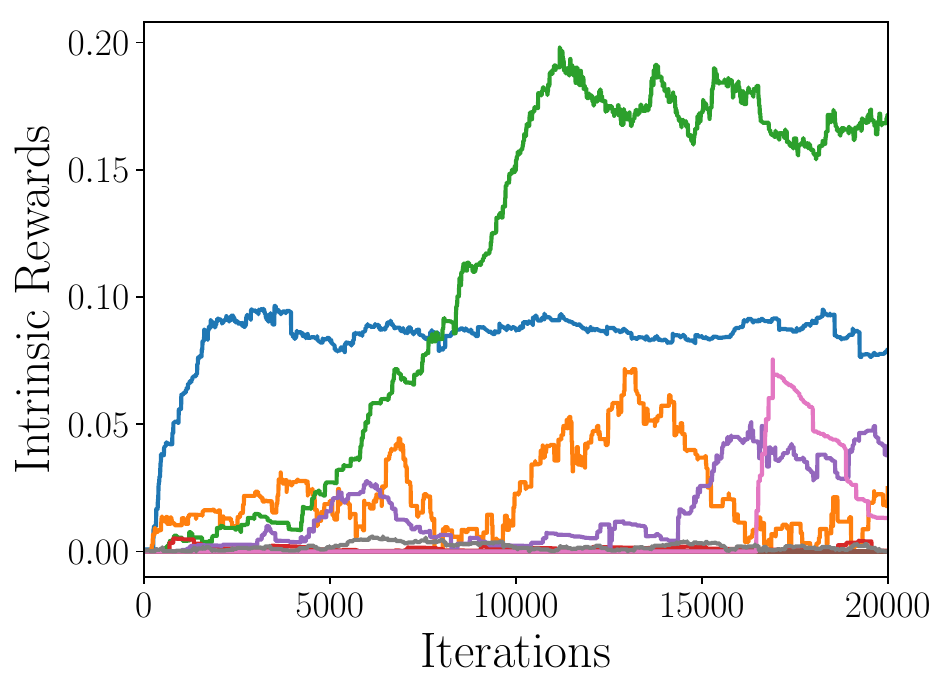}
\caption{Examples of intrinsic rewards in the three environments.
In the 2D Simulated environment (top), agents are first interested in exploring their hand as this is the only object they manage to move, until they discover one of the sticks. 
Then they make progress to move the stick, so the intrinsic reward for moving the stick increases and they focus on it, and then on the other objects they discover: the other stick and the two toys.
They make no progress to move the distractors so those intrinsic reward are always zero.
In the Robotic environment (middle), agents first succeed to move their hand, so they focus on this object at the beginning, until they discover the joysticks. The exploration of the right joystick makes them discover the Ergo toy, which can push the Ball. Some agents also discover how to produce light and sound with the Ball. Agents have a low intrinsic reward for exploring random distractors.
In Minecraft Mountain Cart (bottom), agents first focus on exploring the space of their position until they discover the shovel or the pickaxe and start making progress to move them. When they discover how to mine blocks with the pickaxe and to push the cart, they make progress in those goal spaces, get intrinsic rewards and thus focus more on these.}
\label{interests}
\end{figure}

Fig.\,\ref{interests}\,(top) shows the intrinsic rewards of two agents (different seeds) to explore each object in the 2D simulated environment, computed by the agents as the average of intrinsic rewards based on learning progress to move each object.
We can see that the intrinsic reward of the hand increases first as it is the easiest object to move.
Then, when the sticks are discovered, the agents start to make progress to move them in many directions. Similarly, while exploring the sticks, they discover the reachable toys, so they start making progress in moving those toys.
However, the static objects can't be moved so their learning progress is strictly zero, and the objects moving randomly independently of the agent (cat and dog) have a very low progress.

Fig.\,\ref{interests}\,(middle) shows the intrinsic reward of two agents in the Minecraft Mountain Cart environment. Both agents first explore the simpler agent space and then quickly improves on the shovel and pickaxe spaces. Exploring the pickaxe space leads to discover how to progress in the block space. Finally, after some progress in the block space, the cart is discovered after 14k episodes for the first agent (left figure) and 26k episodes for the other (right figure). The $3$ distracting flowers have an interest strictly equal to zero in both runs.

Fig.\,\ref{interests}\,(bottom) shows the intrinsic reward of two agents in the Robotic environment.
The first interesting object is the robot's hand, followed by the left joystick and then the right joystick.
The left joystick is the easiest to reach and move so it gets interesting before the right one in most runs, but then they have similar learning progress curves.
However, the right joystick can be used as a tool to control other objects, so that one will be touched more often.
Then, the agent can discover the Ergo and Ball while exploring the joysticks.
Finally, some agents also discover that the ball can be used to make light or sound.
Here also, the progress of static objects is zero and the one of random objects is low. \textcolor{black}{Note that, unlike in the 2D simulation and in the Minecraft environment, the intrinsic reward for exploring the robot's hand remains high. This phenomenon is most likely due to the use of full trajectories as goals in the Robotic environment, which creates large goal spaces (compared to when using end-positions as goals). Additionally, the hand space is quite homogeneous: there are many trajectories that are equally learnable, so the hand space remains for a very long time a source of learning progress. For more difficult objects, the space is also large, but the subspace of learnable trajectories is much smaller, thus the decrease in LP can be seen.}

Overall, the evolution of those interests show that evaluating the learning progress to move objects allows agents to self-organize a learning curriculum focusing on the objects currently yielding the most progress and to discover stepping stones one after the other.

\subsubsection{Influence of Goal Modularity}

\begin{figure}[t]
\centering
\subfloat[Magnet Tool]{\includegraphics[width=0.48\textwidth]{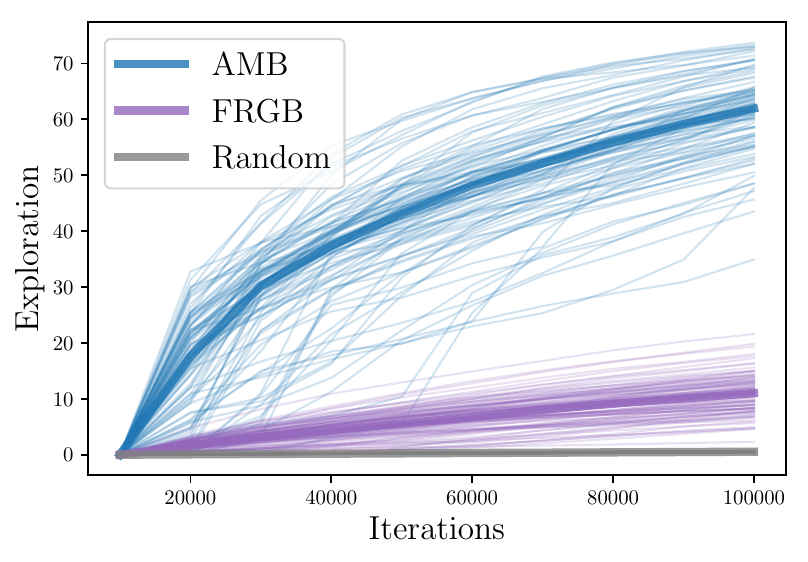}}\hspace{0.2cm}
\subfloat[Magnet Toy]{\includegraphics[width=0.48\textwidth]{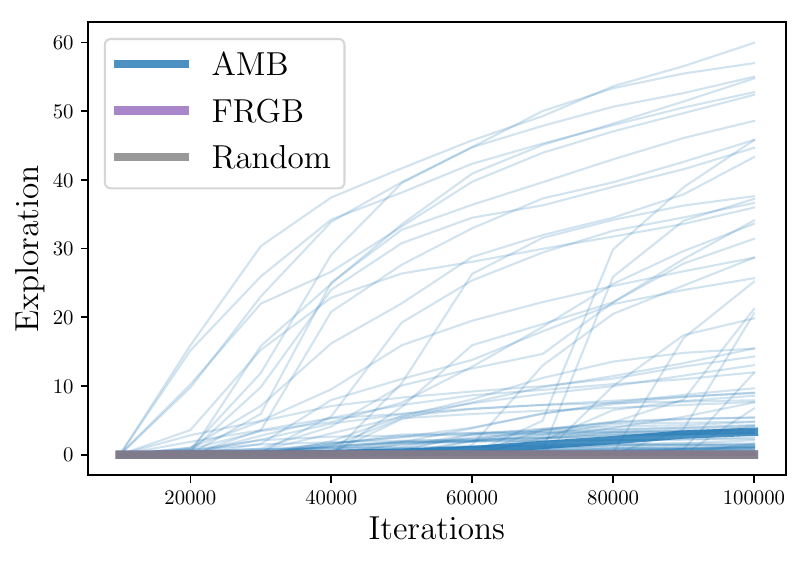}}
\caption{Influence of Goal Modularity on exploration in the 2D simulated environment.
The agents using a modular representation (Active Model Babbling) explore much better the tool and toy spaces than agents with a flat representation (Flat Random Goal Babbling).
Control agents always choosing completely random actions do not manage to touch a toy with the stick.}
\label{explo_2DSimu}
\end{figure}

In this section, we study several algorithms with a different goal space structure in order to evaluate the influence of the modularity of the goal space.
We compare the Active Model Babbling condition to other conditions.
In the Flat Random Goal Babbling (FRGB) condition, the goal space is not modular and contains all variables of all objects. With this non-modular sensory representation, agents choose goals in the whole sensory space, which corresponds to all objects: a goal could mean for instance push \textit{toy1} to the left and \textit{toy2} to the right at the same time, which might be unfeasible. This exploration dynamics results in exploring the most diverse global sensory states, which is akin to novelty search algorithms \citep{lehman2011abandoning}.
We also test the Random control where agents always choose random actions.

In the 2D simulated environment, we run 100 trials of each condition with different random seeds.
We measure the exploration of one stick and its corresponding toy as the cumulative number of reached cells in a discretization of the 2D space of the position of each objects at the end of movements.
Fig. \ref{explo_2DSimu} shows the evolution of the exploration of the stick and the toy in the 100 trials of each condition.
We plot in bold the median over the 100 trials in each condition.
We can see that the modularity of the goal space helps exploration: the median exploration after $100$k iterations is about $60\%$ of the magnet tool space for condition AMB vs about $10\%$ for condition FRGB.
The agents in condition AMB succeeded to reach the magnet toy, with a substantial variance between the 100 trials.
Some AMB agents explored very well the magnet toy (up to $60\%$) and some did not (very low exploration).
Finally, completely random agents did not even manage to explore the magnet tool.

\begin{figure}[t]
\centering
\subfloat[Agent]{\includegraphics[width=0.31\textwidth]{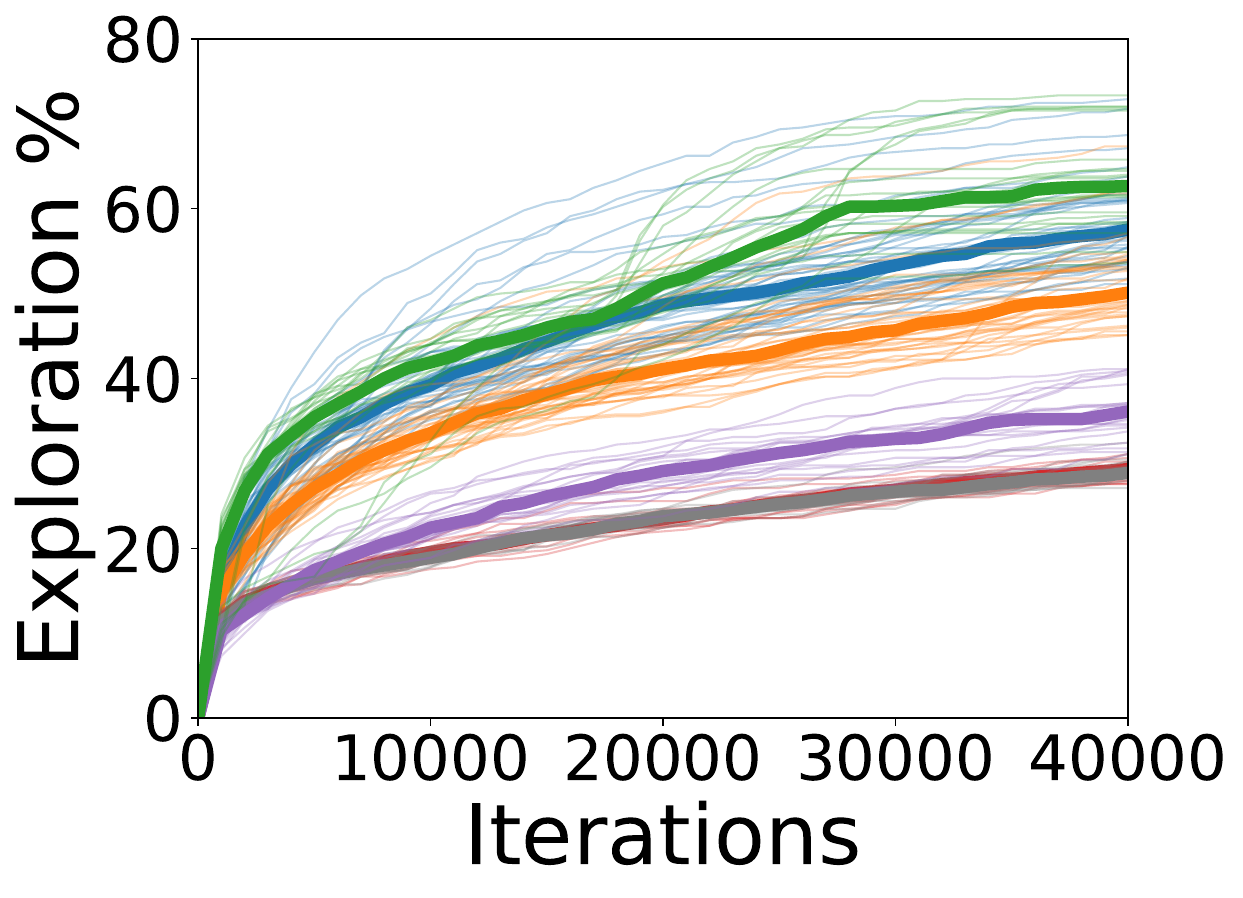}}\hspace{0.2cm}
\subfloat[Shovel]{\includegraphics[width=0.31\textwidth]{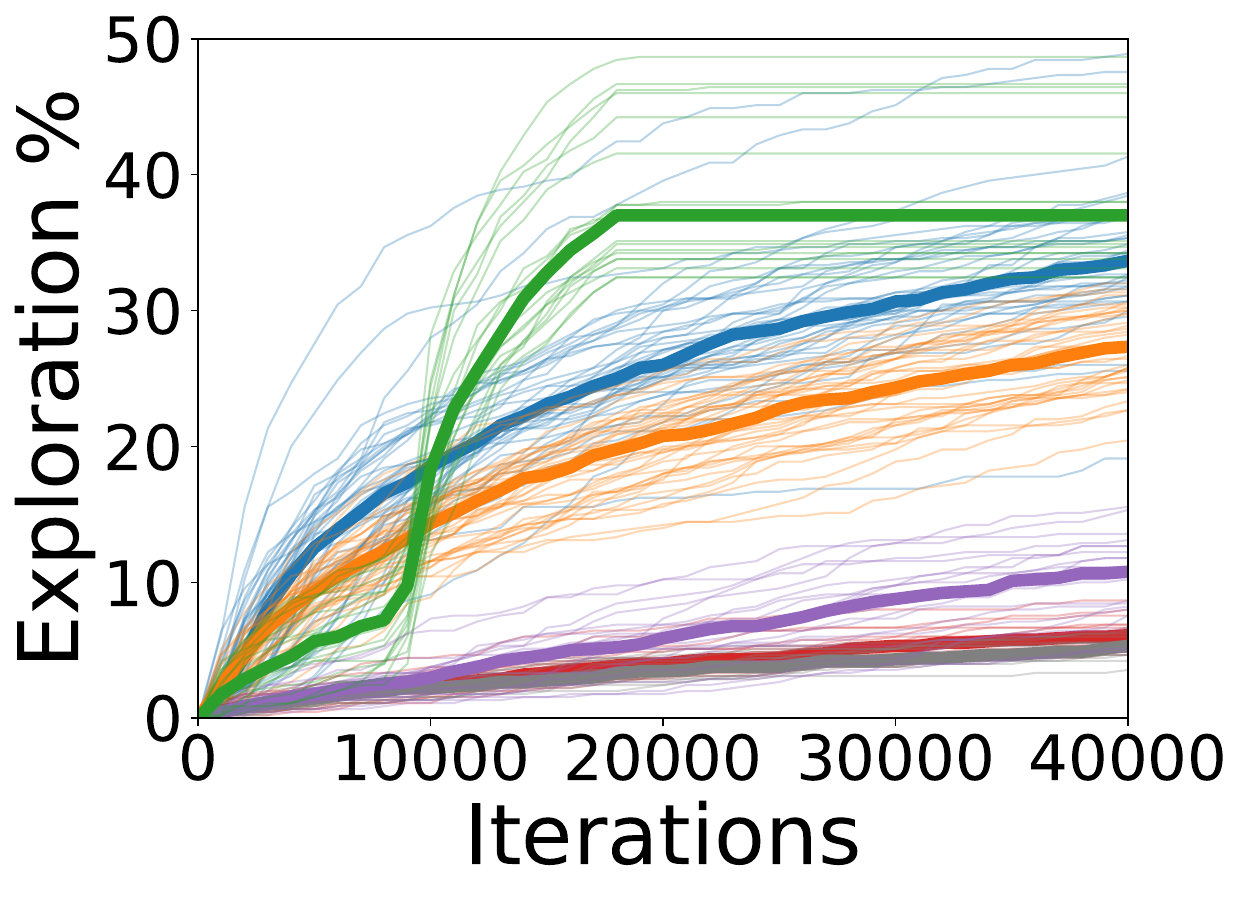}}\hspace{0.2cm}
\subfloat[Pickaxe]{\includegraphics[width=0.31\textwidth]{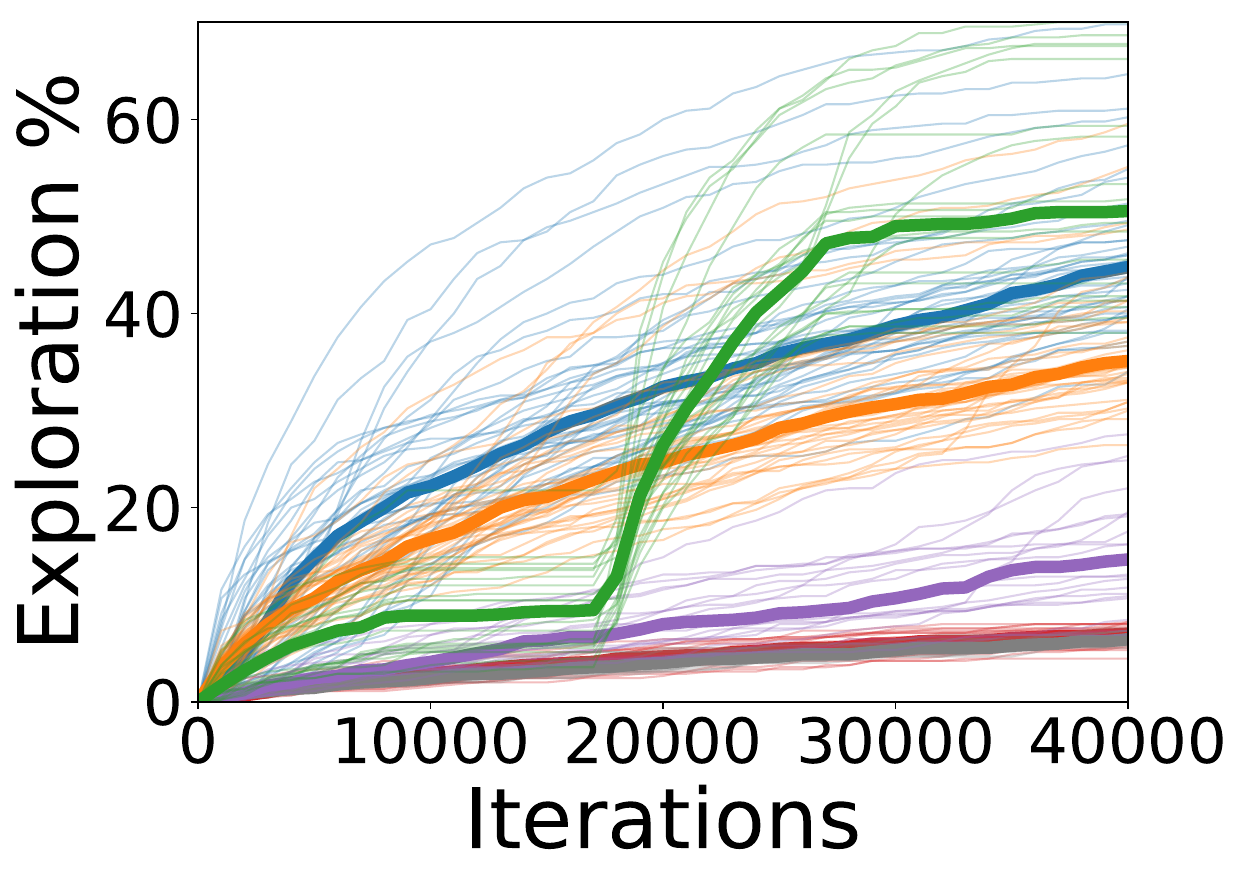}}\\
\subfloat[Blocks]{\includegraphics[width=0.33\textwidth]{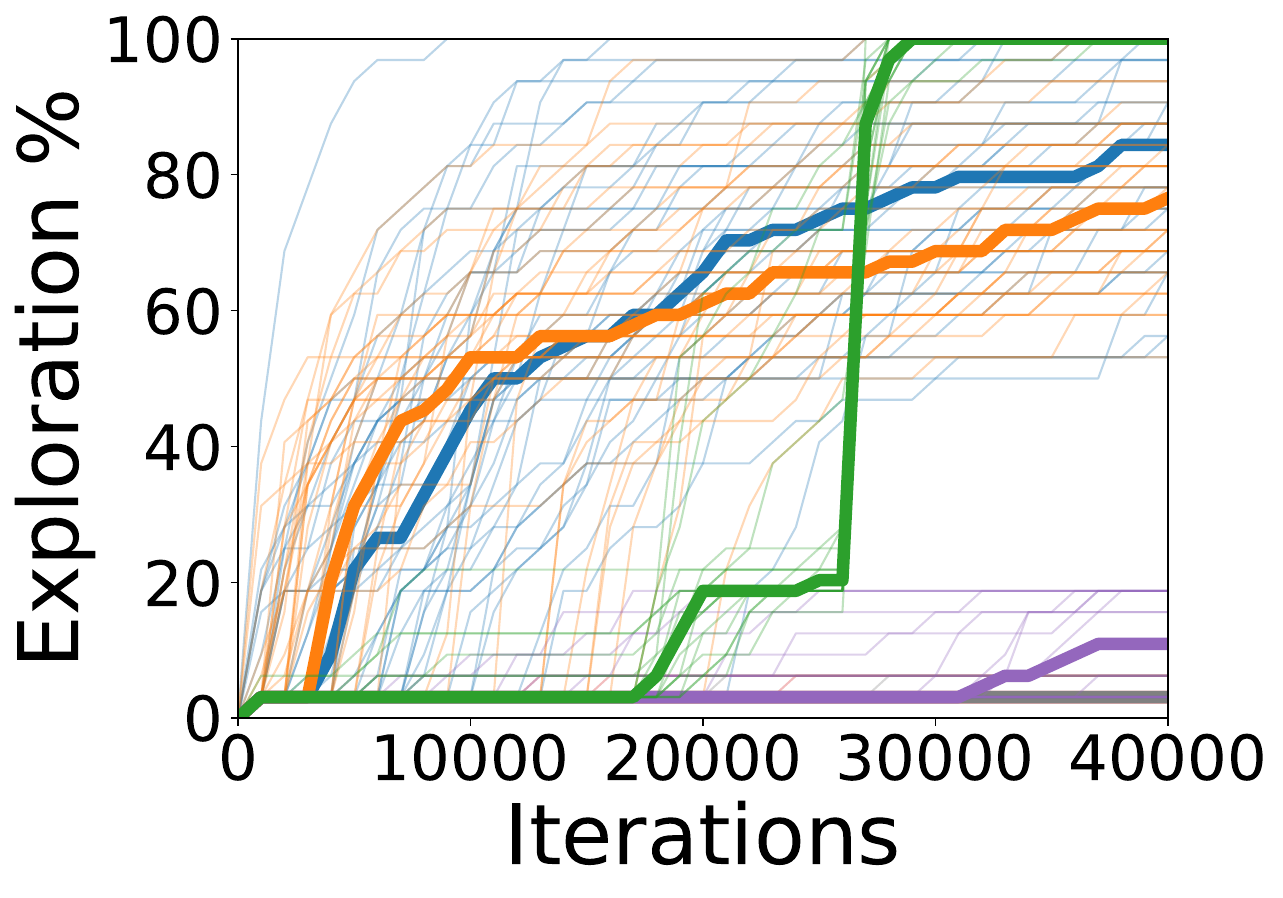}}\hspace{0.2cm}
\subfloat[Cart]{\includegraphics[width=0.34\textwidth]{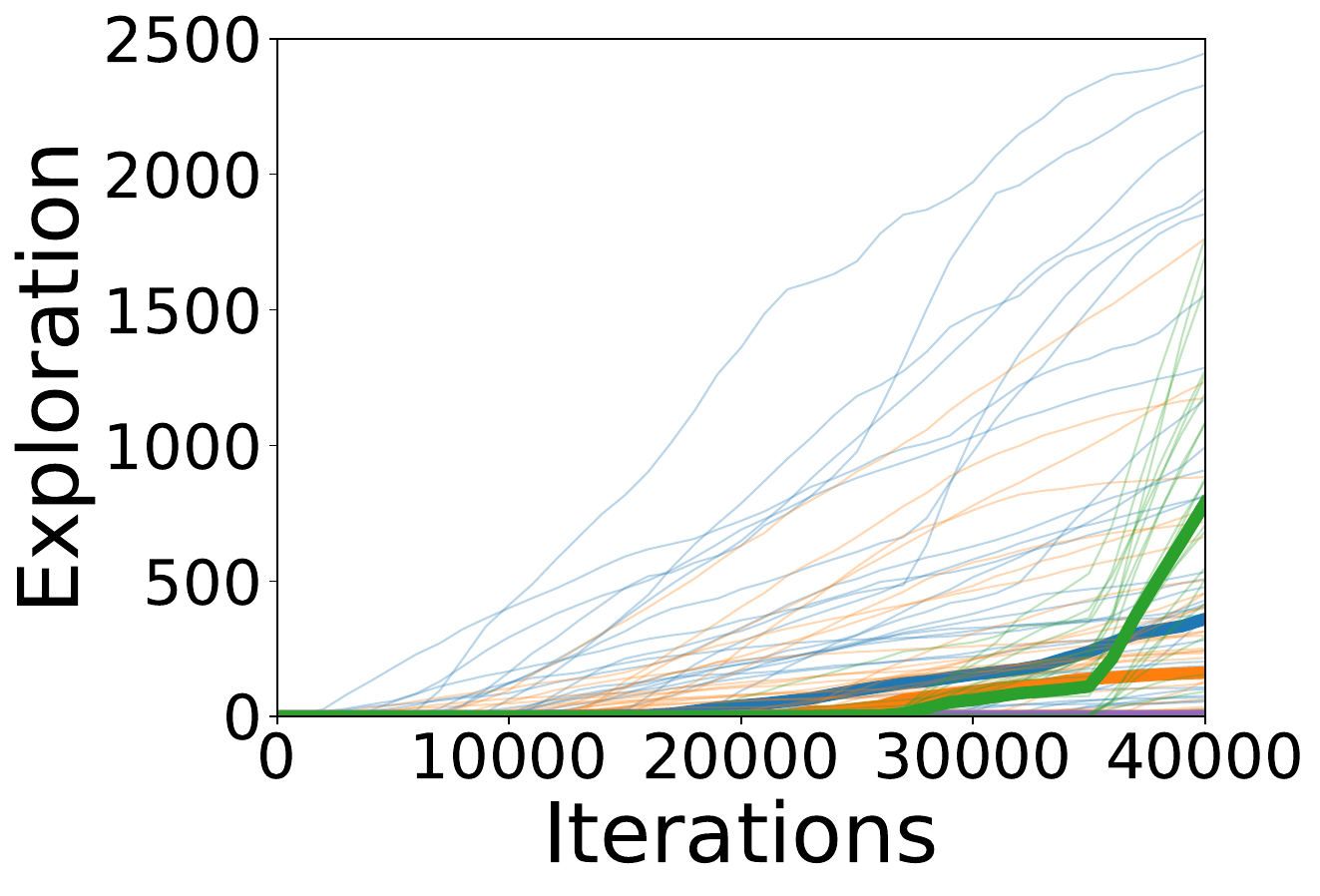}}\hspace{0.5cm}
\subfloat{\includegraphics[width=0.25\textwidth]{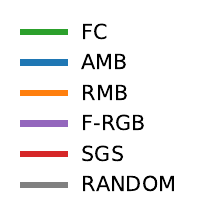}}
\caption{Exploration results in Minecraft Mountain Cart. Modular approaches (AMB and RMB) performs significantly better than the flat (F-RGB) approach. Agents actively generating their curriculum (AMB) perform better overall than agents choosing goal spaces randomly (RMB). Focusing on the cart space (SGS) is equivalent to performing random policies (Random). For the agent, pickaxe and shovel spaces, exploration is measured as the cumulative number of reached cells in a discretization of the $2$D space. For the block and cart spaces we measure the number of unique outcomes reached.}
\label{explo_MMC}
\end{figure}

Fig. \ref{explo_MMC} shows exploration results in the Minecraft Mountain Cart environment for 20 trials of all conditions except for AMB and RMB which were run 42 times. When looking at the median exploration in the pickaxe space, FRGB does not manage to reach more than $15\%$ exploration when AMB and RMB reached $45\%$ and $35\%$, respectively. Modular approaches significantly outperform FRGB across all goal spaces (Welch's t-tests at 40k iterations, $p<0.001$). Random agents did not manage to explore the block and cart spaces.

\begin{figure}[p]
\centering
\subfloat[Hand]{\includegraphics[width=0.45\textwidth]{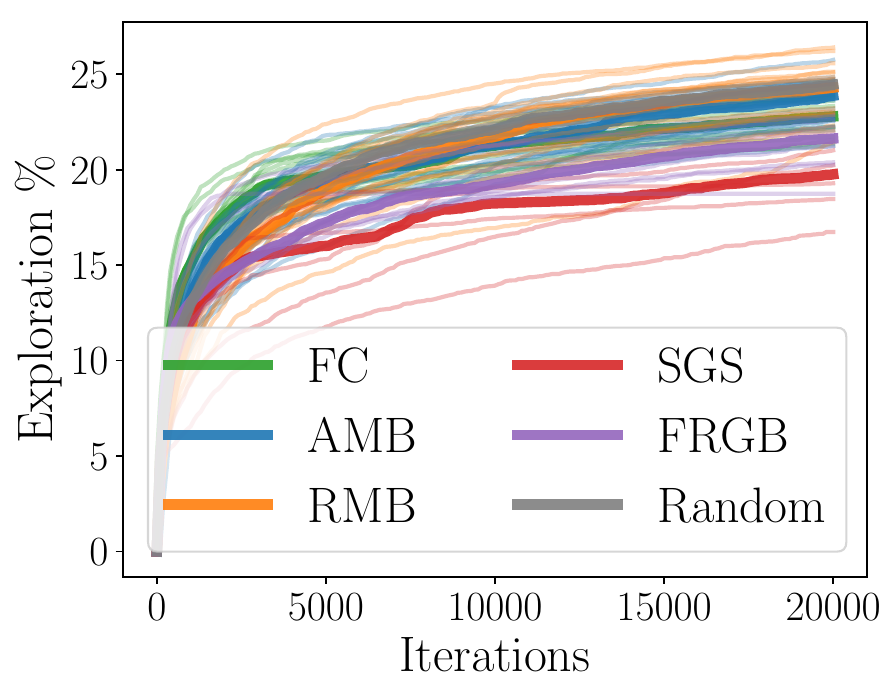}}\hspace{0.2cm}
\subfloat[Joystick Left]{\includegraphics[width=0.45\textwidth]{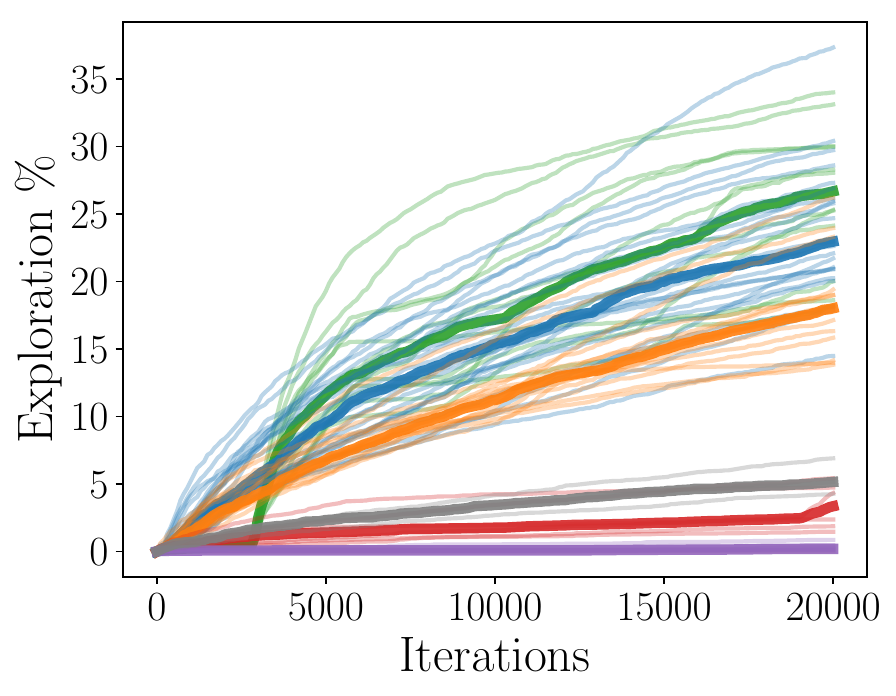}}\\
\subfloat[Joystick Right]{\includegraphics[width=0.45\textwidth]{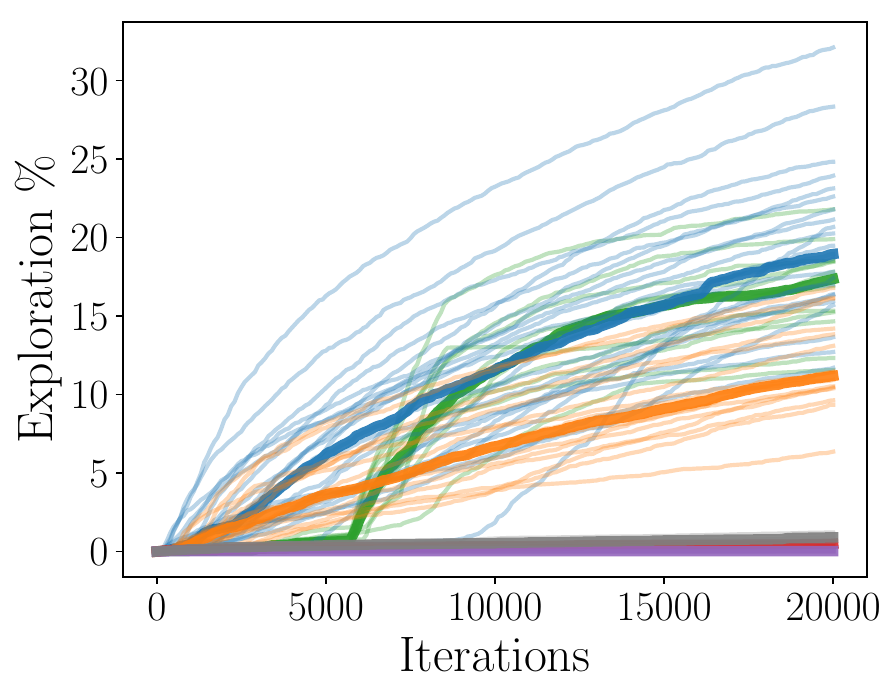}}\hspace{0.2cm}
\subfloat[Ergo]{\includegraphics[width=0.46\textwidth]{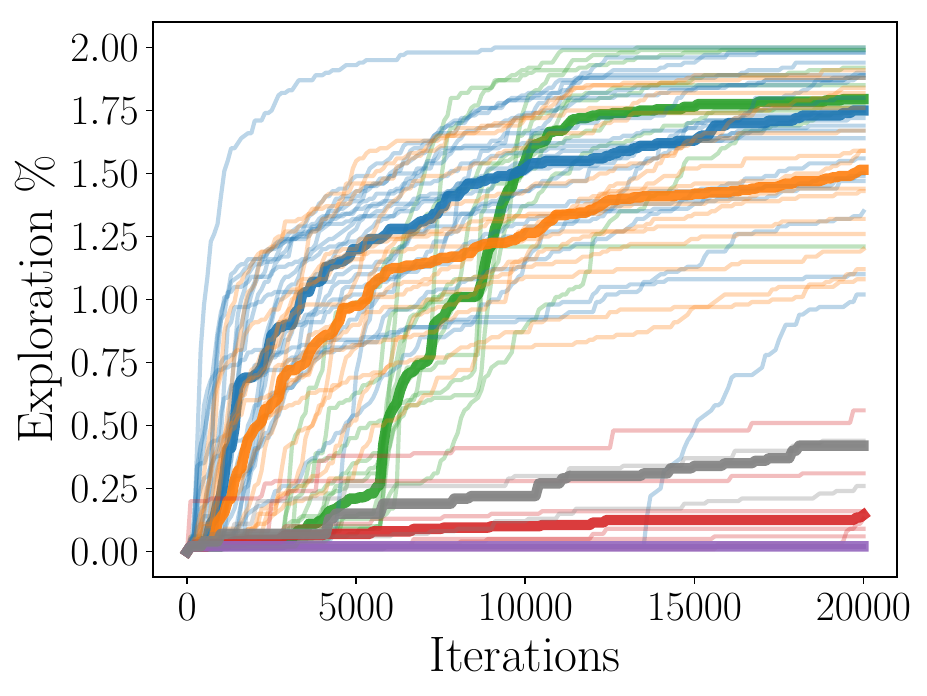}}\\
\subfloat[Ball]{\includegraphics[width=0.31\textwidth]{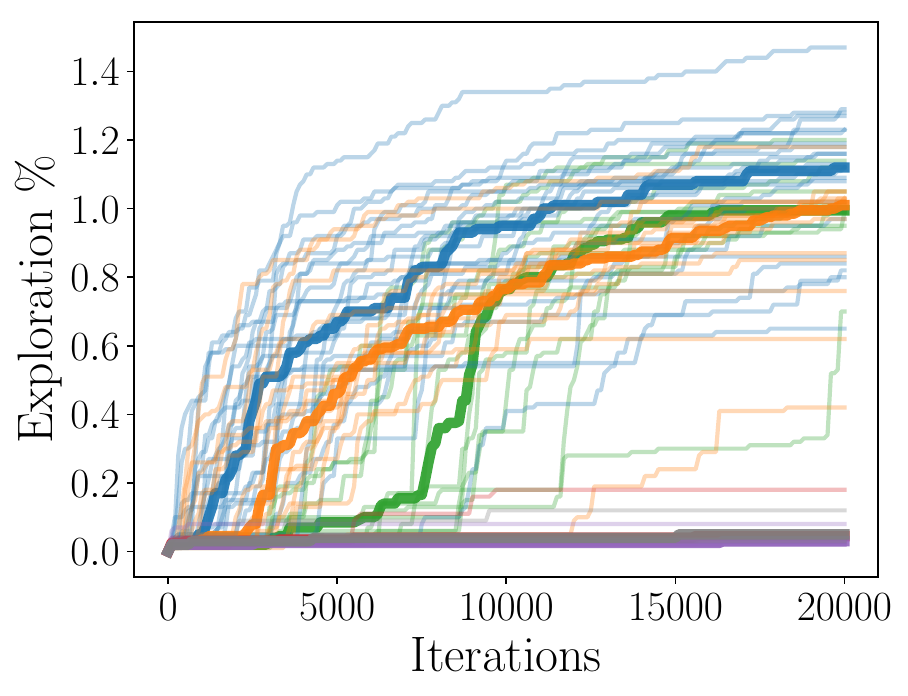}}\hspace{0.2cm}
\subfloat[Light]{\includegraphics[width=0.31\textwidth]{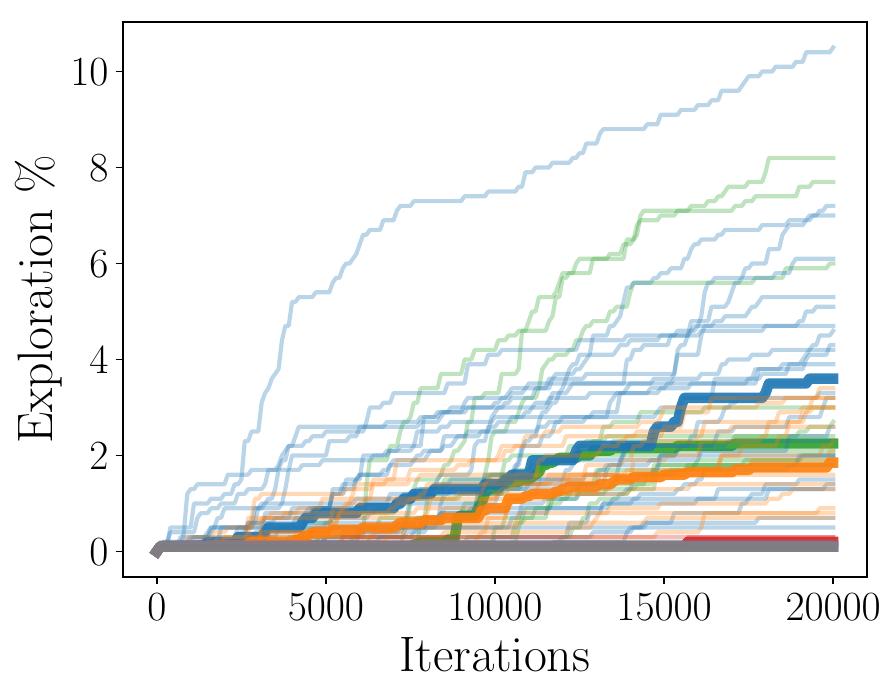}}\hspace{0.2cm}
\subfloat[Sound]{\includegraphics[width=0.31\textwidth]{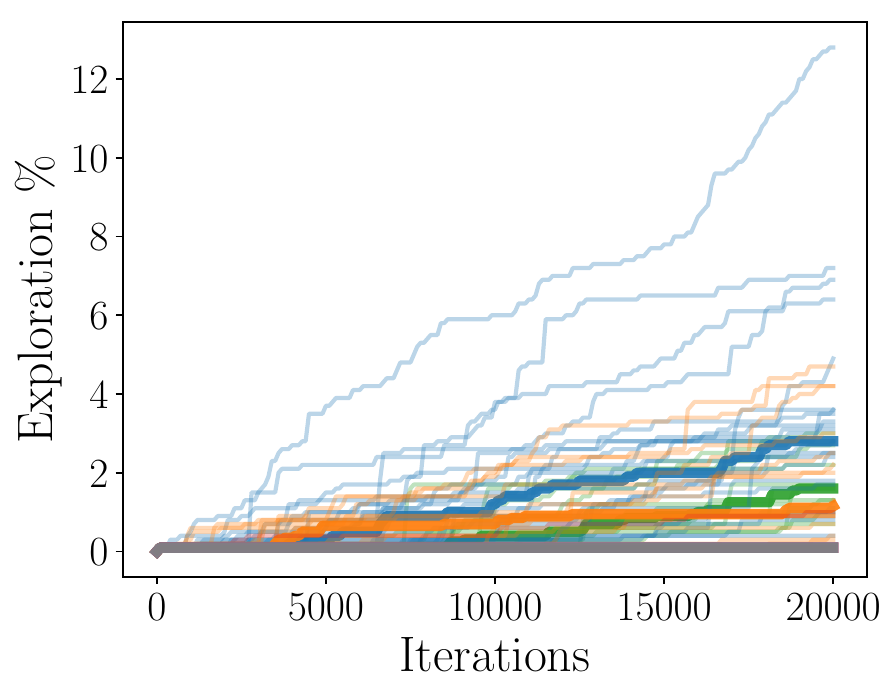}}
\caption{Exploration results in the Robotic environment. Agents self-organizing their curriculum (Active Model Babbling) based on their learning progress explore better than agents choosing random objects (Random Model Babbling) in the joysticks, ball, light and sound spaces, and better than agents with a hard-coded curriculum sequence (FC) in the right joystick and sound spaces. Agents always choosing to explore the ball (Single Goal Space) and agents without a modular representation of goals (FRGB) have low exploration results in all spaces.}
\label{explo_Torso}
\end{figure}

In the robotic environment (see Fig. \ref{explo_Torso}), agents with the flat (intricate) representation of the sensory feedback (FRGB) do not explore objects other than the hand.

The modular representation of the sensory space thus greatly improves exploration efficiency compared to a flat intricate representation of the whole sensory feedback, as it allows to consider the different objects independently to monitor their behavior and select disentangled goals.

\begin{figure}[t]
\centering
\subfloat[Magnet Tool]{\includegraphics[width=0.48\textwidth]{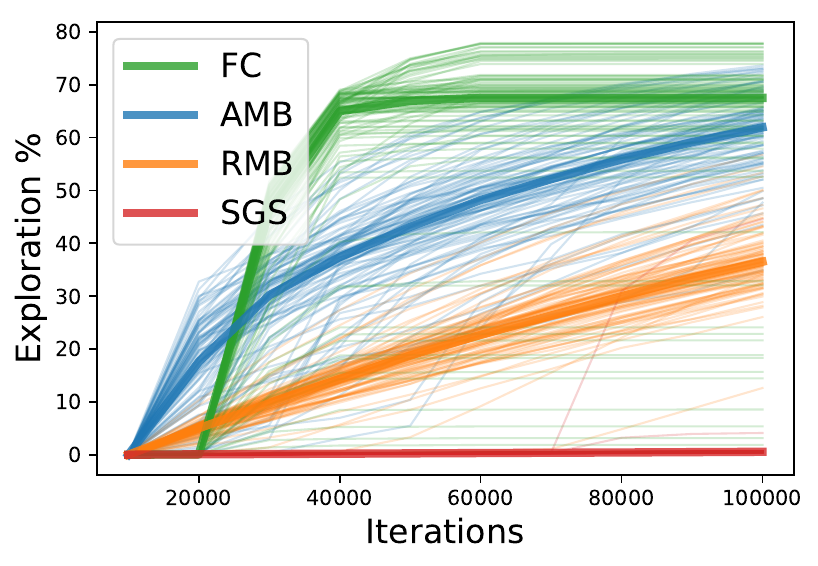}}\hspace{0.2cm}
\subfloat[Magnet Toy]{\includegraphics[width=0.48\textwidth]{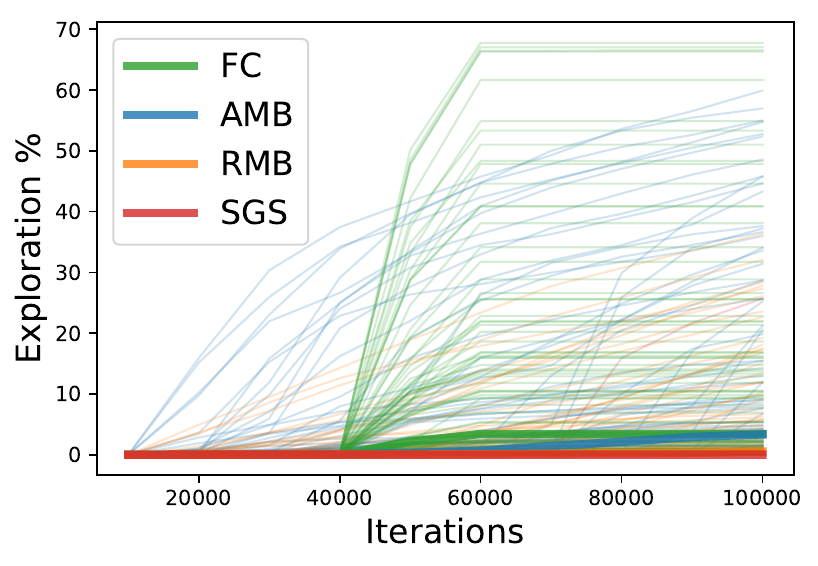}}
\caption{Influence of curriculum learning on exploration in the 2D Simulated environment. Agents self-organizing their curriculum (Active Model Babbling) based on their learning progress explore better than agents choosing to explore random objects (Random Model Babbling) or agents choosing always to explore the magnet toy (Single Goal Space).
Agents with a hard-coded curriculum learning sequence from the simpler objects to the most complex have similar exploration results than autonomous AMB agents after 100k iterations.}
\label{explo_2DSimu2}
\end{figure}

\subsubsection{Curriculum Learning}

A modular sensory representation based on objects allows AMB agents to self-monitor their learning progress to control each object, and to accordingly explore objects with high learning progress.
Here, we compare several variants of agents with a modular sensory representation based on objects, but with a different choice of object to explore.
To evaluate the efficiency of the sampling based on learning progress, we define condition Random Model Babbling (RMB), where agents always choose objects to explore at random.
The sampling based on learning progress of AMB agents makes agents explore any object that shows learning progress, and ignore objects that do not move, are fully predictable, or move independently of the agent.
However if we are only interested in a particular complex skill that we want the agent to learn, such as moving the ball in the robotic environment, then it is not obvious if supervising learning by specifying a curriculum targeted at this skill can accelerate the learning of this skill.
We thus define two control conditions with a hand-designed curriculum.
In condition Single Goal Space (SGS), agents always choose goals for the same complex target object: the magnet toy in the 2D simulated environment, the cart in Minecraft Mountain Cart, or the ball in the robotic environment.
In condition Fixed Curriculum (FC), a particular sequence of exploration is specified, from the easier stepping-stones to the more complex ones.
In the 2D simulated environment, the agent samples goals for 20k iterations on each object in this order: hand, magnet tool, magnet toy, Velcro tool, Velcro toy. 
In the robotic environment, we define the sequence as the following: hand, left joystick, right joystick, ergo, ball, light and sound.

Fig. \ref{explo_2DSimu2} shows the exploration evolution in the 2D simulated environment. 
First, we can see that the sampling based on learning progress (AMB) helps exploration of the tool and the toy compared to the random choice of object to explore (RMB): $62\%$ vs $37\%$ for the tool and $3.3\%$ vs $0.5\%$ for the toy.
Agents in the SGS condition did not manage to explore the tool and the toy.
Agents with a predefined curriculum succeeded to explore the tool and the toy very well, the tool between 20k and 40k iterations and the toy between 40k and 60k iterations, with a median slightly better than in AMB.

Fig. \ref{explo_MMC} shows the evolution of exploration in Minecraft Mountain Cart.
Agents focusing their goal sampling on the cart space (SGS) have low performances across all goal spaces, especially for the cart and block spaces which are never discovered. Agents using learning progress sampling (AMB) explored significantly more than random sampling agents (RMB) across all goal spaces (Welch's t-tests at 40k iterations, $p<0.04$). Agents following a hard-coded curriculum (FC) reached higher median performances than AMB agents on every goal spaces.

Fig. \ref{explo_Torso} shows exploration results in the robotic environment. 
Agents self-organizing their curriculum (Active Model Babbling) based on their learning progress explore better than agents choosing random objects (Random Model Babbling) in the joysticks, ball, light and sound spaces (Welch's t-tests at 100k iterations, $p<0.05$), and better than agents with a hard-coded curriculum sequence (FC) in the right joystick and sound spaces. Agents always choosing to explore the ball (SGS) and agents without a modular representation of goals (FRGB) have low exploration results in all spaces.

Overall, the goal sampling based on learning progress (AMB) improves exploration of most objects of each environment compared to a random choice of object (RMB), as those agents focus on objects that are learnable, ignore the distractor objects and reduce the relative interest of objects already explored for some time.
Specifying the curriculum by hand results in a very bad exploration if the agent always directly focuses on an object hard to discover, however if we carefully design the learning sequence given our knowledge of the task, then the final exploration results are similar to autonomous AMB agents.

\subsubsection{Influence of the Modularity of Exploration Mutations}

\begin{figure}[t]
\centering
\subfloat[\texttt{SSPMutation}]{\includegraphics[width=0.48\textwidth]{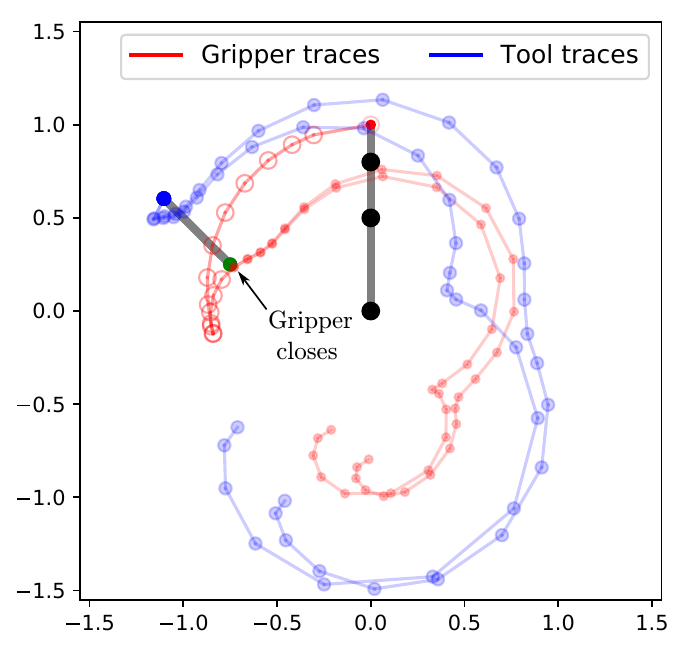}}\hspace{0.2cm}
\subfloat[\texttt{FullMutation}]{\includegraphics[width=0.48\textwidth]{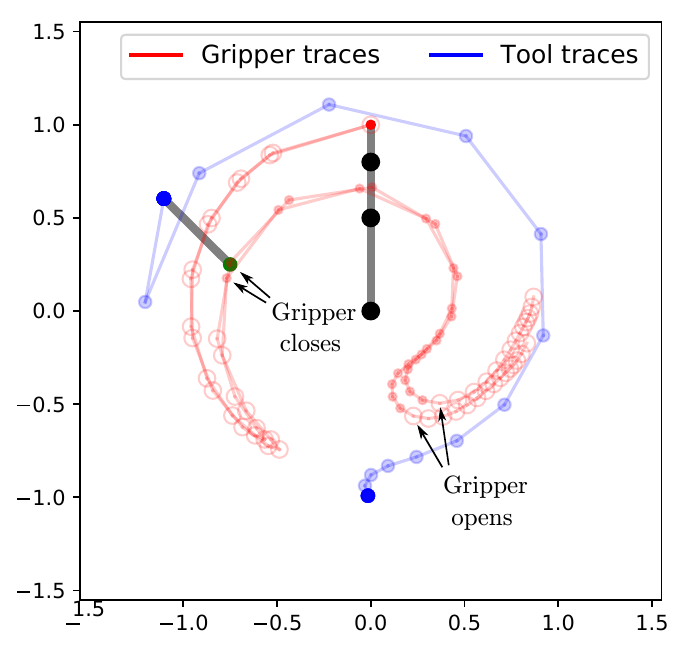}}
\caption{Example of a mutation of a movement reaching the tool, for each mutation operator. With the \texttt{SSPMutation} operator, the two gripper trajectories start to diverge only when the stick is grasped such that the mutated movement also grasps the stick, whereas with the \texttt{FullMutation}, the mutation starts right from the beginning of the movement, which in this case makes the mutated movement miss the stick.}
\label{example_traj}
\end{figure}

The efficiency of the Stepping-Stone Preserving Mutation operator (\texttt{SSPMutation}, see Sec. \ref{temporal}) relies on its ability to preserve the part of the movement that reaches a stepping-stone in the environment, and explore only after the target object started to move in the previous movement being mutated.
For instance, the movement would grab the tool without modification, and explore once the controlled toy started to move.
To illustrate this mechanism, let us look at actual mutations depending on the mutation operator. Fig. \ref{example_traj} shows one movement that reached the magnet tool together with one mutation of this movement, for each mutation operator. The red trajectories are the traces of the gripper (with a circle when open and a point when closed), and the blue trajectories are the traces of the magnet stick. We also plot the initial position of the arm and the magnet stick.
We see that in the case of the \texttt{SSPMutation} operator, the two red trajectories start to diverge only when the stick is grasped such that the mutated movement also grasps the stick, whereas with the \texttt{FullMutation}, the mutation starts right from the beginning of the movement, which in this case makes the mutated movement miss the stick.

\begin{figure}[t]
\centering
\subfloat[Magnet Tool]{\includegraphics[width=0.48\textwidth]{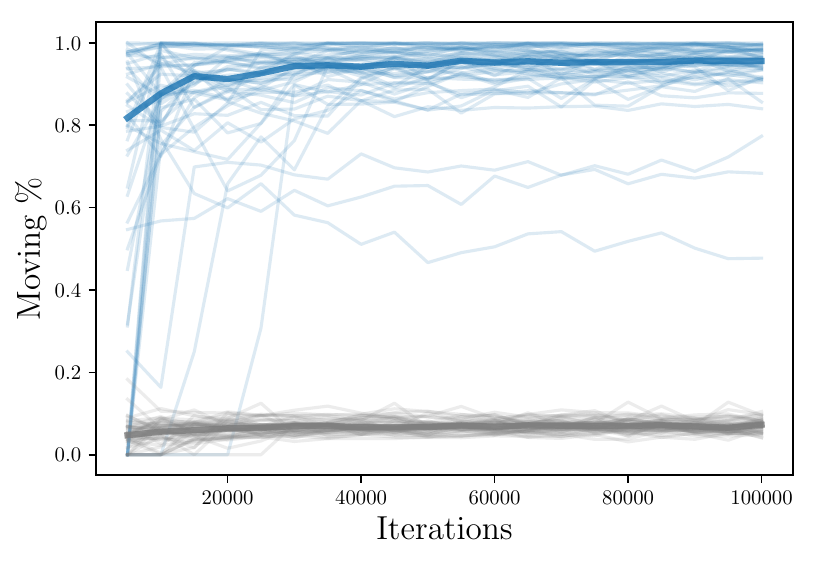}}\hspace{0.2cm}
\subfloat[Magnet Toy]{\includegraphics[width=0.48\textwidth]{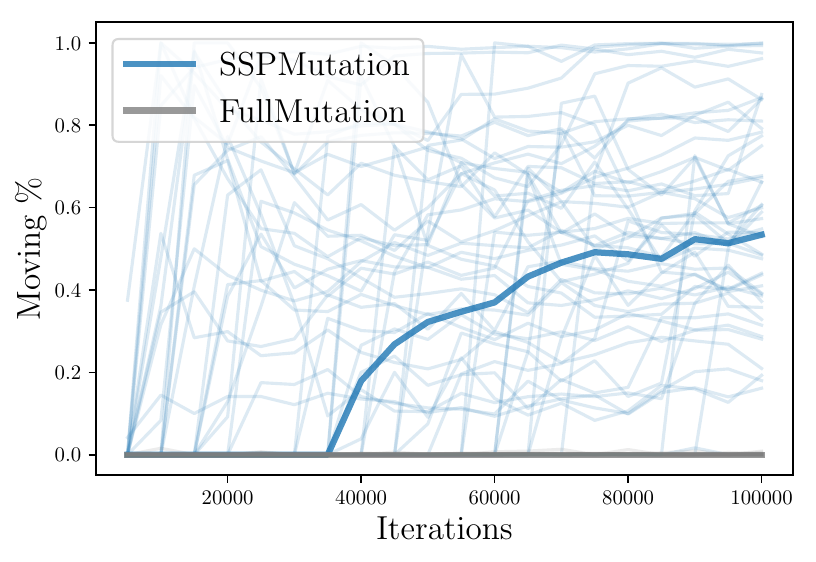}}
\caption{Comparison of the \texttt{SSPMutation} and \texttt{FullMutation} mutation operators. We show the proportion of iterations that allowed to (a) move the magnet tool while exploring this tool, and (b) move the magnet toy while exploring this toy, with $50$ different seeds and median.
With the \texttt{FullMutation} operator, at the end of the runs agents succeed to move the tool in $7\%$ of iterations versus $95\%$ for the \texttt{SSPMutation} operator, and to move the toy in $0.9\%$ of iterations versus $53\%$.}
\label{transfer_2DSimu_FullMutation}
\end{figure}

We measure how many times the agents succeed to move a tool when they are exploring it, or to move a toy when they are exploring the toy, depending on the mutation operator.
Fig. \ref{transfer_2DSimu_FullMutation} shows the proportion of the iterations that allowed to (a) move the magnet tool when this tool is the goal object, (b) move the magnet toy when this toy is the target object, with 50 different runs in the 2D simulated environment (individual runs and median).
We can see that with the \texttt{FullMutation} operator, at the end of the runs agents succeed to move the tool in $7\%$ of iterations targeted at exploring this tool, versus $95\%$ for the \texttt{SSPMutation} operator, and to move the toy in $0.9\%$ of iterations targeted at exploring this toy versus $53\%$.

\begin{figure}[t]
\centering
\subfloat[Magnet Tool]{\includegraphics[width=0.48\textwidth]{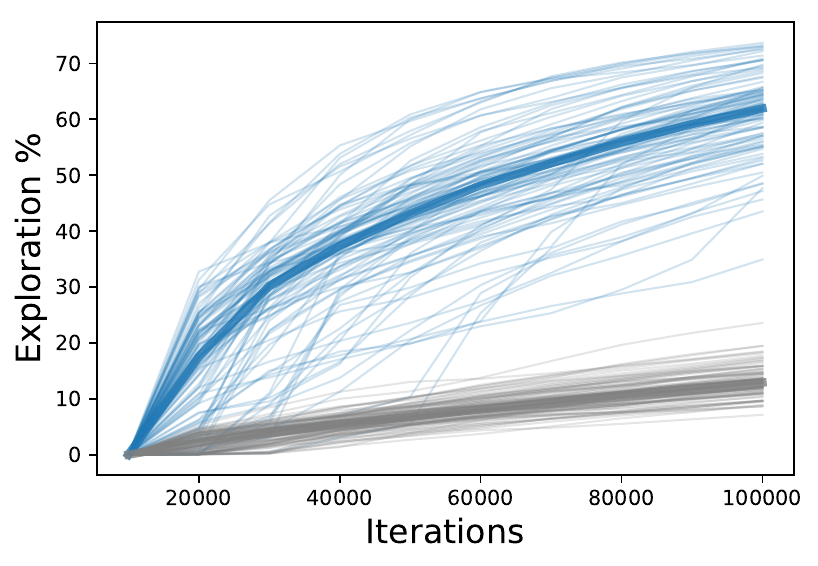}}\hspace{0.2cm}
\subfloat[Magnet Toy]{\includegraphics[width=0.48\textwidth]{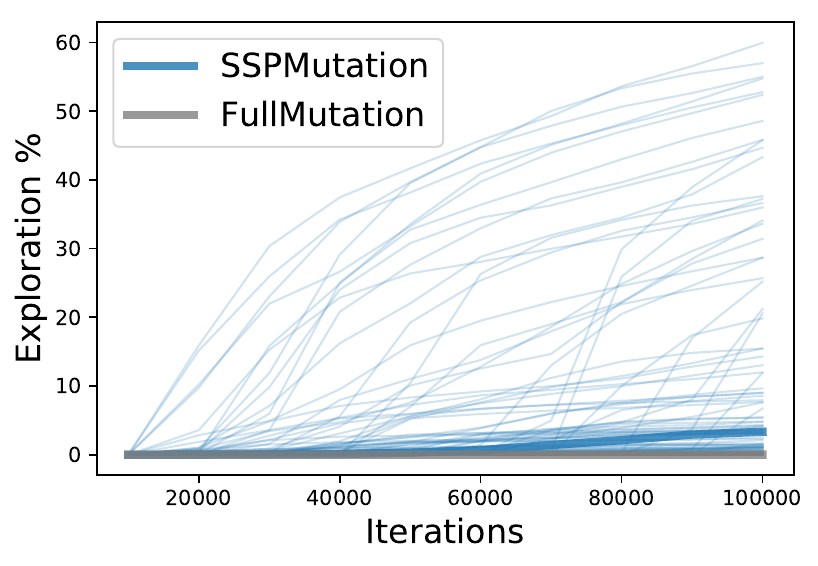}}
\caption{Exploration efficiency depending on the mutation operator. 
\texttt{FullMutation} results in a much lower exploration for the magnet tool and toy compared to the stepping-stone preserving operator.}
\label{explo_operators}
\end{figure}

The ability of \texttt{SSPMutation} to explore while still moving the target object with a high probability directly improves exploration.
Fig. \ref{explo_operators} shows the exploration results of AMB agents with the \texttt{SSPMutation} or \texttt{FullMutation} operators in the 2D simulated environment in 100 runs with different seeds.
The exploration results of the \texttt{FullMutation} operator are much lower for the magnet tool (median $13\%$ vs  $62\%$) and magnet toy (median $0\%$ vs $3\%$, max $0.5\%$ vs $60\%$).

\begin{figure}[t]
\centering
\subfloat[Magnet Tool]{\includegraphics[width=0.48\textwidth]{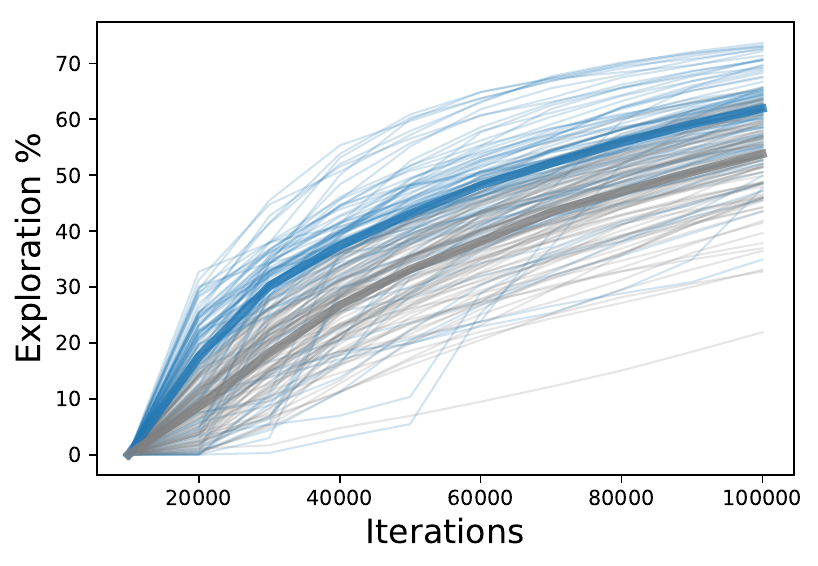}}\hspace{0.2cm}
\subfloat[Magnet Toy]{\includegraphics[width=0.48\textwidth]{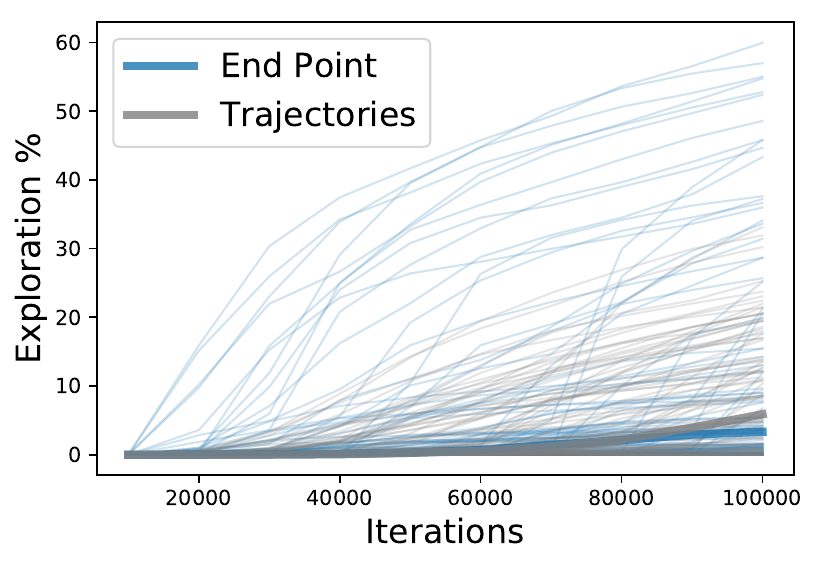}}
\caption{Exploration in 2D simulated environment depending on the goal encoding. 
The encoding with the end position of objects resulted in a slightly better exploration than with trajectories for the magnet tool, and a similar median for the magnet toy but with more variance: standard deviation of $17\%$ vs $8.6\%$ at 100k iterations for the magnet toy.}
\label{explo_traj}
\end{figure}

\subsubsection{Encoding of Goals}

After executing a movement, the agent receives a sensory feedback containing information about the movement of objects in the environment.
The agent then uses the sensory space as an encoding for sampling new goals to reach.
In the 2D simulated environment, we defined the sensory feedback as the position of each object at the end of the movement.
In the robotic environment, as the joysticks may move during the movement but come back by themselves at their rest position, we used a sensory feedback containing information about the whole trajectory of objects as the sequence of their position at $5$ time steps during the movement.
In this section, we study the influence of the goal encoding on exploration efficiency in the 2D simulated environment.

Fig. \ref{explo_traj} shows the exploration evolution of AMB agents depending on the encoding of goals: with object trajectories or end points, in the 2D simulated environment. The exploration is a measure of the proportion of cells reached in a discretization of the space of the last position of each object.
The encoding with the end position of each object resulted in a slightly better exploration than with trajectories for the magnet tool, and a similar median for the magnet toy but with more variance: with standard deviation of $17\%$ vs $8.6\%$ at 100k iterations for the magnet toy.
The trajectory encoding represents the whole trajectory of each object instead of the final point only.
This is not strictly needed to represent if a tool or a toy has moved in this environment so the end point encoding may be more efficient once the objects are discovered, however the trajectory encoding helps to explore trajectories with more diversity, for the hand or other objects, and thus to discover hard objects in the first hand. With the trajectory encoding, the exploration of objects difficult to move the first time is thus slower once discovered, but they are more often discovered.

\begin{figure}[t]
\centering
\includegraphics[width=0.48\textwidth]{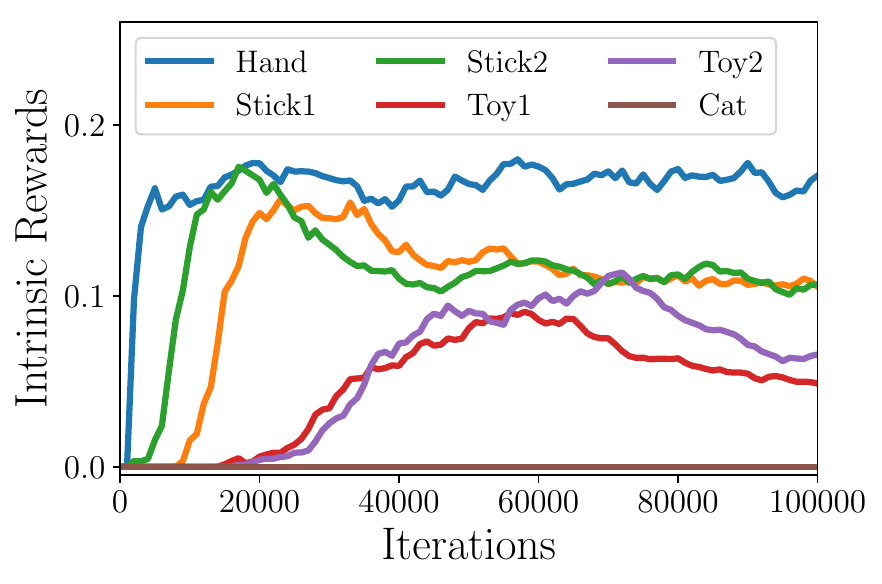}\hspace{0.2cm}
\includegraphics[width=0.48\textwidth]{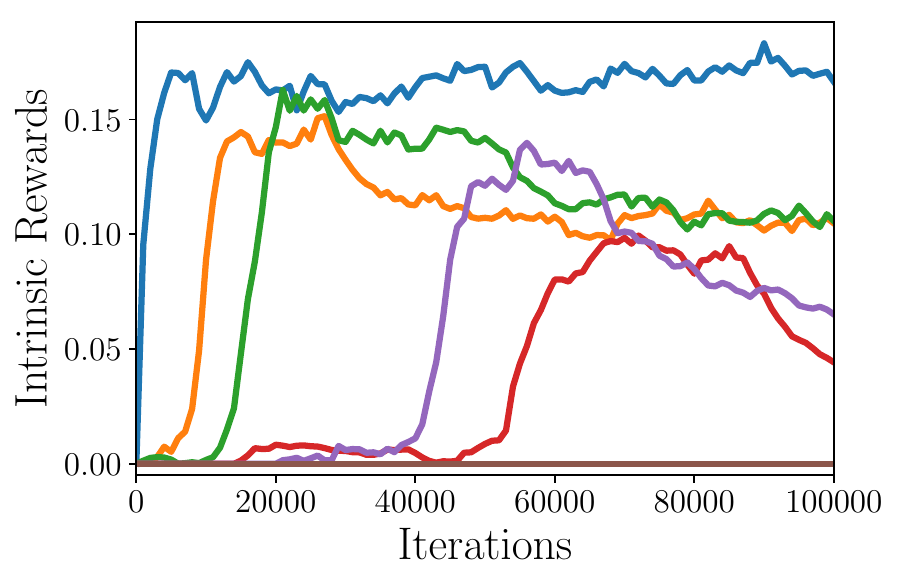}
\caption{Examples of intrinsic rewards in the 2D simulated environment, with an encoding of goals as object trajectories. The sensory spaces are higher-dimensional and take more iterations to be well covered such that the learning progress decreases.}
\label{interests_simu2D_traj}
\end{figure}

Fig. \ref{interests_simu2D_traj} shows examples of interest curves with the goal encoding using trajectories of objects.
As the goal spaces are of much larger dimensionality using the object trajectories than with the end position, it takes a longer time to cover the whole sensory space with reached trajectories such that the self-computed interest to explore the hand is higher than with end positions (comparing with Fig. \ref{interests}(top)) and the interest in all spaces takes more time to decrease.

The trajectory encoding is more interesting in environments where the full trajectory of a tool is of importance to control an object, such as in our robotic environment where joysticks come back at their rest position by themselves such that their end position is not informative to predict the end position of the controlled object. We thus use this trajectory encoding in the robotic environment, but we use the end point encoding in the Minecraft Mountain Cart environment.

\begin{figure}[t]
\centering
\subfloat[Magnet Tool]{\includegraphics[width=0.48\textwidth]{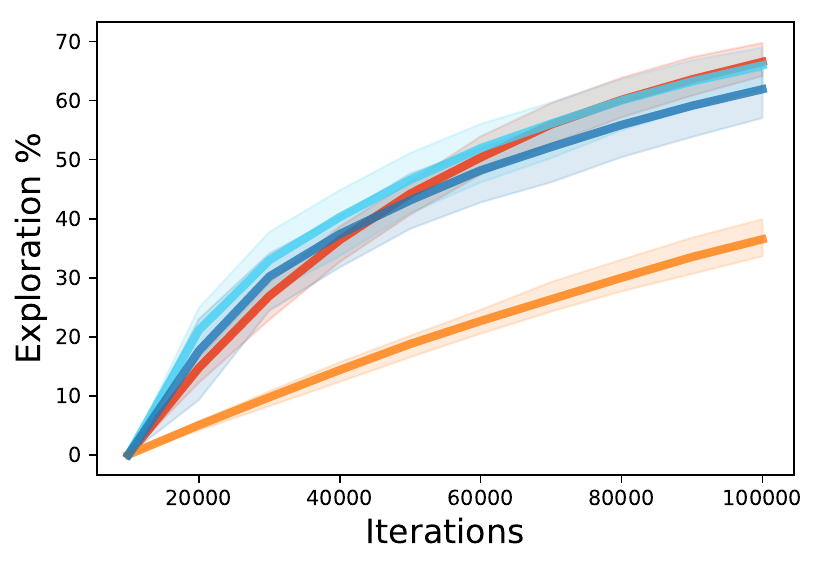}}\hspace{0.2cm}
\subfloat[Magnet Toy]{\includegraphics[width=0.48\textwidth]{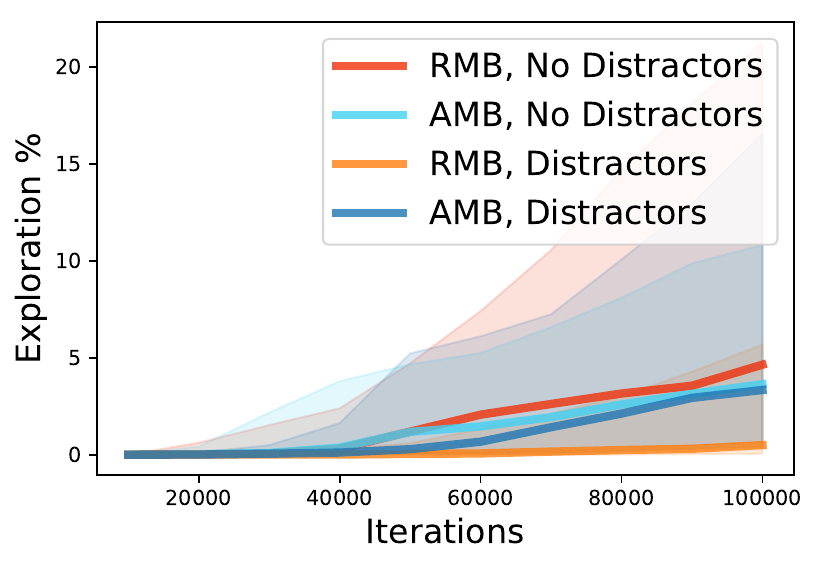}}
\caption{Exploration efficiency of Active Model Babbling and Random Model Babbling depending on the presence of distractors in the 2D Simulated environment. 
The median of $100$ runs is plotted together with a shaded area representing the 25-75 percentiles.
The efficiency of RMB agents decreases when we add distractors, whereas AMB agents, through their self-estimation of their learning progress to move each object, focus on learnable objects despite having distractors in the environment.}
\label{explo_distractors}
\end{figure}

\subsubsection{Static and Random Distractors}

In the three tool-use environments, we included distractor objects to harden exploration as those objects can't be controlled by the agent but are however part of their sensory feedback: some of them are static, and some of them move independently of the agent.
The Active Model Babbling agents monitor their learning progress to move objects, such that their estimation of their progress to move static object is zero, and to move other uncontrollable objects is low compared to controllable objects.
Here we evaluate the exploration efficiency of AMB and RMB agents in the 2D simulated environment in the presence and absence of distractors to evaluate their robustness to distractors.

Fig  \ref{explo_distractors} shows the exploration results depending on the learning condition (RMB vs AMB) and environment condition: with 2 random distractors and 8 static distractors (Distractors) vs without distractors (No Distractors), in the magnet tool and toy spaces (median and 25/75 percentiles of 100 seeds). 
The RMB agents show a similar exploration without distractors compared to AMB agents.
However, we can see that the RMB agents do not cope with distractors while AMB agents do not show a significant decrease in exploration when distractors are added.
The learning progress monitoring is thus an efficient mean to discriminate learnable objects from other objects and thus to focus exploration by choosing most goals for learnable objects.

\section{Related Work}

Early models of intrinsically motivated reinforcement learning (also called curiosity-driven learning) have been used to drive efficient exploration in the context of target tasks with rare or deceptive rewards \citep{jurgen1991possibility, barto2013intrinsic} or in the context of computational modeling of open-ended unsupervised autonomous learning in humans \citep{kaplan2004maximizing,oudeyer_intrinsic_2007}. Reviews of the historical development of these methods and their links with cognitive sciences and neuroscience can be found in \citet{baldassarre2013intrinsically, gottlieb_information-seeking_2013, oudeyer2016intrinsic}.

Several lines of results have shown that intrinsically motivated exploration and learning mechanisms are particularly useful in the context of learning to solve reinforcement learning problems with sparse or deceptive rewards. For example, several state-of-the-art performances of Deep Reinforcement Learning algorithms, such as letting a machine learn how to solve complex video games, have been achieved by complementing the extrinsic rewards (number of points won) with an intrinsic reward pushing the learner to explore for improving its predictions of the world dynamics \citep{bellemare2016unifying, houthooft2016vime}. An even more radical approach for solving problems with rare or deceptive extrinsic rewards has been to completely ignore extrinsic rewards, and let the machine explore the environment for the sole purpose of learning to predict the consequences of its actions \citep{jurgen1991possibility,oudeyer_intrinsic_2007}, to achieve self-generated goals \citep{baranes_active_2013,oudeyer_what_2007}, or to generate novel behavioral features \citep{lehman2011abandoning}. This was shown for example to allow agents to learn to play some video games without ever observing the extrinsic reward \citep{pathak2017curiosity}. 

Some approaches to intrinsically motivated exploration have used intrinsic rewards to value visited actions and states through measuring their novelty or the improvement of predictions that they provide, e.g. \citet{sutton1990integrated, dayan1996exploration,jurgen1991possibility,oudeyer_intrinsic_2007} or more recently \citet{bellemare2016unifying,houthooft2016vime,pathak2017curiosity}. However, organizing intrinsically motivated exploration at the higher level of goals, by sampling goals according to measures such as competence progress \citep{oudeyer_what_2007}, has been proposed and shown to be more efficient in contexts with high-dimensional continuous action spaces and strong time constraints for interaction with the environment \citep{baranes_active_2013}.

Several proposed methods are related to IMGEP, including \citet{gregor2016variational}, \citet{dosovitskiy2016learning} and \citet{kulkarni2016hierarchical}, however they have considered notions of goals restricted to the reaching of states or direct sensory measurements, did not consider goal-parameterized rewards that can be computed for any goal, used different intrinsic rewards, and did not evaluate these algorithms in robotic setups. The notion of auxiliary tasks is also related to IMGEP in the sense that it allows a learner to acquire tasks with rare rewards by adding several other objectives which increase the density of information obtained from the environment \citep{jaderberg2017reinforcement,riedmiller2018learning}. Another line of related work \citep{srivastava2013first} proposed a theoretical framework for automatic generation of problem sequences for machine learners, however it has focused on theoretical considerations and experiments on abstract problems. 

\textcolor{black}{The present work focused on proposing AMB, an IMGEP instantiation able to deal with modular goal spaces. Another interesting and parallel perspective is to focus on designing intrinsically motivated agents able to learn modular skills, e.g. hierarchical RL agents able to combine atomic skills. For instance, in a simulated end-effector positioning environment, \cite{santucci2019autonomous} used intrinsic signals to generate sub-goals allowing their agent to derive skill-chains towards end-effector goals. In \cite{blaes2019control}, authors combine a learning progress signal with planning strategies to learn modular skills in a simulated tool use environment. Closely related to our work, \cite{forestier2016curiosity} extends a former version of our AMB agent in a simulated tool use environment with hierarchical goal spaces, enabling to decompose explicitly trajectory generation. Note that such aforementioned hierarchical learners can be formalized as specific forms of IMGEPs featuring multistep goal sampling policies.}

Several strands of research in robotics have presented algorithms that instantiate such intrinsically motivated goal exploration processes \citep{baranes2010intrinsically, rolf}, using different terminologies such as contextual policy search \citep{kupcsik2017model,queisser2016incremental}, or formulated within an evolutionary computation perspective such as Novelty Search \citep{lehman2011abandoning} or Quality Diversity \citep{cully2015robots, cully2017quality} (see next sections).
In \citet{forestier2016modular}, the implemented modular population-based version of IMGEP was called Model Babbling.
Several variants of Model Babbling were evaluated: Random Model Babbling is a variant where the goal space is chosen randomly and goals are chosen randomly in the goal space and Active Model Babbling one where the goal space is chosen based on the learning progress to control each object.
Both implementations are instances of IMGEP as the goal spaces are generated autonomously from the sensory spaces and no ``expert knowledge'' has been given to the algorithm.

In machine learning, the concept of curriculum learning \citep{bengio2009curriculum} has most often been used in the context of training neural networks to solve prediction problems. Many approaches have used hand-designed learning curriculum \citep{Sutskever14}, but recently it was shown how learning progress could be used to automate intrinsically motivated curriculum learning in LSTMs \citep{graves2017automated}. However, these approaches have not considered a curriculum learning of sets of reinforcement learning problems, which is central in the IMGEP framework formulated with goals as fitness functions, and assumed the pre-existence of a database with learning exemplars to sample from.
In recent related work, \citep{matiisen2017teacher,portelas2019teacher} studied how intrinsic rewards based on learning progress could also be used to automatically generate a learning curriculum with discrete and continuous sets of reinforcement learning problems, respectively. In contrast to the present study, both works did not consider high-dimensional modular problem spaces. See \citet{portelas2020automatic} for a review of other recent curriculum learning methods applied to reinforcement learning scenarios. The concept of ``curriculum learning'' has also been called ``developmental trajectories'' in prior work on computational modeling of intrinsically motivated exploration \citep{oudeyer_intrinsic_2007}, and in particular on the topic of intrinsically motivated goal exploration \citep{baranes_active_2013,forestier2017unified}.

The concepts of goals and of learning across goals have been introduced in machine learning in \citet{kaelbling1993learning} with a finite set of goals.
Continuous goals were used in Universal Value Function Approximators \citep{schaul2015universal}, where a vector describing the goal is provided as input together with the state to the neural network of the policy and of the value function.
\textcolor{black}{In \cite{her2017}, authors drastically improve the sample efficiency of such goal-conditioned RL learners using the Hindsight Experience Replay (HER) algorithm, which implements a form of hindsight learning related to ours.
HER is a passive and offline mechanism. It reinterprets trajectories collected with a given target goal with
respect to a different goal, in order to retrospectively generate positive rewards used to compute meaningful learning gradients for the policy.
In contrast, we consider an active and online form of hindsight learning used for exploration. Given a new goal, IMGEP agents condition their exploration policy based on a retrospective analysis of previous goal attempts, whose corresponding policies can potentially be fitting for the new goal.
In classical goal-conditioned works \citep{her2017,schaul2015universal}, goals are not modular, and are considered extrinsic to the agent, with extrinsic rewards that can contain expert knowledge about the tasks being learned.
The learning problem is not formulated as an autonomous learning problem where the agent has to explore the most diverse set of states and skills on its own.
Although not modular and population-based, other recent works do consider goal-conditioned mechanisms to control exploration for autotelic deep reinforcement learning agents, e.g. \citep{pong2020skewfit,pitis2020, choi_variational_2021}, in a variant of the IMGEP framework called RL-IMGEP \cite{colas2020intrinsically}.
Most related to the present work is the CURIOUS system \citep{colas2018curious}, building on an early version of this paper to transpose the population-based IMGEP algorithms we introduced within the reinforcement learning framework, leveraging an extension of Universal Value Function Approximators \citep{schaul2015universal}. 
This is a particular implementation of the IMGEP architecture using reinforcement learning techniques, using a unique monolithic (multi-task multi-goal) policy network, that learns from on a replay buffer filled with rollouts on task and goals of high learning progress.}
Also, using a population-based IMGEP architecture can help bootstrap a deep RL agent \citep{pmlr-v80-colas18a}. Filling the replay buffer of a deep RL agent with exploratory trajectories collected by an IMGEP algorithm kick-starts the RL agent by enhancing its exploratory abilities. It combines the efficient exploration of population-based IMGEP agents with the efficient fine tuning of policies offered by deep RL agents with a function approximator based on gradient descent.

\subsection{IMGEP and Novelty Search}

In Novelty Search evolutionary algorithms, no objective is given to the optimization process, which is driven by the novelty or diversity of the discovered individuals \citep{lehman2011abandoning}.
In this implementation, an archive of novel individuals is built and used to compute the novelty of the individuals of the current generation of the evolutionary algorithm.
If the novelty of a new individual is above a threshold, it is added to the archive.
Different measures of novelty can be used, a simple one being the average distance of the individual to its closest neighbors in the archive, the distance being measured in a behavioral space defined by the user.
Then, to generate the population of the next generation, the individuals with a high measured novelty are reused, mutated or built upon.

Although designed in an evolutionary framework, the Novelty Search (NS) algorithm can be framed as a population-based IMGEP implementation, assuming that the behavioral space and its distance measure can be self-generated by the algorithm.
Indeed, we can define an IMGEP goal space based on the NS behavioral space, with each behavior in that space generating the corresponding goal of reaching that behavior, with a fitness function defined as the negative distance between the target behavior and the reached behavior.
In IMGEP, if the goal $g$  (defining the target behavior) is chosen randomly, the algorithm can then reuse the previous reached behaviors that give the highest fitness to reach the current goal $g$, which are the closest reached points in the behavioral space.
The key similarity between our population-based implementations of IMGEP and Novelty Search is that the previous behavior the closest to the current random target behavior is a behavior with high novelty on average.
Indeed, a random point in a space is more often closer to a point at the frontier of the explored regions of that space which is thus a high-novelty point.
Randomly exploring behaviors or mutating high-novelty behaviors are therefore efficient for the same reasons.

Abandoning the external objectives and focusing on the novelty of the behaviors in \citet{lehman2011abandoning} can be seen in the lens of the IMGEP framework as embracing all self-generated objectives.

\subsection{IMGEP and Quality Diversity}

The Novelty Search approach stems from the fact that in many complex optimization problems, using a fitness function to define a particular objective and relying only on the optimization of this function may not be efficient: indeed, there may be unknown complex successive stepping-stones that need to be reached before the final objective can be approached \cite{stanley2015greatness}.
Relying on novelty allows to reach stepping-stones and build upon them to explore new behaviors even if the objective does not get closer.
However, when the behavioral space is high-dimensional, pursuing the final objective is still useful to drive exploration together with the search for novelty \citep{cuccu2011novelty}.
The Quality Diversity approach combines the search for diversity from Novelty Search approaches and the use of an external objective function to ensure the Quality of the explored individuals \citep{cully2017quality,lehman2011evolving,cully2015robots}. 

In the MAP-Elites algorithm \citep{cully2015robots}, the behavioral space is discretized into a grid of possible behaviors, and a fitness function is provided to assess the quality of individuals according to a global objective.
Each new individual is assigned to a behavioral cell in this grid and is given a quality value with the quality function.
The population of the next generation of the evolutionary algorithm is mutated, in its simplest version, from a random sample of the set of the best quality individual of all cells.
In more sophisticated versions, the parents used for evolving the next generation are selected based on their quality, the novelty of the cells, or a tradeoff between quality and novelty.

In the applications of this algorithm, the fitness function is an extrinsic objective.
For instance, in \citet{cully2015robots} robot controllers are evolved to find efficient robot walking behaviors.
The fitness function given to the algorithm is the speed of the robot, while the descriptors of a behavior can be the orientation, displacement, energy used, deviation from a straight line, joint angles, etc.
The algorithm thus tries to find efficient walking behaviors for a diverse set of of behavioural constraints.

The concept of Quality Diversity algorithms is thus different from the concept of intrinsically motivated exploration. However Quality Diversity algorithms could be used with a fitness function that is intrinsically generated by the algorithm.
The functioning of the algorithm given this fitness function can then be seen as a population-based instantiation of the IMGEP framework.
Indeed, each cell of the behavioral grid can generate one different IMGEP goal with a particular fitness function returning the quality of the individual if its behavior falls into that cell and zero otherwise.
In MAP-Elites \citep{cully2015robots}, the next generation of individuals is mutated from a random sample of elites (the best quality individual of each non-void cell).
In an IMGEP settings with those goals, the MAP-Elites sampling is equivalent to selecting a random goal from the set of goals that had a non-zero fitness in the past.
When such a goal is selected, the next IMGEP exploration loop then reuses, in its simplest version, the sample with the best fitness for that goal, which corresponds to the elite, and also mutates it.

A common process among Novelty Search, Quality Diversity and IMGEP approaches is that exploration is made efficient by leveraging two principles: 1) a diversity of solutions continues to be explored even if they are non-optimal, and 2) when exploring solutions to a given region/cell/goal, the algorithm can find solutions to other regions/cells/goals, which are registered and can be leveraged later.


\subsection{IMGEP and Reinforcement Learning}

In our setting, the fitness functions $f_g$ have two particularities in comparison with the concept of ``reward function'' as often used in the RL literature.
The first particularity is that these fitness functions are computed based on the trajectory $\tau$ resulting from the execution of policy $\Pi$, and thus consider the whole interaction of the agent and its environment during the execution of the policy, for instance taking into account the energy used by the agent or the trajectory of an object.
Therefore they are not necessarily Markovian if one considers them from the perspective of the level of state transitions $s_t$.

The second particularity is that since the computation of the fitness $f_g(\tau)$ is internal to the agent, it can be computed any time after the experiment and for any goal $g \in \mathcal{G}$, not only the particular goal that the agent was trying to achieve.
Consequently, if the agent stores the observation $\tau$ resulting from the exploration of a goal $p_1$, then when later on it self-generates goals $g_2,g_3,...,g_i$ it can compute, without further actions in the environment, the associated fitness $f_{g_i}(\tau)$ and use this information to improve over these goals $g_i$.
This property is essential as it enables direct reuse of data collected when trying to achieve a goal for later exploring other goals.
It is leveraged for curriculum learning in Intrinsically Motivated Goal Exploration Processes.

\section{Discussion}

This paper provides a formal framework for an algorithmic architecture called Intrinsically Motivated Goal Exploration Processes (IMGEP), enabling to build autotelic agents.
This framework enables a unified description of various related algorithms that share several principles: exploration is driven by self-generated goals, exploring towards a goal gives information that can be reused to improve solutions for other goals, and intrinsic rewards can help the construction and selection of goals.

The IMGEP framework is both compact and general.
The goals are defined through parameterized fitness functions and therefore can represent any kind of objective that can be computed from the information stored and available to the agent.
The policies can be implemented by any algorithm that can learn a function that takes a goal as input and outputs actions to explore this goal, such as a monolithic neural network \citep{colas2018curious} or a population-based policy \citep{forestier2016modular}.

We presented a particular type of modular IMGEP architecture (AMB) that formalizes the Model Babbling approach described in \citet{forestier2016modular}. It leverages population-based policies and a goal sampling process that uses learning-progress and is modular: spatial modularity (the agent generates one goal space for each object in the environment), and temporal modularity (the temporal structure of objects' movements is leveraged for a more efficient goal exploration).

The IMGEP framework is well suited to organize exploration in environments where, in order to avoid local optima and find advanced behaviors or phenotypes, enough time should be allocated to the continued search for diverse non-optimal solutions. Interesting unexpected stepping-stones can be discovered in the exploration process and built upon afterwards.
The framework is most useful when the stepping-stones or the targets are unknown to the expert user or too complex such that they cannot easily be directly represented and optimized.
In that case the use of intrinsic motivations for the exploration of goals can help to discover and build upon a diverse set of stepping-stones.
Furthermore, the use of intrinsic rewards, e.g. based on the monitoring of the learning progress in achieving goals, can further improve the efficiency of exploration by focusing on the most interesting problems and avoiding the ones that bring no more information.

We studied IMGEP implementations in different tool-use environments.
We designed the first robotic experiment where an intrinsically-motivated humanoid robot discovers a complex continuous high-dimensional environment and succeeds to explore and learn from scratch that some objects can be used as tools to act on other objects. We also created the Minecraft Mountain Cart environment, where an intrinsically motivated Minecraft agent discovered various tool use skills, and chaining them to further discover more complex skills.

We evaluated different variants of Intrinsically Motivated Goal Exploration Processes and showed the efficiency of variants that do not use a hand-designed learning (RMB and AMB conditions) for discovery of the most complex affordances. Furthermore, when the agent monitors its learning progress with intrinsic rewards (AMB), it autonomously develops a learning sequence, or curriculum, from the easiest to the most complex tasks, and explores more efficiently than without those intrinsic rewards. Also, the comparison between agents only exploring one interesting problem space (SGS) versus all spaces (AMB) shows that if an engineer were to specify the target problems to solve (e.g. move the ball), then it would be more efficient to also explore all other possible intrinsic goals to develop new skills that can serve as stepping-stones to solve the target problem.

Compared to other approaches, our approach to implement population-based IMGEPs is sample-efficient, with a number of iterations of 20k in the real robotic setup, 40k in the Minecraft environment, and 100k in the 2D simulated one. 
Approaches such as Quality Diversity need 40M iterations for example for the learning of a hexapod's locomotion \citep{cully2015robots}, and deep reinforcement learning agents also require millions of training steps, e.g. 2M steps in the Atari game Montezuma's Revenge \citep{kulkarni2016hierarchical}, or 50M in a Doom-like first-person shooter \citep{dosovitskiy2016learning}, including for RL-based IMGEPs \cite{colas2018curious,pong2020skewfit}.
The population-based IMGEP implementation in its simplest form, with a Nearest Neighbor look-up used as inverse models, is also computationally efficient, as we have run the 20k iterations of the real robotic experiment on a raspberry Pi 3.

A limitation of IMGEP algorithms presented in this paper is that we suppose agents already have a perceptual system allowing them to see and track objects, as well as associated handcrafted modular spaces of representations to encode goals. 
Recent works studied in simulation how the learning of a representation of objects from pixels and its use as a goal space could be achieved for population-based intrinsically motivated agents \citep{pere2018unsupervised}, including modular goal space representations \cite{pmlr-v87-laversanne-finot18a}. Population-based IMGEPs also inspired subsequent works on learning goal spaces from pixels for monolithic deep reinforcement learning agents in real-world robotic manipulation scenarios \citep{nair2018rig,pong2020skewfit}.

\textcolor{black}{Another limitation of our approach originates from adopting a population-based policy generation system -- i.e. a set of frozen task-experts -- which hinders generalization abilities compared to when training a single multi-purpose DRL policy through backpropagation. Although our experiments showcased that our agents were able to operate on varying initial contexts, considering more complex initial state distributions, in which only a small subset of goals (or goal spaces) are achievable in each episode would impair the performances of our current agents. An interesting avenue would be to find ways to integrate context in goal selection, e.g. using a neural network based goal selection as in the Setter-Solver architecture \citep{racaniere2019automated}.}

The IMGEP framework has also been applied to the exploration of very different setups in other scientific domains.
In \citet{grizou2020exploration}, an IMGEP implementation allowed to discover a variety of droplet behaviors in a chemical system of self-propelling oil droplets in water, where the exploration parameters were the concentrations of the different components of the oil droplet among others.
In yet another domain, \citet{reinke2019intrinsically} showed that the IMGEP framework with a goal representation learned online could find self-organized patterns in the complex morphogenetic system Lenia, a continuous game-of-life cellular automaton.

\acks{
We would like to thank Alexandre Péré, Cédric Colas, Adrien Laversanne-Finot and Olivier Sigaud for their valuable comments on an earlier version of this manuscript and Damien Caselli for his technical help.
}



\vskip 0.2in
\bibliography{sample}

\end{document}